\documentclass[10pt,journal,compsoc]{IEEEtran}

\ifCLASSOPTIONcompsoc
  \usepackage[nocompress]{cite}
\else
  \usepackage{cite}
\fi

\ifCLASSINFOpdf
\else
\fi
\usepackage{amsmath}
\usepackage{array}

\makeatletter
\let\MYcaption\@makecaption
\makeatother

\usepackage[font=normalsize,labelfont=sf,textfont=sf]{subcaption}

\makeatletter
\let\@makecaption\MYcaption
\makeatother

\usepackage{amssymb}

\usepackage{MnSymbol}

\usepackage{multirow}
\usepackage{siunitx}
\usepackage{tabularx} 
\usepackage{makecell} 
\usepackage{booktabs}
\usepackage{pifont}

\usepackage{hyphenat}
\usepackage{enumitem}
\usepackage{xcolor,soul}
\usepackage{colortbl}

\usepackage[nolist,nohyperlinks,printonlyused]{acronym}
\usepackage[export]{adjustbox}
\usepackage{esvect}
\usepackage{pgfplotstable} 
\usepackage{tikz}
\usetikzlibrary{shadows,intersections}

\usepackage[framemethod=tikz]{mdframed} 

\usepackage{etoolbox}  

\newtoggle{draft_mode}
\toggletrue{draft_mode}

\togglefalse{draft_mode}

\iftoggle{draft_mode}
{
    \usepackage{todonotes}
    \paperwidth=\dimexpr \paperwidth + 6cm\relax
    \oddsidemargin=\dimexpr\oddsidemargin + 3cm\relax
    \evensidemargin=\dimexpr\evensidemargin + 3cm\relax
    \marginparwidth=\dimexpr \marginparwidth + 3cm\relax
}{
    \usepackage[disable]{todonotes}
}

\usepackage[breaklinks=true,bookmarks=false]{hyperref}
\usepackage[shortcuts]{extdash}  

\usepackage{pbox}

\DeclareSIUnit{\nothing}{\relax}

\newtoggle{use_paragraphs}
\toggletrue{use_paragraphs}

\newcommand{\dittotikz}{%
    \tikz{
        \draw [line width=0.12ex] (-0.2ex,0) -- +(0,0.8ex)
            (0.2ex,0) -- +(0,0.8ex);
        \draw [line width=0.08ex] (-0.6ex,0.4ex) -- +(-1.5em,0)
            (0.6ex,0.4ex) -- +(1.5em,0);
    }%
}

\newcommand{\methodname}{Local-DIFs}%
\newcommand{\methodabbr}{LDIF}%
\newcommand{\nn}{\methodabbr{}}%
\newcommand{\scssnetUpsample}{\methodname{}-CBN}
\newcommand{\scssnetSingle}{\methodname{}-c3}

\iftoggle{use_paragraphs}
{
    \newcommand{\psection}[1]{{\boldparagraph{{#1}}}}
}
{
    \newcommand{\psection}[1]{{\subsubsection{{#1}}}}
}

\pgfplotsset{compat=1.3}
\pgfplotsset{
  tick label style = {font=\rmfamily},
  every axis label = {font=\rmfamily},
  legend style = {font=\rmfamily},
  label style = {font=\rmfamily}
}

\newcolumntype{R}[2]{%
    >{\adjustbox{angle=#1,lap=\width-(#2)}\bgroup}%
    l%
    <{\egroup}%
}
\newcommand*\rotsem{\multicolumn{1}{R{55}{1em}}}
\newcommand*\rotsemXX{\multicolumn{1}{R{25}{1em}}}
\newcommand*\rotsemXXX{\multicolumn{1}{R{20}{1em}}}

\definecolor{car}{rgb}{0.39215686274509803, 0.5882352941176471, 0.9607843137254902}
\definecolor{bicycle}{rgb}{0.39215686274509803, 0.9019607843137255, 0.9607843137254902}
\definecolor{motorcycle}{rgb}{0.11764705882352941, 0.23529411764705882, 0.5882352941176471}
\definecolor{truck}{rgb}{0.3137254901960784, 0.11764705882352941, 0.7058823529411765}
\definecolor{othervehicle}{rgb}{0.0, 0.0, 1.0}
\definecolor{person}{rgb}{1.0, 0.11764705882352941, 0.11764705882352941}
\definecolor{bicyclist}{rgb}{1.0, 0.1568627450980392, 0.7843137254901961}
\definecolor{motorcyclist}{rgb}{0.5882352941176471, 0.11764705882352941, 0.35294117647058826}
\definecolor{road}{rgb}{1.0, 0.0, 1.0}
\definecolor{parking}{rgb}{1.0, 0.5882352941176471, 1.0}
\definecolor{sidewalk}{rgb}{0.29411764705882354, 0.0, 0.29411764705882354}
\definecolor{otherground}{rgb}{0.6862745098039216, 0.0, 0.29411764705882354}
\definecolor{building}{rgb}{1.0, 0.7843137254901961, 0.0}
\definecolor{fence}{rgb}{1.0, 0.47058823529411764, 0.19607843137254902}
\definecolor{vegetation}{rgb}{0.0, 0.6862745098039216, 0.0}
\definecolor{trunk}{rgb}{0.5294117647058824, 0.23529411764705882, 0.0}
\definecolor{terrain}{rgb}{0.5882352941176471, 0.9411764705882353, 0.3137254901960784}
\definecolor{pole}{rgb}{1.0, 0.9411764705882353, 0.5882352941176471}
\definecolor{trafficsign}{rgb}{1.0, 0.0, 0.0}

\newcommand{\tikzcircle}[1]{\tikz[baseline=-0.7ex]\draw[#1,fill=#1,radius=1.4mm] (0,0) circle ;}%
\newcommand\semcolor[1][black]{\tikzcircle{#1}}

\newcommand{\simplecolorbar}[6]
{   
        \foreach \x [count=\c] in {#3}{ \xdef\numcolo{\c}}
        \pgfmathsetmacro{\pieceheight}{#1/(\numcolo-1)}
        \xdef\lowcolo{}
        \foreach \x [count=\c] in {#3}
        {   \ifthenelse{\c = 1}
            {}
            {   \fill[bottom color=\lowcolo,top color=\x] (0,{(\c-2)*\pieceheight}) rectangle (#2,{(\c-1)*\pieceheight});
            }
            \xdef\lowcolo{\x}
        }
        \draw (0,0) rectangle (#2,#1);
        \pgfmathsetmacro{\secondlabel}{#4+#6}
        \pgfmathsetmacro{\lastlabel}{#5+0.01}
        \pgfkeys{/pgf/number format/.cd,fixed,precision=0}
        \foreach \x in {#4,\secondlabel,...,\lastlabel}
        { \draw (#2,{(\x-#4)/(#5-#4)*#1}) -- ++ (0.05,0) node[right] 
        {\footnotesize{\pgfmathprintnumber{\x}\si{\per\meter}}};
        }
}

\newcommand{\makescaleimage}[3]{%
\begin{tikzpicture}
      \node[anchor=south west,inner sep=0] (image) at (0,0) {\includegraphics[width=\linewidth]{#1}};
      \begin{scope}[x={(image.south east)},y={(image.north west)}]
        \fill [black] ({\scalemargin}, {\scalemarginh}) rectangle ++({\scalesection}, {\scaleheight});
        \fill [white] ({\scalemargin}, {\scalemarginh+\scaleheight}) rectangle ++({\scalesection}, {\scaleheight});
        \fill [white] ({\scalemargin+\scalesection}, {\scalemarginh}) rectangle ++({\scalesection}, {\scaleheight});
        \fill [black] ({\scalemargin+\scalesection}, {\scalemarginh+\scaleheight}) rectangle ++({\scalesection}, {\scaleheight});
        \fill [black] ({\scalemargin+2.0*\scalesection}, {\scalemarginh}) rectangle ++({\scalesection}, {\scaleheight});
        \fill [white] ({\scalemargin+2.0*\scalesection}, {\scalemarginh+\scaleheight}) rectangle ++({\scalesection}, {\scaleheight});
        \fill [white] ({\scalemargin+3.0*\scalesection}, {\scalemarginh}) rectangle ++({\scalesection}, {\scaleheight});
        \fill [black] ({\scalemargin+3.0*\scalesection}, {\scalemarginh+\scaleheight}) rectangle ++({\scalesection}, {\scaleheight});
        \draw ({\scalemargin}, {\scalemarginh+(2.0*\scaleheight)+0.04}) node {\footnotesize{0}};
        \draw ({\scalemargin+2.0*\scalesection}, {\scalemarginh+(2.0*\scaleheight)+0.04}) node {\footnotesize{#2}};
        \draw ({\scalemargin+(4.0*\scalesection)}, {\scalemarginh+(2.0*\scaleheight)+0.04}) node {\footnotesize{#3}};
        \draw ({\scalemargin+(4.0*\scalesection)+0.04}, {\scalemarginh+0.02}) node {\tiny{m}};
      \end{scope}
\end{tikzpicture}%
}

\newcommand{\clsrule}[3]{\multicolumn{#1}{r}{\rule[0.1ex]{0.2ex}{.5em}\rule[0.2pt]{#2}{.5pt}\hspace{1pt}#3\hspace{1pt}\rule[0.2pt]{#2}{.5pt}\rule[0.1ex]{0.2ex}{.5em}}}

\newcommand{\bc}{\mathbf{c}}

\newcommand{\bff}{\mathbf{f}}

\newcommand{\bo}{\mathbf{o}}
\newcommand{\bp}{\mathbf{p}}\newcommand{\bP}{\mathbf{P}}

\newcommand{\bx}{\mathbf{x}}
\newcommand{\by}{\mathbf{y}}
\newcommand{\bz}{\mathbf{z}}

\newcommand{\nR}{\mathbb{R}}

\newcommand{\cV}{\mathcal{V}}
\newcommand{\cP}{\mathcal{P}}
\newcommand{\cL}{\mathcal{L}}
\newcommand{\figref}[1]{Fig.~\ref{#1}}
\newcommand{\secref}[1]{Section~\ref{#1}}
\newcommand{\eqnref}[1]{Eq.~\eqref{#1}}
\newcommand{\tabref}[1]{Table~\ref{#1}}

\makeatletter
\def\onedot{.}
\def\eg{e.g\onedot} 
 
\def\wrt{wrt\onedot}
\def\etal{et~al\onedot}
\makeatother

\newcommand{\boldparagraph}[1]{\vspace{0.2cm}\noindent{\bf #1:} }
\definecolor{darkgreen}{rgb}{0,0.7,0}

\begin{document}



\begin{acronym}
\acro{SDF}[SDF]{signed distance function}
\acroplural{SDF}[SDFs]{signed distance functions}

\acro{TSDF}[TSDF]{truncated signed distance function}
\acroplural{TSDF}[TSDFs]{truncated signed distance functions}

\acro{DNN}[DNN]{deep neural network}

\acro{CNN}[CNN]{convolutional neural network}
\acroplural{CNN}[CNNs]{convolutional neural networks}
\acro{ADNN}[ADNN]{alternating direction neural network}
\acroplural{ADNN}[ADNNs]{alternating direction neural networks}

\acro{JSD}[JSD]{Jensen-Shannon divergence}

\acro{MISE}[MISE]{multiresolution IsoSurface Extraction}
\acro{MLP}[MLP]{Multi-layer perceptron}


\acro{mIoU}[mIoU]{mean intersection-over-union}
\acro{IoU}[IoU]{intersection-over-union}

\acro{bev}[BEV]{bird's-eye view}

\acro{cbn}[CBN]{conditioned batch normalization}

\acro{DIF}[DIF]{Deep Implicit Function}
\acroplural{DIF}[DIFs]{Deep Implicit Functions}

\acro{TS3D}{Two-Stream}

\acro{TTA}[TTA]{test-time augmentation}

\end{acronym}

%

%
\title{Semantic Scene Completion using\\ Local Deep Implicit Functions on LiDAR Data}
%
%
%

\author{Christoph~B.~Rist, David~Emmerichs, Markus~Enzweiler and Dariu~M.~Gavrila
\IEEEcompsocitemizethanks{%
\IEEEcompsocthanksitem C. Rist is with the Intelligent Vehicles Group, TU Delft (NL) and with Mercedes-Benz AG, Stuttgart (DE).
\IEEEcompsocthanksitem D. Emmerichs is with Mercedes-Benz AG, Stuttgart (DE). %
\IEEEcompsocthanksitem M. Enzweiler is with Esslingen University of Applied Sciences (DE).
\IEEEcompsocthanksitem D. Gavrila is with the Intelligent Vehicles Group, TU Delft (NL).}
}

\IEEEtitleabstractindextext{%
\begin{abstract}

Semantic scene completion is the task of jointly estimating 3D geometry
and semantics of objects and surfaces within a given extent.
This is a particularly challenging task on real-world data
that is sparse and occluded.
We propose a scene segmentation network based on local \aclp{DIF}
as a novel learning-based method for scene completion.
Unlike previous work on scene completion, our method produces a continuous scene
representation that is not based on voxelization.
We encode raw point clouds into a latent space locally and at multiple spatial resolutions. A global scene completion function is subsequently assembled from the localized function patches.
We show that this continuous representation is suitable to encode
geometric and semantic properties
of extensive outdoor scenes without the need for spatial discretization
(thus avoiding the trade-off between level of scene detail and the scene extent that can be covered). 

We train and evaluate our method on
semantically annotated LiDAR scans from the Semantic KITTI dataset.
Our experiments verify that our method generates a powerful representation
that can be decoded into a dense 3D description of a given scene.
The performance of our method surpasses the state of the art
on the Semantic KITTI Scene Completion Benchmark in terms of
geometric completion \acf{IoU}.
\end{abstract}

\begin{IEEEkeywords}
LiDAR, semantic scene completion, semantic segmentation, geometry representation,
deep implicit functions
\end{IEEEkeywords}%
}

\maketitle

\IEEEdisplaynontitleabstractindextext

%
\IEEEpeerreviewmaketitle


\IEEEraisesectionheading{\section{Introduction}\label{sec:introduction}}
%
%
%
%

\IEEEPARstart{A}{utonomous} mobile robots have to base their
actions almost exclusively on an internal representation of their current environment.
Perception systems are built to create and update
such a representation from real-time raw sensor data.
We are interested in a model of the current environment that preferably
condenses the information that is important for the task at hand or makes
it easy to extract relevant information.
For robot navigation it is required to estimate whether a certain area is occupied by an object
and what semantic meaning different objects and surfaces hold.
Even non-mobile settings, \eg{} mapping applications, benefit from an effective
geometric and semantic completion of low-resolution or incomplete sensor data.
To fulfill this need 3D completion aims to map and infer the true geometry of objects from sensor input.
Semantic scene completion extends this task to larger arrangements of multiple
objects and requires to predict the corresponding semantic classes.


\begin{figure}[t]
  \centering
  \includegraphics[width=\linewidth]{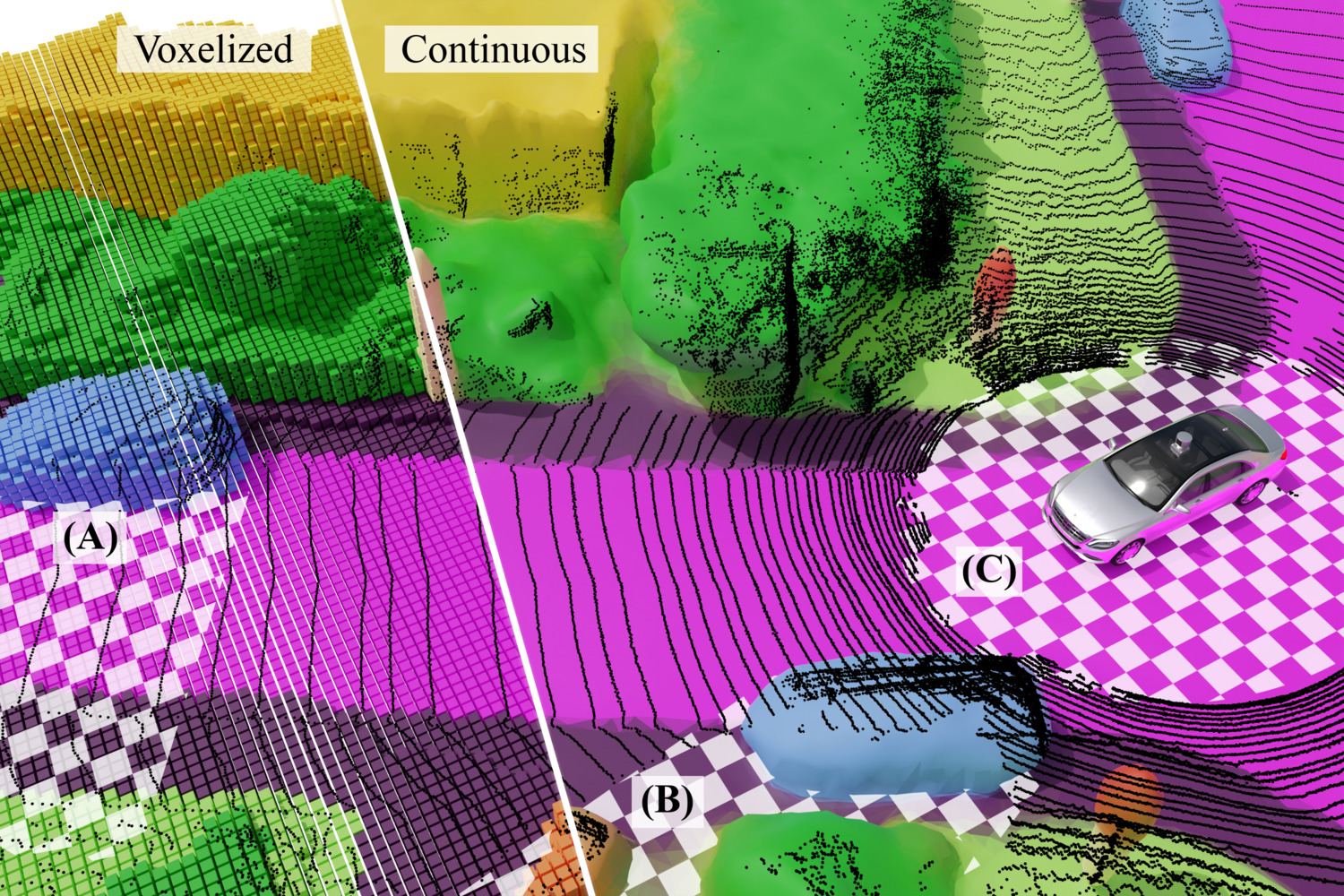}
  \caption{Illustration of the semantic scene completion task and the output of our method.
  Sensors are limited in their resolution and restricted to a single perspective of their
  surroundings.
  A LiDAR scan (herein depicted as black points) is
  characterized by a varying
  degree of sparsity caused either by distance (A), occlusions from
  objects (B) or sensor blind spots (C).
  Our method is able to complete the sparse scan geometrically and semantically
  and can be applied to large spatial extents as typically found in outdoor environments.
  The underlying representation is not tied to a fixed output resolution
  and describes the scene using a continuous function
  (right side, color indicates semantic class).
  Therefore the geometry does not exhibit quantization artefacts
  resulting from a discretization into voxels (left side).
  }
  \label{fig:cover_image}
\end{figure}

Sensor data can only reflect partial observations of the real world.
First, this is because of the physical properties of the sensors themselves
which impose limits on their ultimate resolution, frequency,
and minimal amount of noise with which they capture data.
Second, it is because every sensor is restricted to its current perspective.
Thus, after the point-of-view sensor data is mapped into the 3D scene, the result
will always be characterized by a distance-decreasing sampling density,
occlusions and blind spots
(see regions marked A, B, C in \figref{fig:cover_image} respectively).
Multiple sensors mounted on a single vehicle do not alleviate that issue significantly. They are usually positioned rather close together,
so that their view of the surroundings still exhibits almost the same degree of
occlusions and shadows.
Hence, the completion task in 3D Euclidean space represents
a key challenge for perception in real-time cognitive robotics:
Making predictions about currently unobserved areas by the use of context and experience.
This ability is only necessary for real-time perception systems.
In a static world without time constraints it would be possible to just move
the sensors towards areas of interest to gain evidence of their true appearance.
But unlike static worlds, mobile robots need to reason about the nature
of objects given only the current observations.

The semantic scene completion task is based on
a correlation between the semantic class
of an object or surface and its physical 3D geometry.
In the case of LiDAR, the sensor observes a part of the scene's geometry.
The semantics that can be deduced from this geometry can be used to
then again complete the missing geometry.
Regardless of the dataset in use, hidden geometry can only be completed by
means of what is probable but never with absolute certainty.
This probability is in turn associated with the type of objects within the scene.
Naturally, human perception exhibits the same inherent limitations as computer sensors
when it comes to physical limitations and the laws of 3D geometry.
However, humans make up for this by fitting a powerful model to infer even large
missing pieces of geometric and semantic information about their surroundings.

Our approach is a deep learning method that we train on a large number of
semantically annotated LiDAR measurements.
The model leverages the training data as prior knowledge to reason
about the geometry and semantics of the complete 3D scene from a single LiDAR
scan as input.
We propose to represent the scene completion output with localized \acp{DIF}.
A \ac{DIF} is a continuous function over 3D space which
classifies individual positions.
The composed scene completion function
$\bff^\bc_{\text{\nn{}}}:\nR{}^3\rightarrow[0,1]^{N+1}$
is defined over all scene positions and outputs a classification vector
over $N$ semantic classes and free space.
This continuous representation
avoids a trade-off between achievable spatial output resolution
and the extent of the 3D scene that can be processed.
\figref{fig:cover_image} presents a visualization of the resulting function and a comparison
to a voxelized output.

When it comes to the representation of geometry,
existing works on object or scene completion focus most commonly on voxelization
\cite{Dai2018CVPR,Firman2016CVPR,Garbade2019CVPRWorkshops,Song2017CVPR,Stutz2018CVPR,Behley2019ICCV,Roldao2020,yan2021sparse,Cheng2020}.
However, this results in satisfactory output resolutions only for volumes of limited extent.
Approaches using \acp{DIF} to represent shapes
\cite{Chen2019CVPR,Mescheder2019CVPR,michalkiewicz2019deep,Park2019CVPR}
only encode single objects into a fixed size latent vector.
Most previous work completes 3D geometry on the assumption that the scene in question
is covered evenly with sensor measurements,
such as indoor scenes recorded with RGB-D cameras.
In comparison, the density of a LiDAR scan decreases steadily with
distance so that gaps between measurements get larger.
Distance to the sensor and occlusions lead to areas
where the actual ground truth geometry cannot be inferred anymore from
the measurements.
This label noise and the varying sparsity is a challenge for current models \cite{Behley2019ICCV}.

Our method requires accurate 3D measurements of a scene to be trained
for geometric completion.
These measurements can be obtained from one or multiple LiDAR sensors,
or a LiDAR sensor that is moved through the scene,
provided that all measurements can be transformed into a single
reference coordinate system.
If semantic annotations are not available our method can still
be trained for pure completion of scene geometry.

This paper builds upon our earlier work on LiDAR-based scene segmentation \cite{Rist2020IV}.
For this work, we created a training procedure for semantic scene completion based on accumulated LiDAR data and conducted an extensive experimental evaluation of our design choices and parameters.
%
In summary, our contributions are:
\begin{itemize}[leftmargin=4mm]
    \item We produce a representation for both geometry and semantics of 3D scenes by \aclp{DIF} with spatial support derived from a 2D multi-resolution grid. Our combination with continuous output coordinates make dense decoding of large spatial extents feasible.
    \item We generate point-like training targets from time-accumulated real-world LiDAR data and the included free space information. Dynamic objects are considered separately to ensure consistency.
    \item In experiments on the Semantic~KITTI Scene completion benchmark, we show that the proposed approaches outperform voxel-based methods on geometric completion accuracy.
\end{itemize}


\section{Related Work}

First, this section discusses ways to represent geometry and surfaces within
the context of reconstruction algorithms.
Second, related work about geometric completion is categorized into completion
of single object shapes and completion of indoor scenes from synthetic or RGB\=/D data.
Finally, we take a look at the state of the art in semantic segmentation
and scene completion of outdoor scenes from real-world LiDAR data.

\subsection{Geometry and Surface Representation}
Most commonly the output representation for 3D scene completion
is a voxel occupancy grid \cite{Song2017CVPR}, voxelized (truncated) \acp{SDF} \cite{Stutz2018CVPR,Dai2018CVPR,Firman2016CVPR,Garbade2019CVPRWorkshops,Dai_2020_CVPR},
or interpolation and CRFs \cite{Tchapmi2017THREEDV} for sub-voxel accuracy.
A differentiable deep marching cubes algorithm replaces
the \ac{SDF} as an intermediate representation and enables to
train the surface representation end-to-end \cite{Liao2018CVPR}
but the resulting representation is still constrained to the underlying
voxel resolution.
The general trade-off between output resolution and
computational resources is an issue for 3D representations \cite{Dai2018CVPR}.
Octree-based \acp{CNN} have been proposed to represent space at different resolutions and to perform gradual shape refinements~\cite{Haene2017THREEDV,Riegler2017CVPR,Tatarchenko2017ICCV,Wang2017OCNN}.

Recent works represent 3D shapes and surfaces implicitly
as isosurfaces of an output function which classifies single points in Euclidean 3D space~\cite{Chen2019CVPR,Mescheder2019CVPR,michalkiewicz2019deep,Park2019CVPR}.
Depending on the output function's complexity this approach has the capacity and expressiveness to represent fine geometric details.
An encoder creates a parameter vector that makes the output function
dependent on the actual input data for geometric reconstruction.
Both the output function and encoder are represented as \acp{DNN} and trained by backpropagation.
They either use oriented surfaces \cite{michalkiewicz2019deep} or  watertight meshes \cite{Mescheder2019CVPR} from ShapeNet \cite{Chang2015ShapeNetAI} as synthetic full-supervision training targets.
These methods have improved the state of the art significantly for shape reconstruction
and completion. However, their scope is limited to the reconstruction of single objects.
These approaches do not generalize or scale well because of the nature
of a single fixed-size feature vector that represents a shape globally.

Recently, \acp{DIF} are combined with grid structures or other support positions that improve their
spatial capabilities to describe larger scene extents \cite{Jiang_2020_CVPR,Peng2020ECCV} or
more complex geometric details of individual objects \cite{Chibane_2020_CVPR,Genova_2020_CVPR,Peng2020ECCV} instead of only simple shapes.

To represent more complex details in 3D shapes, a set of local analytic 3D functions with limited support can be combined with deep implicit functions to predict occupancy
\cite{Genova_2020_CVPR}.
The latent representations of individual small synthetic object parts can be used to assemble a large 3D scene~\cite{Jiang_2020_CVPR}.
For this purpose, synthetic objects are first auto-encoded to generate the latent space. Then, a possible representation of a scene is found by iterative inference. This setup only requires a decoder from latent grid to the 3D scene.
Concurrent to our work, \cite{Peng2020ECCV,Chibane_2020_CVPR} encode 3D points into a 2D grid or 3D feature volume and perform bilinear or trilinear interpolation on this feature space.
Here \cite{Chibane_2020_CVPR} explicitly considers features from multiple resolutions and the query position in only used for interpolation, not in the decoder.
\cite{Peng2020ECCV} uses the query position for interpolation and again as concatenation to the latent feature in the decoder. The feature grid is single-resolution. For geometric reconstruction of indoor RGB-D data, the full volumetric grid performs best.
With a focus on representation and reconstruction of geometry, the method is trained on synthetic watertight-meshes and uniformly sampled point clouds are used as input.


\subsection{Shape Completion}
Poisson Surface Reconstruction is a state-of-the-art reconstruction algorithm for an object's surface from
measured oriented points \cite{10.5555/1281957.1281965}.
As with other implicit representations the resulting geometry needs to be extracted by
marching cubes or an iterative octree variant of marching cubes \cite{Mescheder2019CVPR}.
Poisson Surface Reconstruction handles noise and imperfect data well and
adapts to different local sampling densities.
However, it is of limited use on real-time real world data as it is unable
to leverage prior knowledge to complete unseen or sparse regions
unlike methods based on learned shape representations.

Many data-driven, learning-based and symmetry-based approaches have been proposed
for shape completion. We refer to Stutz~\etal{}~\cite{Stutz2018ARXIV} for an
overview and focus on shape completion on LiDAR scans.
3D models can be used to train a \ac{DNN} for the shape completion problem
on synthetic data and perform inference on real LiDAR scans \cite{Yuan2018ThreeDV}.
Alternatively, a shape prior from synthetic data can be used
for amortized maximum likelihood inference to avoid the domain gap
between synthetic and real data \cite{Stutz2018ARXIV}.
Recently, it has been shown that synthetic data can be avoided altogether
by using a multi-view consistency constraint
to train shape completion only from LiDAR scans without full supervision \cite{Gu2020ECCV}.

\subsection{Semantic Scene Completion}

For a recent comprehensive survey on semantic scene completion we refer to \cite{Roldao2021Arxiv}.
The subject of scene completion has first gotten momentum from the
wide availability of RGB\=/D cameras
leading to the advent of indoor semantic segmentation datasets
such as the NYUv2 Depth Dataset \cite{Silberman2012ECCV} and ScanNet~\cite{dai2017scannet}.
\cite{Firman2016CVPR}~is a pioneering work to infer full scene geometry
from a single depth image
in an output space of voxelized \acp{SDF}.
Generalization to entirely new shapes is data-driven and implemented with
voxel occupancy predicted by a structured random forest.
A specially created table-top scene dataset
with ground truth from a Kinect RGB\=/D camera is used as
full-supervision training target.

A volumetric occupancy grid with semantic
information can be predicted from voxelized \acp{SDF} as input in an end-to-end manner~\cite{Dai2018CVPR,Song2017CVPR}.
They apply their methods to synthetic indoor data from the SUNCG dataset.
While \cite{Song2017CVPR} is appropriate only on single RGB\=/D images,
\cite{Dai2018CVPR} extends to larger spatial extents.
Multiple measures improve geometric precision and consistency:
Using \acp{SDF} as output representation per voxel,
an iterative increase of voxel resolution, and the division of space into interleaving
voxel groups.

Voxelized \acp{SDF} and semantic segmentation can be inferred
by explicit fusion of single depth images with RGB data \cite{Garbade2019CVPRWorkshops}.
\cite{Jiang_2020_CVPR} validates the geometric representation power of \acp{DIF} in 
combination with a structured latent space approach on indoor RGB\=/D data
of the Matterport3D dataset \cite{Matterport3D}.
The details in the completion of RGB\=/D scans from Matterport3D
can be improved by progressive spatial upsampling in the decoder and a deliberate loss formulation that does not penalize unseen areas \cite{Dai_2020_CVPR}.

\subsection{Segmentation and Scene Completion on LiDAR Data}
Numerous prior works focus on semantic segmentation of all observed data points
resulting in a pixel-wise or point-wise classification of LiDAR data.
These methods do not predict any labels for invisible parts of space from the sensor's perspective.
However, datasets and benchmarks on real-world road scenes have
defined a standard of semantic classes that is significant
while simultaneously advancing the state of the art~\cite{Alhaija2018IJCV,Behley2019ICCV,Cordts2016Cityscapes}.
CNN-architectures on RGB-Images for segmentation and detection
\cite{Chen2018ECCV,redmon2018arxiv} have inspired sensor-view based approaches
in the more recent LiDAR-based segmentation task \cite{Tatarchenko2018CVPR,Wu2018ICRA,milioto2019iros}.
Neural network architectures adjust to the three dimensional nature of a segmentation or detection problem through voxelization of input data \cite{Lang2019CVPR,Rist2019IV,Zhou2018CVPR,liu2019pvcnn},
combination with sensor-view range images \cite{gerdzhev2020tornadonet}, and use of surface geometry \cite{Tatarchenko2018CVPR}.
Computation, memory efficiency and representation of details 
of voxel architectures can be improved
by combining a coarser voxel structure with a point-feature branch
for details \cite{liu2019pvcnn} and neural architecture search~\cite{tang2020searching}.

The scene completion problem on real-world data has only recently been advanced by the large-scale Semantic KITTI dataset \cite{Behley2019ICCV} featuring point-wise semantic annotations on LiDAR together with a private test set and a segmentation benchmark for
semantic scene completion.
Methods originally applied to scene completion from depth images \cite{Song2017CVPR, Garbade2019CVPRWorkshops}
can be adapted for LiDAR scene completion:
The Semantic KITTI authors \cite{Behley2019ICCV} adapt the
\ac{TS3D} approach \cite{Garbade2019CVPRWorkshops} which 
is originally applied to depth images of indoor scenes of the NYUv2 dataset.
\ac{TS3D} combines geometric information from a depth image and
a predicted semantic segmentation from an RGB image in a volumetric voxel grid.
For Semantic KITTI outdoor scenes, they use a state-of-the-art
DeepNet53 segmentation network trained on Cityscapes \cite{Cordts2016Cityscapes}
and SatNet \cite{Liu2018NIPS} for voxel output.

The three recent methods LMSCNet \cite{Roldao2020}, JS3CNet \cite{yan2021sparse}, and S3CNet \cite{Cheng2020} only use LiDAR data as input.
The usage of U-net architectures for down-, upsampling, and spatial context is a common architectural pattern.
LMSCNet~\cite{Roldao2020} operates on the voxelized LiDAR input and uses a 2D-\ac{CNN} backbone for feature extraction.
The voxelized output is inferred with a monolithic hybrid-network that predicts the completion end-to-end.
LMSCNet can output a lower-resolution coarse version of a scene at an intermediate stage.
However, their experiments show that the single-output version
trained only on the highest resolution performs slightly better than
the multi-scale version trained with multiple-resolution losses.

S3CNet \cite{Cheng2020} and JS3CNet \cite{yan2021sparse} both use the raw LiDAR scan as input.
Both also propose to use a lower resolution scene representation internally
which is subsequently upsampled into the full output voxel resolution.
JS3CNet proposes a two-stage approach:
First, a semantic segmentation of the input LiDAR scan is inferred.
Second, a neural network fuses the voxelized semantic segmentation and point-wise feature
vectors into the voxelized representation of the completed scene.
S3CNet augments the input LiDAR scan with a calculation of normal surface vectors
from the depth-completed range image and TSDF values. These are stored in a sparse tensor.
A semantic 2D BEV map and a 3D semantic sparse tensor are predicted in parallel.
These are then subsequently fused into a full 3D tensor. The final scene completion is
obtained after a second semantically-based post-processing.
The authors conduct ablations and attribute a large share of the final results to the post-processing.

\section{Proposed Approach}

\subsection{Overview}
\begin{figure*}[t]
  \centering
  \includegraphics[width=\linewidth]{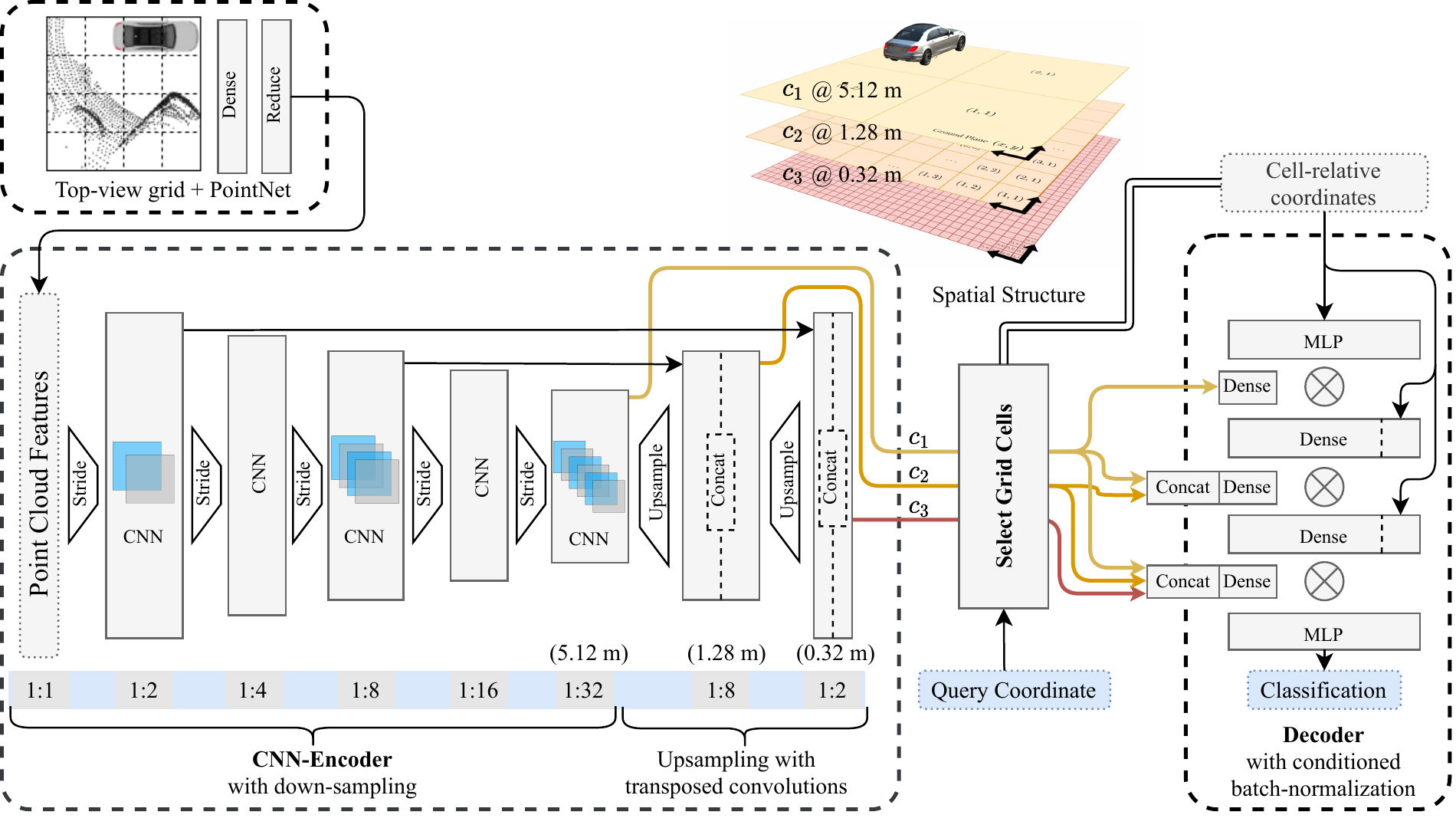}%
  \caption{%
  \textbf{Network architecture:} %
  The feature extractor creates a top-view feature map of the input point cloud.
  The CNN-encoder outputs feature maps at three different resolutions that make
  up the latent representation of the 3D scene.
  The decoder classifies individual coordinates within the 3D scene extent.
  Latent feature vectors and relative-coordinates are processed by conditioned
  batch normalization in the decoder.
  \label{fig:architecture_v2}%
  }
\end{figure*}

Our method takes as input a LiDAR scan and outputs
the corresponding scene completion function
$\bff{}^\bc_{\text{\nn}}:\nR{}^3\rightarrow{}[0,1]^{N+1}$.
This function maps every 3D position $\bp$ within
the scene to a probability vector
that we define to represent the semantic class of the position $\bp$.
The dependence of the completion function $\bff{}^\bc_{\text{\nn}}$ on the input
data is expressed by the superscript vector $\bc$.
Positions belonging to objects in the scene are
categorized into $N$ semantic classes.
The additional class \textit{free space} represents positions that are not
occupied by any object (instead they are occupied by air).
The resulting total of $N+1$ classes is able to describe every position within the scene.
Hence the $\bff{}^\bc_{\text{\nn}}$ function uniformly represents
the geometric and semantic segmentation of space
instead of only the physical boundaries of objects.
The global $\bff{}^\bc_{\text{\nn}}$ function is built from many
local functions $\bff{}_{\text{L}}$.
Every local function has two distinctive inputs:
The coordinate of interest $\Delta\bp$ and a parameterization vector $\bc_V$.
In the context of \acp{DIF}, producing an output function $\bff{}^\bc_{\text{L}}$ means
generating a parameterization (\textit{conditioning}) vector $\bc_V$.
When the parameterization vector $\bc_V$ is fixed we obtain
the conditioned function $\bff{}^\bc_{\text{L}}$ which is only dependent on the remaining input coordinate $\Delta\bp$.
Our approach to the composition of the $\bff{}^\bc_{\text{\nn}}$ function
is designed to encode large outdoor scenes.
While related works on single object shape representation
encode geometry information in a fixed size conditioning vector,
we add spatial structure to the latent space through the use of a 2D feature grid.
Each grid entry is a conditioning vector for a local function.
The grid is chosen to be two-dimensional, uniform and represents the
$xy$-coordinates of a flattened scene that omits the vertical dimension.
We use three grids, each with its own feature resolution.
An illustration is given in \figref{fig:architecture_v2}.
As a consequence of the grid approach, the amount of conditioning information is tied
to the spatial extent of the scene.
The intuition is that each individual conditioning vector now describes only a
small part of the complete scene in the vicinity of its own position.
Each grid entry always encodes a volume of the same size,
regardless of the overall scene extent.

We propose a convolutional encoder to generate the feature maps that make up
the conditioning grid.
Outdoor scenes are mainly composed from objects at different locations
on the ground plane ($xy$).
Therefore the configuration of outdoor scenes is assumed to be translation-invariant
in $x$ and $y$ direction.
Intuitively, the encoding of the front of a car or a part of a tree can be the same
regardless of the absolute position of the object within the scene.
For this reason we consider the implementation of the encoder as
a convolutional neural network as appropriate.
\figref{fig:architecture_v2} gives a schematic overview over
the point cloud encoding stage, feature selection,
and decoding a position into a coordinate classification.

The next section describes the details of the composition of the
global completion function $\bff{}^\bc_{\text{\nn}}$
from multiple conditioning vectors and grid resolutions.
A sampling-based supervised training method from real-world LiDAR data is proposed
and details on the used network architecture and inference procedure follow.

\subsection{Spatial Structure of Latent Feature Grid} %
\boldparagraph{Composition of $\bff^\bc_{\text{\nn{}}}$}%
Centerpiece of our method is the formulation of latent conditioning vectors that are
spatially arranged in a grid
and generated by a convolutional encoder network on LiDAR point clouds.
Each individual conditioning vector $\bc_V$ parameterizes a
\emph{local segmentation function
$\bff^{\bc}_{\text{L}}(\Delta\bp_V)$}
to classify a position of interest $\bp$.
Even though the domain of individual
local functions is $\nR^3$ and therefore infinite,
the classification will only be meaningful for positions that are close to
the conditioning vector's position within the scene.

It is necessary to define how a conditioning vector is selected
for a given query coordinate $\bp$.
It is straightforward to use the single vector of the grid cell that
contains the coordinate $\bp$ when projected onto the ground plane.
But with this approach the resulting global function would exhibit
discontinuities between grid cells.
Instead, we select the four grid cells with 
the closest center coordinates for the query coordinate $\bp$.
Thus we obtain four individual classifications for $\bp$ and perform
bilinear interpolation according to $\bp$'s position within the square of the
surrounding grid cell center points.
We denote the set of the four closest conditioning cells 
the \textit{support region} $\cV_\bp$ of the coordinate $\bp$ and
the corresponding coefficients for bilinear interpolation $w$.
This yields the global classification function
\begin{align}
    \bff^\bc_{\text{\nn{}}}(\bp) \label{eq:prediction}&= \sum_{V \in \cV_\bp} w(\Delta\bp_V) \bff_{\text{L}}(\bc_V{}, \Delta\bp_V) \\
    \label{eqn:local}\text{with}\quad \Delta\bp_V &= \bp - \bo_V
\end{align}
for a coordinate $\bp$. $\bo_V$ is the center position of a cell $V$ and $\bc_V$ is the
conditioning vector at cell $V$.
The coefficients for bilinear weighing $w(\Delta\bp_V)$ sum to $1$.
Intuitively, the spatial extent of a scene can be thought of as covered by overlapping function patches $\bff_{\text{L}}$.
Each function $\bff_{\text{L}}$ has its own coordinate origin $\bo_V$ at the center of its grid cell $V$.
\eqnref{eqn:local} conveys the translation of scene coordinates $\bp$
into the coordinate system of the conditioning vector's grid cell that shall
describe $\bp$.

\boldparagraph{Multi-resolution scene representation}%
An important aspect of the composition of $\bff^\bc_{\text{\nn{}}}$ is the
use of three individual conditioning vectors from three different resolutions levels.
The intuition behind this is that the geometric structure of a
scene is composed of different levels of detail.
There is the coarse positioning of the ground level and large structures
as well as more fine-grained details like curbstones, small objects and poles.
We reproduce this range in the network structure
to facilitate learning of a smooth representation with more details and
more consistency over cell boundaries.
The conditioning information for a single local function $f_L$ is composed
from three resolution-specific feature vecors.
We opt for features $\bc_V\!=\!(\bc_1,\bc_2,\bc_3)$ from the resolution ratios
$1\!\!:\!\!16$, $1\!\!:\!\!4$, and $1\!\!:\!\!1$ that originate
from a U-net-structured~\cite{long2015cvpr} convolutional feature encoder, as illustrated in \figref{fig:architecture_v2}.
The resolution ratios correspond to grid cells with \SI{5.12}{\meter},
\SI{1.28}{\meter}, and \SI{0.32}{\meter} edge length respectively.

For a scene position $\bp$, we select the four closest feature vectors
at the highest resolution feature map as support region.
The single cell where the coordinate resides
in is selected in each case of the two lower resolutions.
This $2\!\times\!2$ square of support positions at the highest
resolution is used for bi-linear interpolation.
Hence there are four local segmentation functions that are able to describe
the single position $\bp$ in the scene.
All four need to be evaluated to obtain the final interpolated classification result.

Each conditioning vector $\bc_i,i\in\{1,2,3\}$ belongs to a grid cell $V_i$ at resolution $i$ defining a coordinate system relative to its own position through its origin $\bo_{V_i}$.
Due to the hierarchical set of vectors $(\bc_1,\bc_2,\bc_3)$ at different resolutions,
we also obtain a corresponding 3-tuple of relative coordinates
$\Delta\bp_V=(\bp_1,\bp_2,\bp_3)$ with $\bp_i=\bp - \bo_{V_i}$
as input for $\bff_{\text{L}}$.

\subsection{Training on LiDAR Point Clouds}
\begin{figure}[tb]
    \centering{}
    \renewcommand{\arraystretch}{.5} 
    \setlength{\tabcolsep}{1pt} 
    
    \begin{tabularx}{\linewidth}{@{}XX@{}}
        \includegraphics[width=\linewidth]%
        {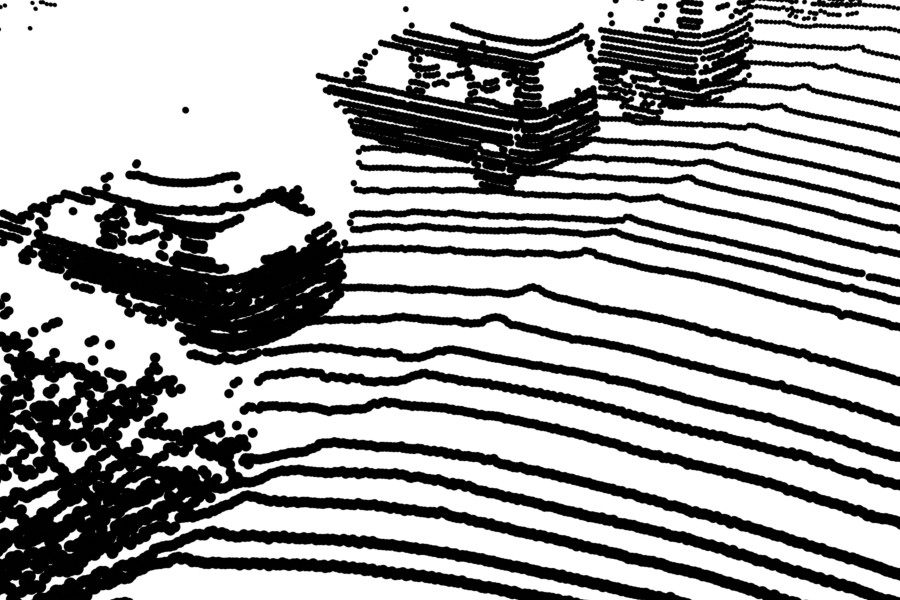} &
        \includegraphics[width=\linewidth]%
        {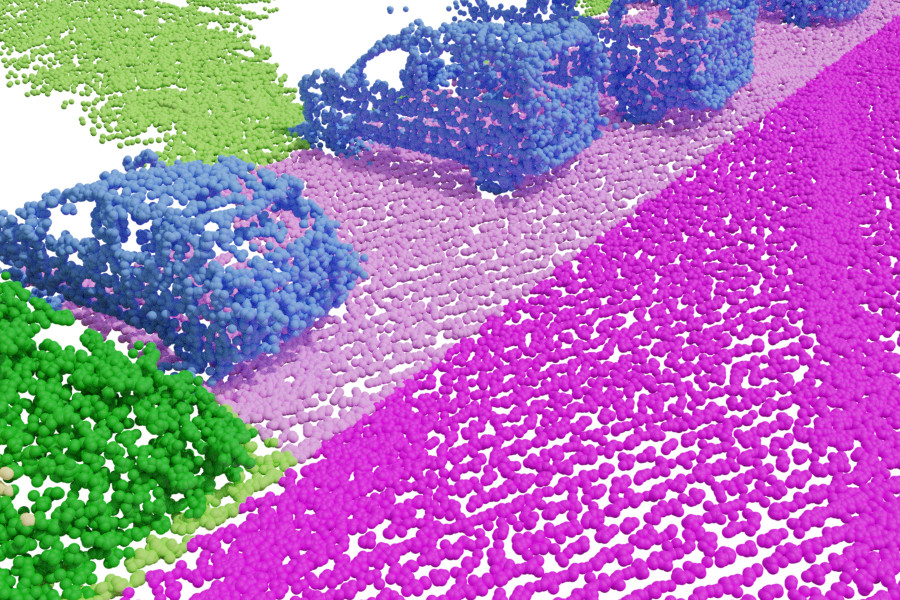} \\
        \includegraphics[width=\linewidth]%
        {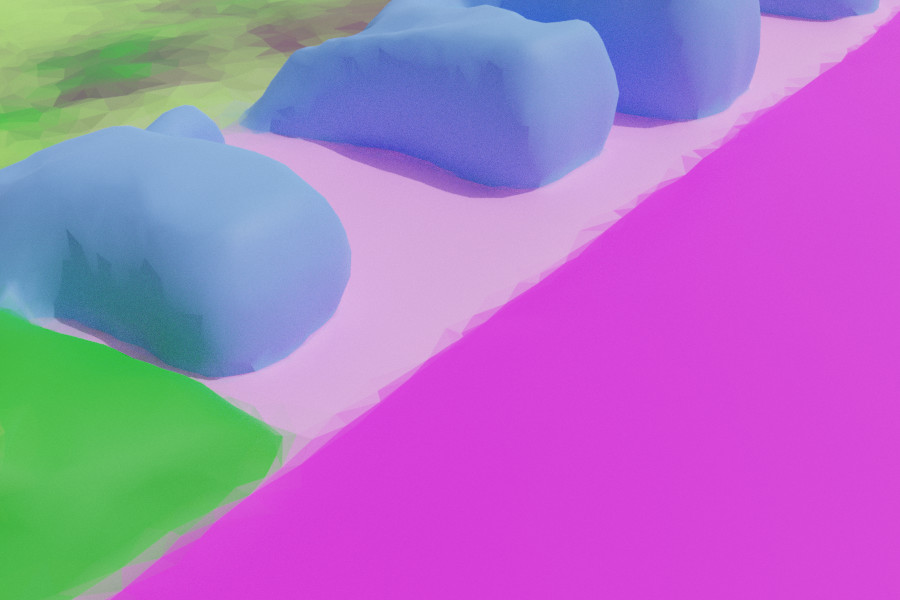} &
        \includegraphics[width=\linewidth]%
        {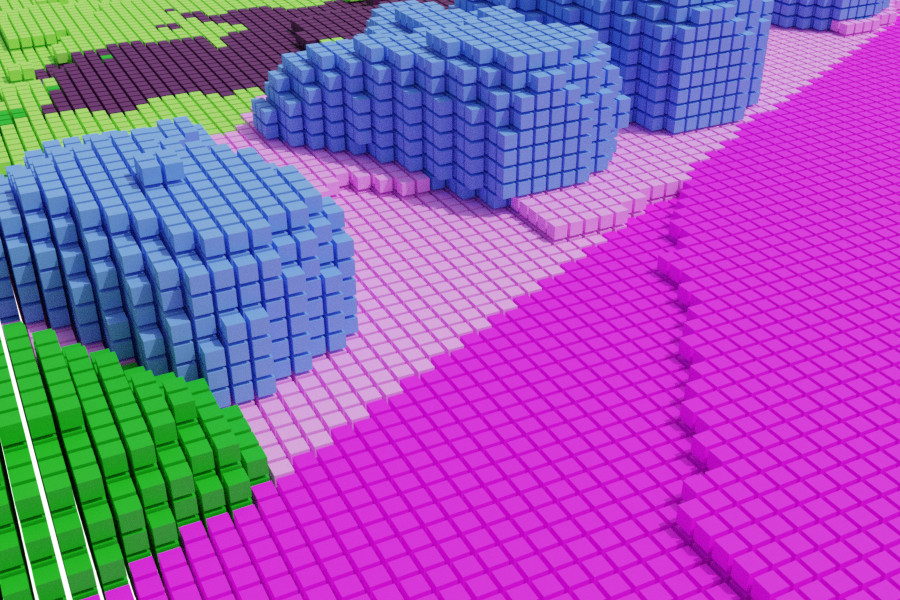} \\
    \end{tabularx}
    \caption{%
    Left to right, top to bottom: Input points, ground truth accumulated points, mesh visualization of
    continuous output function, derived voxelization at \SI{20}{\centi\meter} edge length.
    Geometric details can be represented more accurately by our continuous output function
    as compared to the voxelization resolution of the Semantic KITTI dataset.
    Our method does not cause artefacts
    on slanted surfaces (\eg{} road plane) or edges between objects. 
    \label{fig:voxelization}%
    }

\end{figure}

\boldparagraph{Sampling targets for supervised training} %
The decoder neural network and feature encoder are trained
end-to-end
using individual coordinates within the scene and their associated training labels.
This set of coordinate-label tuples is generated from different data sources.
The large number of time-accumulated LiDAR measurements is used as 
primary training target. Each LiDAR point has a position in the reference coordinate
frame and an associated semantic label.
Together, these positions make up all training targets for the occupied classes.
The top row of \figref{fig:voxelization} shows the single input point cloud
and the accumulated training targets with semantic annotations.

Next, we need to obtain positions that are of the free space class, so not
occupied by any object.
The pre-processing that accumulates LiDAR points keeps track of all voxels
that are observed at least once, but empty.
In every such empty voxel we sample a free space position target uniformly at random.
This ensures that the scene extent is evenly covered with free space information.

We use the input point cloud as a second source of free space positions.
The straight line between a LiDAR measurement
and the sensor's position at time of measurement is empty, meaning not occupied by any object.
We exploit this reality for self-supervised training of object geometry.
The goal of our scene completion function is to resemble physical boundaries.
Wherever surfaces are scanned by the LiDAR sensor we would like to have a sharp
transition of the completion function from the prediction of an occupied class
to a free space prediction.
Therefore we sample free space positions on the straight lines between
LiDAR measurement and sensor position.
We use an exponential decaying probability distribution to sample the free
space positions close to the surfaces of objects.
The approach of close surface sampling of free space targets
and the combination of surface sampled and global training positions is similar to \cite{Genova_2020_CVPR}.

\boldparagraph{Loss function} %
Training the classifier involves three separate loss terms: semantic $L_{S}$,
geometric $L_{G}$, and consistency $L_{C}$ loss.
Semantics and geometry could also be covered by a single
cross-entropy classification problem.
However, 
the formulation with individual losses allows to include positions that are known to be
occupied by an object without information about an object class,
\eg{} unlabeled LiDAR points.
Moreover, geometric and semantic loss terms can be weighted more easily against each other.
The overall loss
\begin{equation}
  L = \lambda_S \sum_{\cP}L_{S}
  + \lambda_{G} \sum_{\cP}L_{G}
  + \lambda_C \sum_{\cP}L_{C}    
\end{equation}
is the weighted sum of the individual losses that are each in turn summed over
all training targets.
We write the predicted probability vector at position $\bp$ as $[f_1,\dots,f_N,f_{N+1}]^{\intercal} = \bff^\bc_{\text{\nn{}}}(\bp)$. The scalar $f_{N+1}$ is the predicted probability of the free space class.

The semantic loss $L_S$ is a cross-entropy loss between the
classification output vector $[f_1,\dots,f_{N}]^{\intercal}$ and semantic ground truth.
The ground truth free space probability for LiDAR targets is always zero
as LiDAR measurements $\cL{}$ are assumed to be located on objects.
This loss is not evaluated for free space targets.

The geometric reconstruction loss
\begin{align}
\label{eq:reconstruction}
L_G &= H\left(\left[l_{\text{occupied}}, l_{\text{free}}\right]^{\intercal}, \left[\sum_{i=1}^{N}f_{i},f_{N+1}\right]^{\intercal}\right)
\end{align}
is the binary cross-entropy $H$ between
the sum of the semantic class probabilities $[f_1,\dots{},f_{N}]$ for all objects
and the remaining free space probability $f_{N+1}$.
It is available for all free space points with
$[l_{\text{occupied}}, l_{\text{free}}]^{\intercal} = [0, 1]^{\intercal}$
and all LiDAR points with
$[l_{\text{occupied}}, l_{\text{free}}]^{\intercal} = [1, 0]^{\intercal}$.

The consistency loss 
\begin{align}
    L_{C} & = \mathrm{JSD}\left(\bff_{\text{L},0}(\bp), \dots, \bff_{\text{L},m}(\bp)\right) \\
     & = \phantom{-{}} H\left(\frac1{m}\sum_{V \in \cV_\bp}\bff_{\text{L}}(\bc_V{}, \Delta\bp_V)\right)\nonumber\\
    & \phantom{={}} - \frac1{m} \sum_{V \in \cV_\bp} H\left(\bff^{\bc}_{\text{L}}(\Delta\bp_V)\right)\quad \label{eq:consistency}
\end{align}
for a given coordinate $\bp$ is the \ac{JSD}
between $m = |\cV_\bp|$ probability distributions
predicted by the 
local segmentation functions $\bff_{\text{L}}$ on
the support region $\cV_\bp$ of a consistency point $\bp$.
$\mathrm{H}(\bP{})$ denotes the entropy of distribution $\bP{}$.
The \ac{JSD} is symmetric and always bounded.
Multiple local functions $f_L$ make a prediction for the same position in the scene.
The unweighted output of these local functions $f_L$ exhibit grid artefacts between
neighboring cells.
The consistency loss acts as a regularizer by penalizing divergence between the grid cells
without the need to specify any particular
semantic or free space target label.
Thereby, this loss term is available at any position within the scene,
not only at regions where training targets from LiDAR points or sampled
free space targets are occurring.
We provide our numerically stable formulations of the geometric and consistency
loss terms in the appendix (\secref{sec:geometric_loss}, \secref{sec:consistency_loss}).

\subsection{Implementation}
All \ac{DNN} network details and the hyperparameters are listed in the appendix in \tabref{tab:network_architecture} and \tabref{tab:hyperparameters}.

\boldparagraph{LiDAR point cloud encoding} %
At the base layer we use a voxel-wise point cloud feature encoder from
recent literature \cite{Lang2019CVPR,Zhou2018CVPR}.
The encoder transforms the raw input point set
into a fixed-size \acl{bev} feature representation (\figref{fig:architecture_v2}, top-left)
that corresponds to the spatial extent of the scene and is a suitable input for a convolutional feature extractor.
Note that the encoder input feature space is in principle unrelated to the $\nR{}^3$ domain
of the generated completion function.
This means that the point cloud encoder can make use of additional information of
the sensor. We supply the reflectivity value of every LiDAR point as an extra feature.
The positions of LiDAR points are encoded as
separate coordinates relative to the mean position
of the points within the voxel and the voxel center.

\boldparagraph{Decoder for batch-norm conditioned classification} %
Spatial encoding is implicitly modeled with a local output function $\bff{}_{\text{L}}$
that needs to be conditioned on the latent vector $\bc_V{}$ of the feature extractor.
This single-position classification function is implemented as a \ac{MLP} that uses \ac{cbn} layers
to express its dependency on the latent vectors \cite{NIPS2017_7237}.
Hereby, the resulting mean and variance of feature maps is generated by an affine transformation
of the respective conditioning vectors.
Our method divides the latent coding $\bc_V$ into resolution-specific latent vectors $\bc_V\!=\!(\bc_{V1}, \bc_{V2}, \bc_{V3})$
and their associated relative positions $\Delta\bp_V \!=\! (\Delta\bp_1, \Delta\bp_2, \Delta\bp_3)$.
This information then conditions the output function from coarse to fine: Thus beginning with the
lowest resolution latent vector and adding more fine-grained information in the later
layers of the \ac{MLP}.
The decoder diagram on the right of \figref{fig:architecture_v2} illustrates
this setup.

\boldparagraph{Training details}%
Training the architecture involves common spatial augmentations of the input LiDAR
point clouds in sensor coordinates.
We use random uniform rotation over full \SI{360}{\degree},
random uniform scaling between $\pm$\SI{5}{\percent}, random uniform translations
between $\pm$\SI{5}{\centi\meter}.
When training we use a top-view input grid with $256\times256$ voxels
which results in a square with edge length of \SI{40.96}{\meter} within the scene.
The grid is initially centered over the area where the accumulated training targets
have been generated.
The voxel grid is shifted off-center
using normally-distributed offsets with standard deviation $\sigma=\SI{8}{\meter}$.
We sample a single free space point for each point in the input LiDAR point cloud and a single free space point within each empty voxel.
Additionally, \num{2500} random scene locations are sampled and contribute to the
consistency loss term, but do not have any other annotations.
When training, only two out of the four nearest local functions $f_L$ are evaluated
for each query point to be able to include almost twice as many query training targets in a single batch.
The two selected weighting coefficients $w$ are scaled up accordingly.
Depending on available VRAM and desired batch size
the total number of training targets is clipped to a maximum value.
For a KITTI scan with around \num{120000}~points and GPUs with 16~GB VRAM
we selected a batch size of two and \num{400000} training targets per GPU.
Training on four Tesla-V100-GPUs with an effective batch size of eight took around four days to complete.


\begin{figure*}[t]
\setlength{\tabcolsep}{7.5pt} 
\centering
\begin{tabular}{@{}lllllllllllllllllll@{}}
         \rotsemXX{\semcolor[road] Road} &
         \rotsemXX{\semcolor[sidewalk] Sidewalk} &
         \rotsemXX{\semcolor[parking] Parking} &
         \rotsemXX{\semcolor[otherground] other gr.} &
         \rotsemXX{\semcolor[building] Building} &
         \rotsemXX{\semcolor[fence] Fence} &
         \rotsemXX{\semcolor[car] Car} &
         \rotsemXX{\semcolor[truck] Truck} &
         \rotsemXX{\semcolor[othervehicle] other veh.} &
         \rotsemXX{\semcolor[bicycle] Bicycle} &
         \rotsemXX{\semcolor[motorcycle] Motorcycle} &
         \rotsemXX{\semcolor[person] Person} &
         \rotsemXX{\semcolor[bicyclist] Bicyclist} &
         \rotsemXX{\semcolor[motorcyclist] Motorcycl.} &
         \rotsemXX{\semcolor[vegetation] Vegetation} &
         \rotsemXX{\semcolor[trunk] Trunk} &
         \rotsemXX{\semcolor[terrain] Terrain} &
         \rotsemXX{\semcolor[pole] Pole} &
         \rotsemXX{\semcolor[trafficsign] Tr. Sign} \\
\end{tabular}
\setlength{\extrarowheight}{0pt} 
\setlength{\tabcolsep}{1pt} 
\def\scalemargin{0.04}
\def\scalemarginh{0.04}
\def\scalesection{6.25 / 51.2}
\def\scaleheight{0.02}
\def\customrowspace{-.5mm}
\begin{tabularx}{\linewidth}{@{}XXXX@{}}
\includegraphics[width=\linewidth]{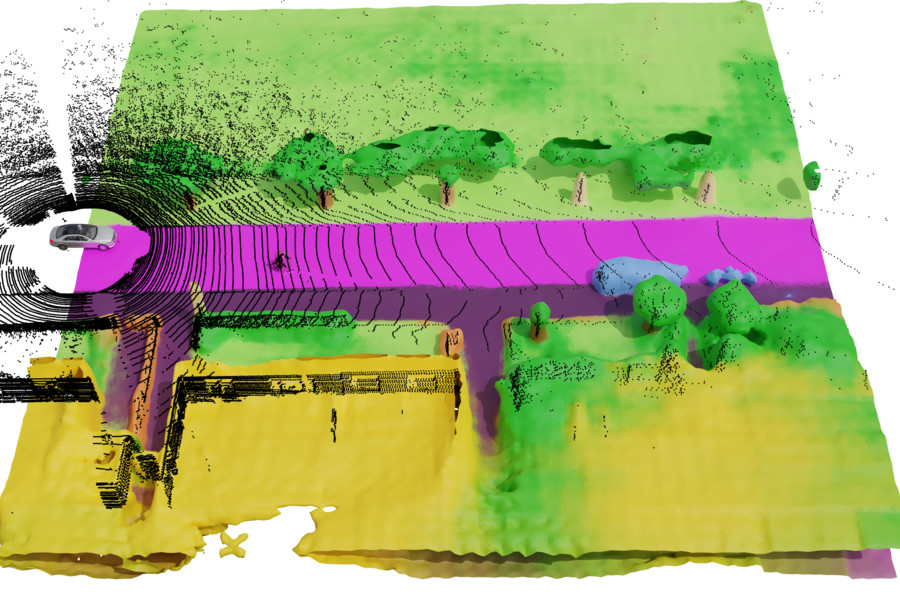} &
\includegraphics[width=\linewidth]{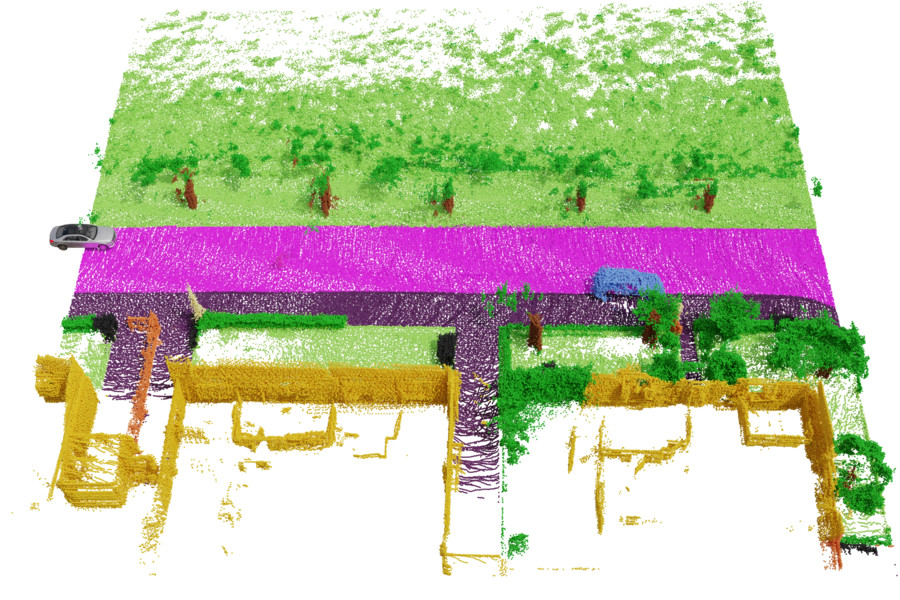} &
\includegraphics[width=\linewidth]{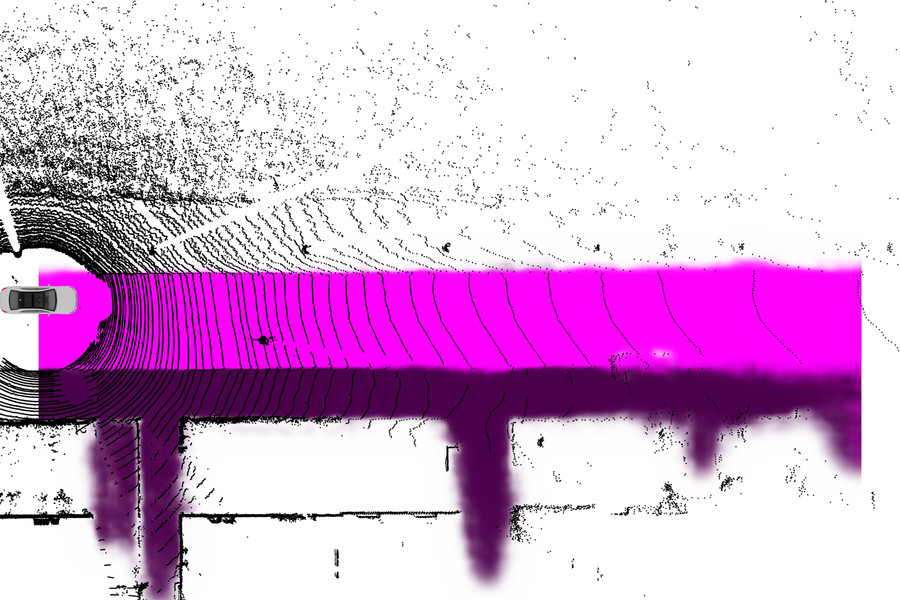} &
\makescaleimage{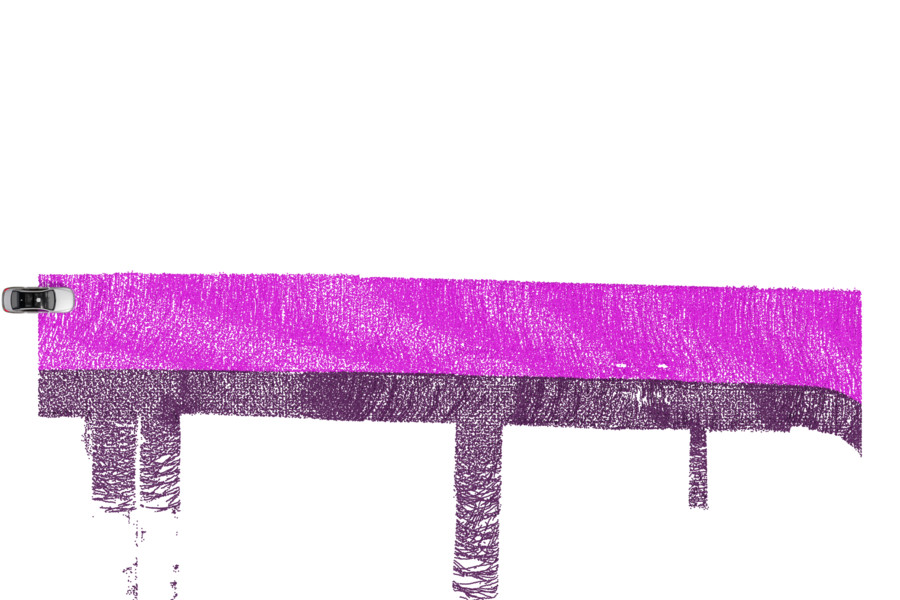}{12.5}{25} \\[\customrowspace]
\includegraphics[width=\linewidth]{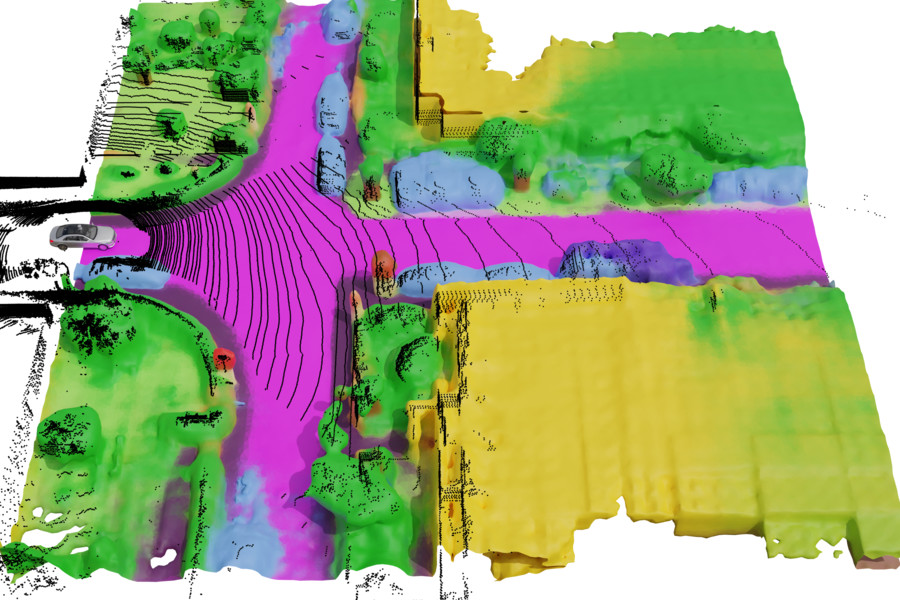} &
\includegraphics[width=\linewidth]{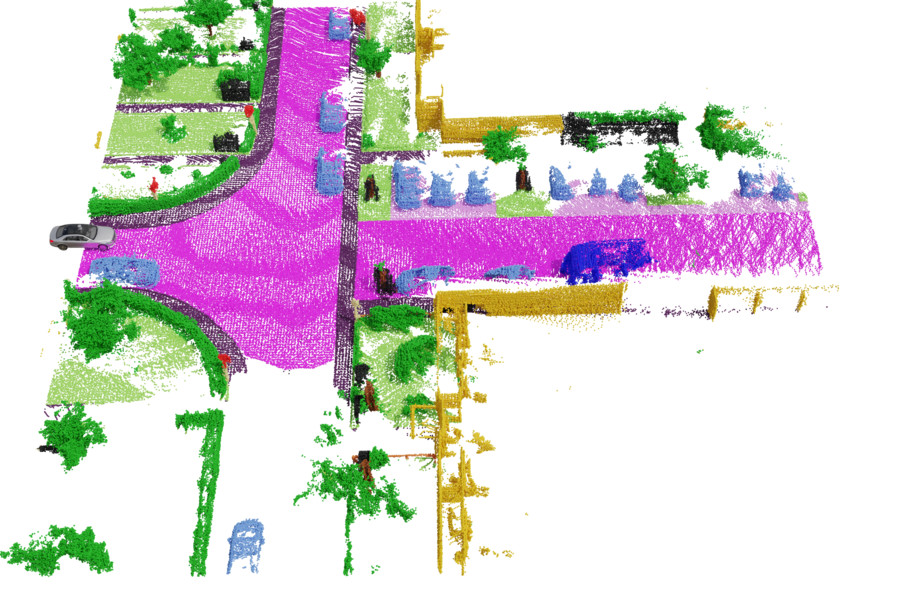} &
\includegraphics[width=\linewidth]{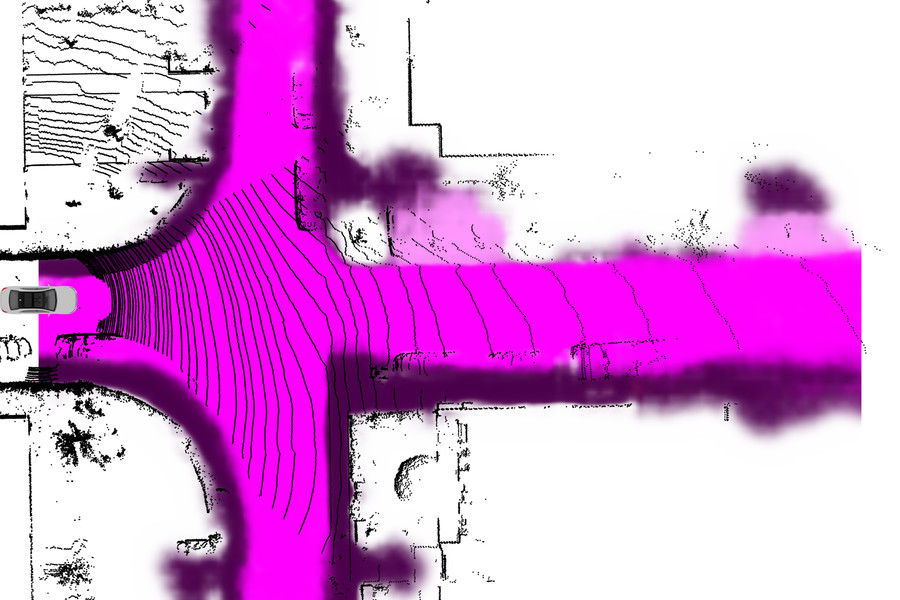} &
\makescaleimage{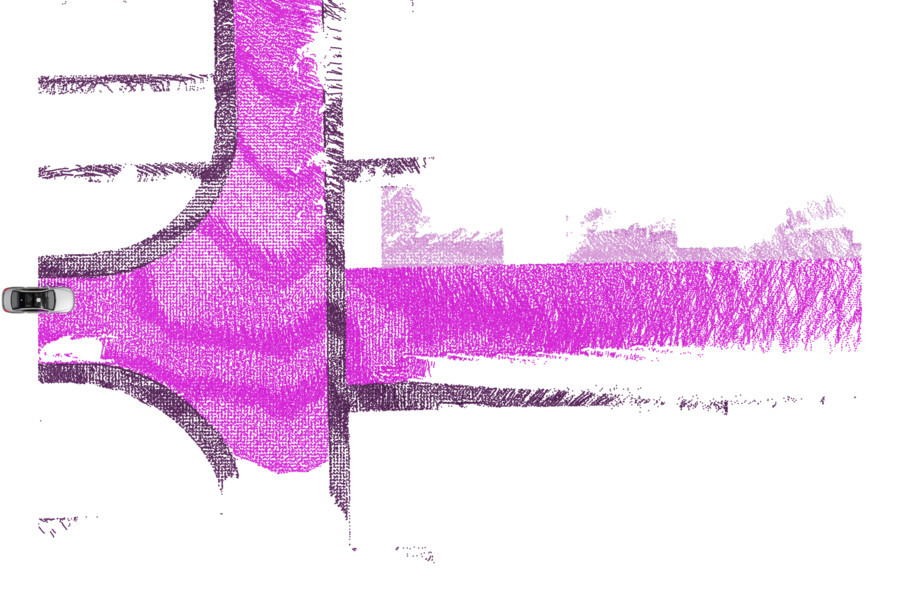}{12.5}{25} \\[\customrowspace]
\includegraphics[width=\linewidth]{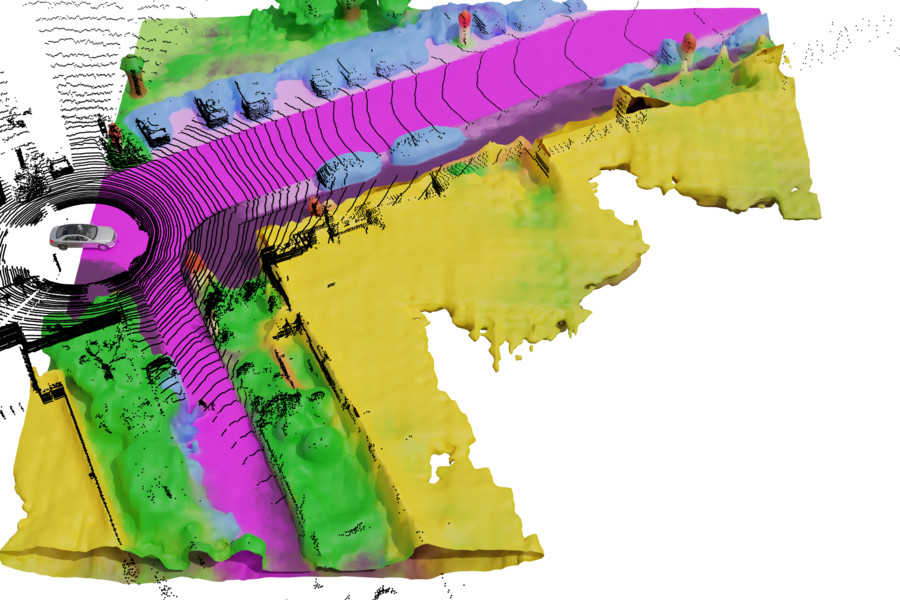} &
\includegraphics[width=\linewidth]{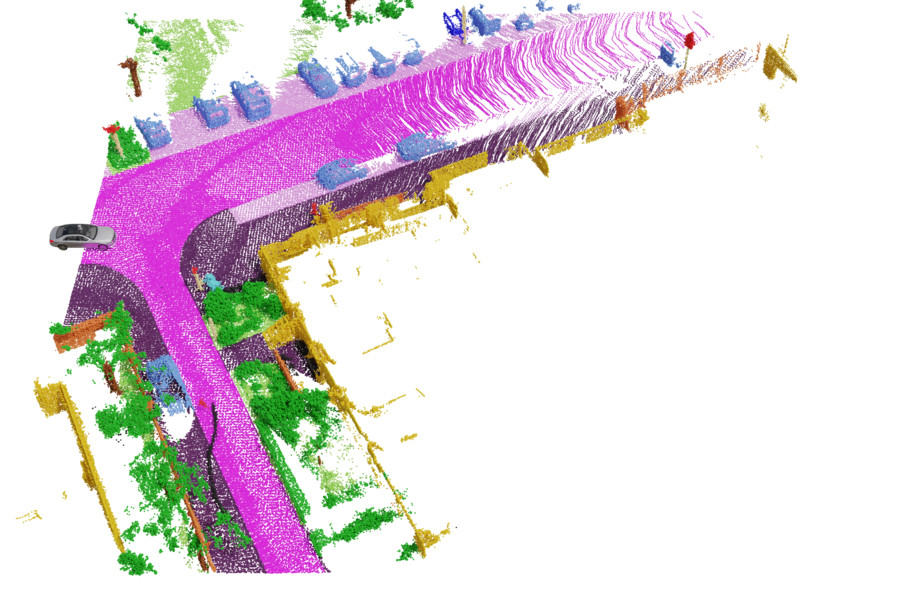} &
\includegraphics[width=\linewidth]{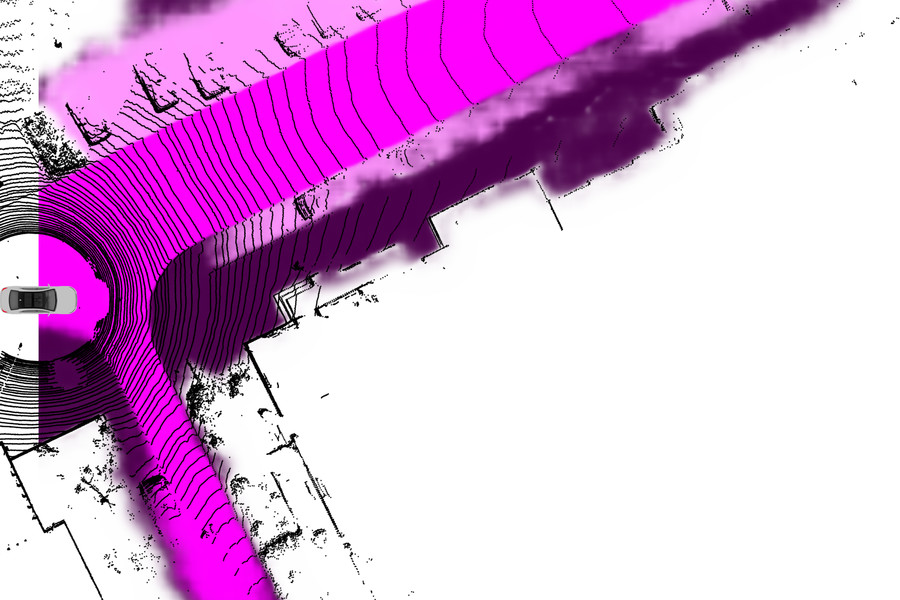} &
\makescaleimage{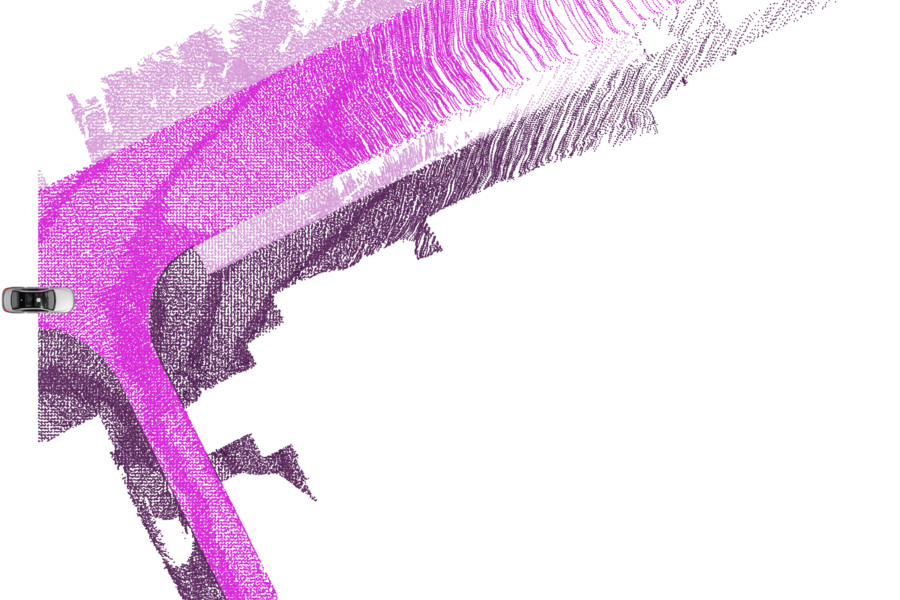}{12.5}{25} \\[\customrowspace]
\includegraphics[width=\linewidth]{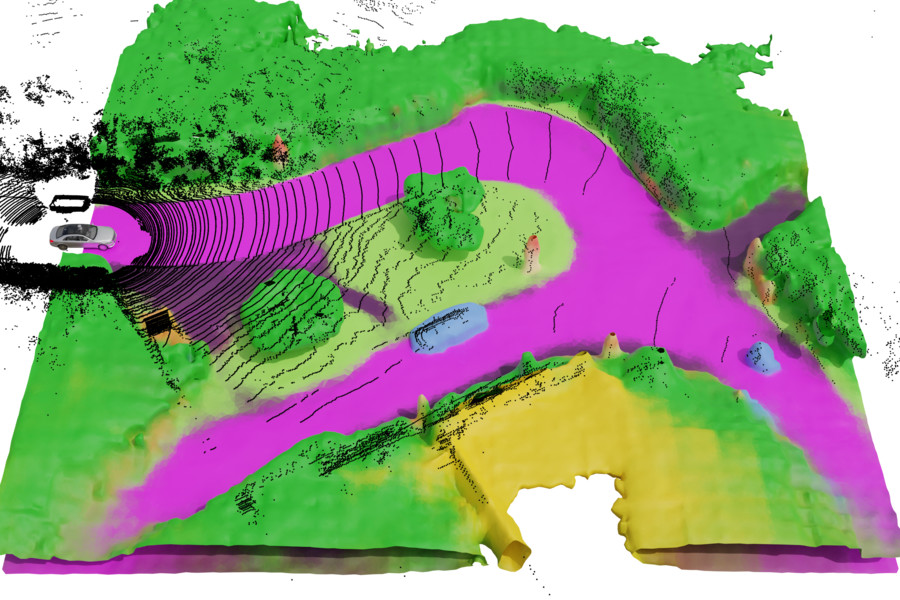} &
\includegraphics[width=\linewidth]{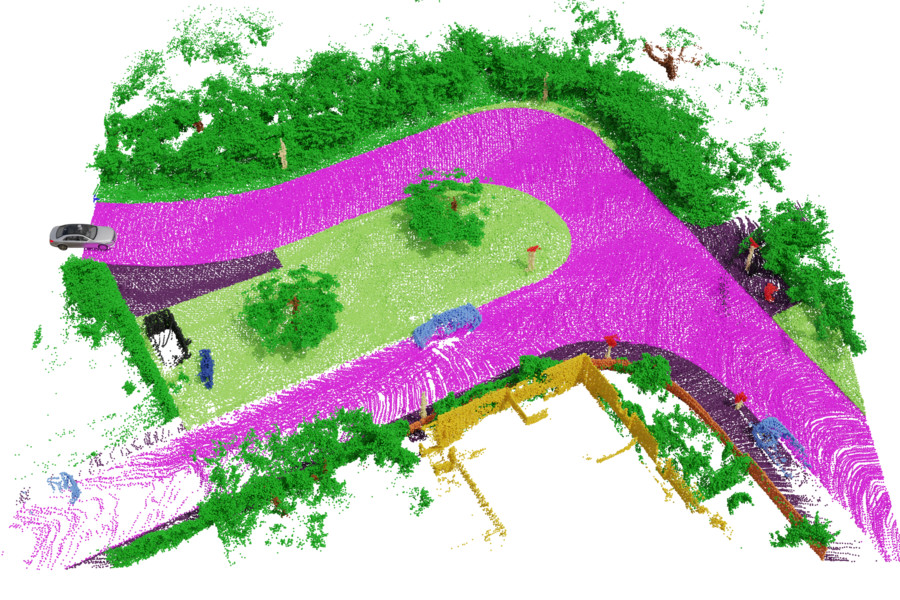} &
\includegraphics[width=\linewidth]{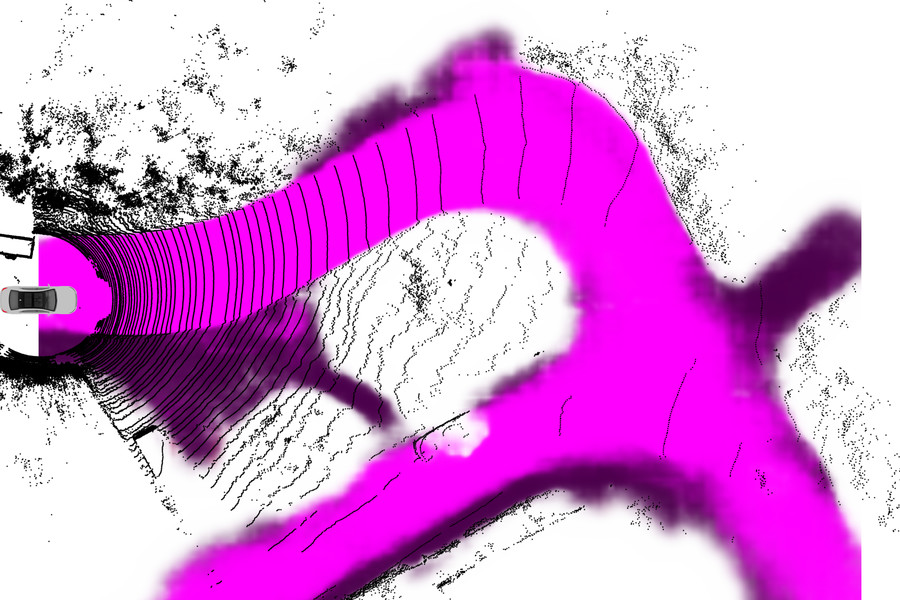} &
\makescaleimage{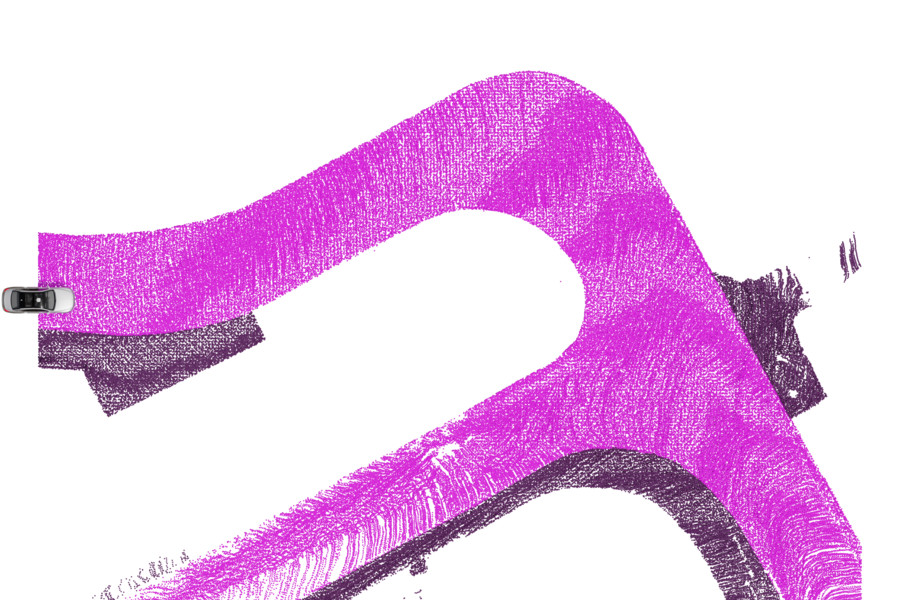}{12.5}{25} \\[\customrowspace]
\includegraphics[width=\linewidth]{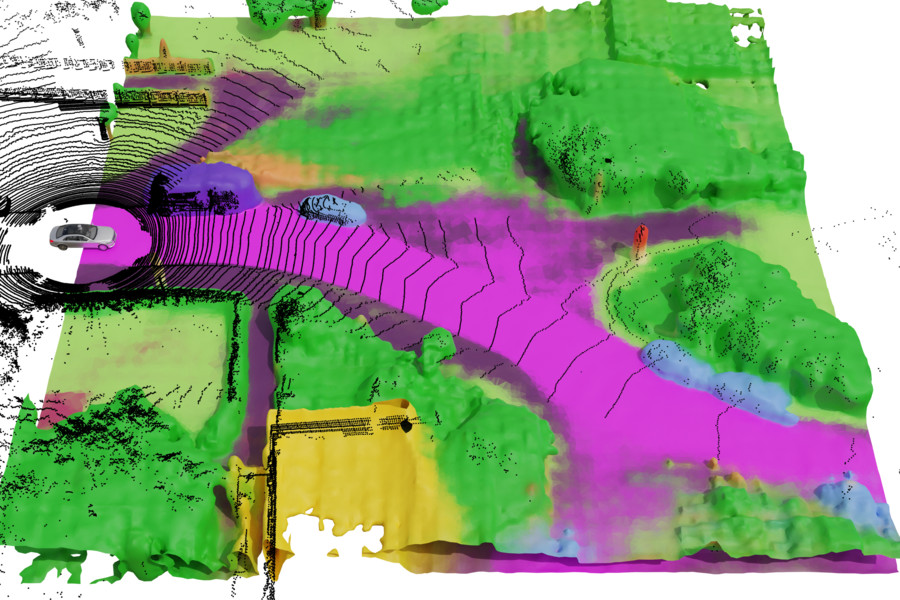} & %

\includegraphics[width=\linewidth]{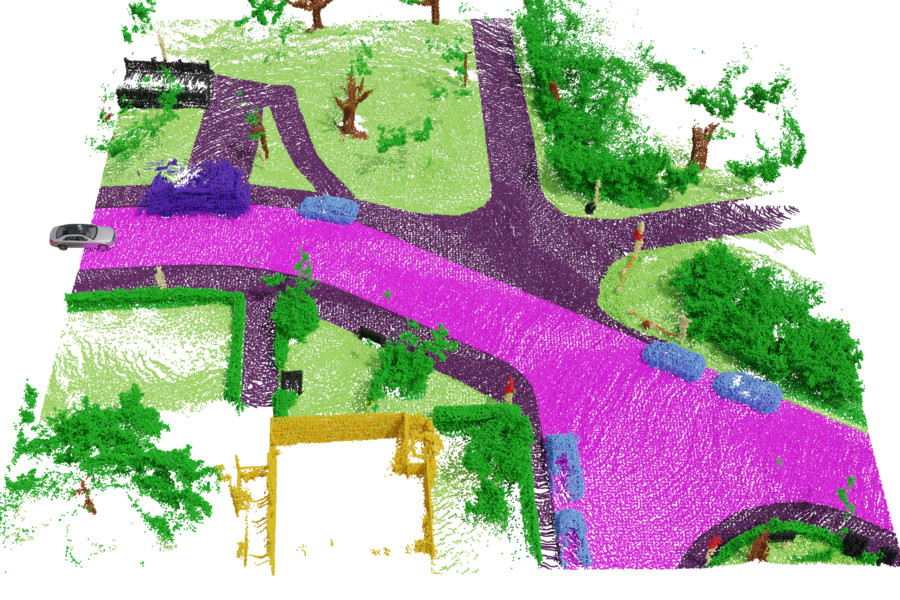} & %
\includegraphics[width=\linewidth]{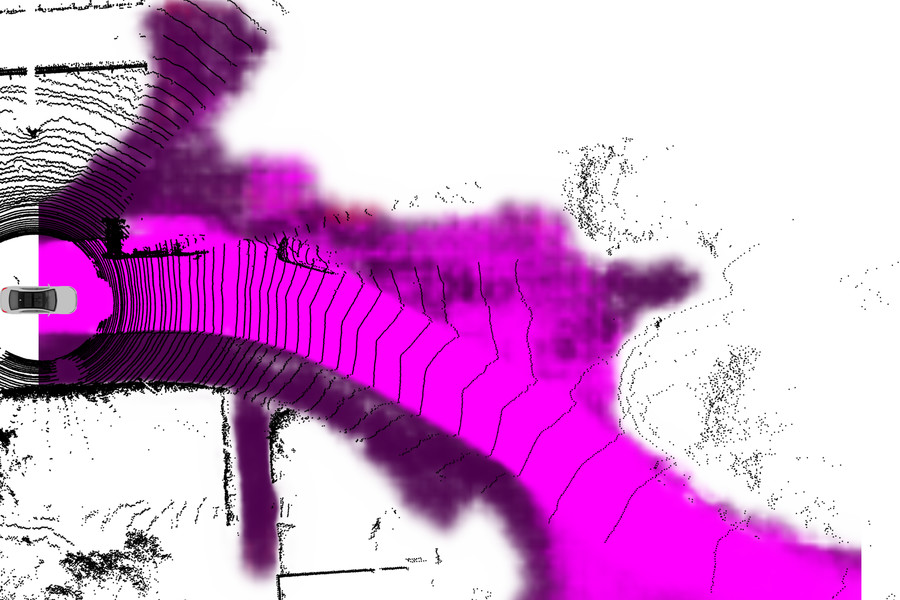} & %
\makescaleimage{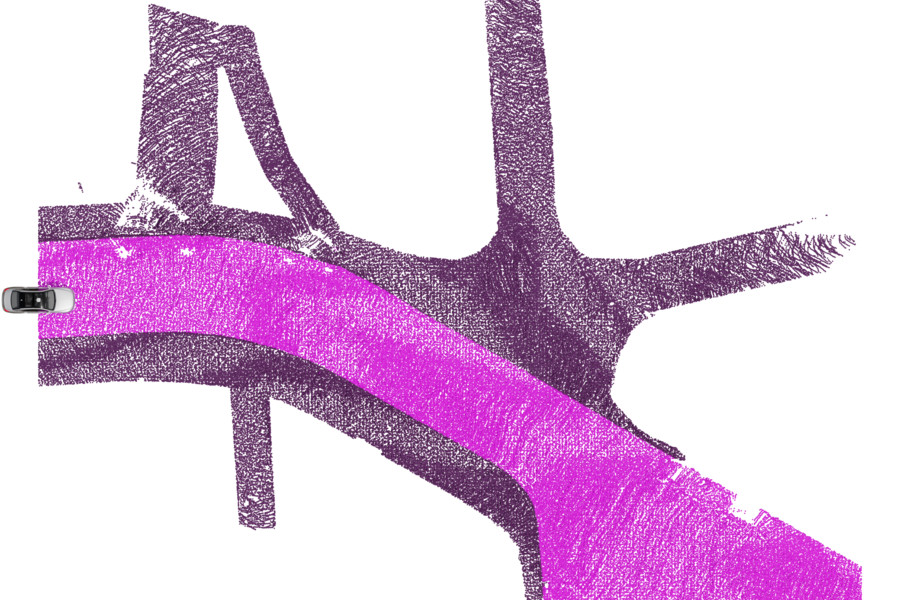}{12.5}{25} \\
\end{tabularx}
\caption{%
Columns from left to right: Completed scene, accumulated LiDAR as ground truth, ground segmentation, and corresponding ground truth.
Each row displays qualitative results and ground truth for a single scene on the Semantic KITTI validation set.
The single LiDAR scan used as input for our method is depicted as
an overlay of black points.
The far-right section
in each scene view demonstrates that
our approach is able to operate on areas that include hardly any LiDAR
measurements anymore.
The method is data-based and takes advantage of experience from the training dataset to facilitate predictions based on the larger context of the scene.
This is particularly visible from the completed courses of streets and sidewalks.
}
\label{fig:val_qualitative}
\end{figure*}

\subsection{Inference and Visualization}

We use latent conditioning vectors to define a function $\bff{}^\bc_{\text{\nn}}$ over $\nR{}^3$ to represent geometry
and semantics in a single classification vector.
Depending on the task at hand this implicit representation necessitates different procedures
to obtain explicit results.
In any case, the completion function is evaluated for an
arbitrary number of query coordinates at test time.
For point-wise semantic segmentation the positions of
the LiDAR points themselves are used as query points at test time to obtain semantic predictions
for the point cloud itself.
In this mode, it is previously known that none of the query positions can accurately be
classified as \emph{free space}.
Therefore, the predicted class value is just the argmax over all non-free-space semantic classes.

A regular voxel grid with per-voxel semantic information is a common
dense output structure for
semantic scene completion.
We query the completion function for all
corner points of all voxels on the voxel grid.
Every corner point is shared by eight voxels.
A voxel is marked as occupied when at least a single
corner of the voxel is assigned any occupied class.
The semantic label is averaged from
all corners which are predicted as occupied.
A threshold $\theta_{\text{empty voxel}}\in (0, 1)$ declares the
free space probability under which a coordinate is considered occupied.
This hyper-parameter controls the position on the precision-recall curve
for the occupied class and is tuned on the training set to
reach the maximum IoU of the occupied class.

To visualize the $\bff^\bc_{\text{\nn}}$ function we generate 3D meshes
to represent the isosurface of the scalar free space
function as close as possible (see \figref{fig:val_qualitative}, left column).
From the $N\!+\!1$ semantic classes of the vector-valued $\bff^\bc_{\text{\nn}}$ function
we extract the free space probability isosurface at a threshold
$\theta_{\text{free space}} \in (0, 1)$.
This isosurface
$\{ \bp \in \nR{}^3 | \left(\bff^\bc_\text{\nn}(\bp)\right)_{N+1} = \theta_{\text{free space}} \}$
resembles the estimated boundaries of all
objects in the scene and therefore
gives an idea of the learned scene representation.
To extract the mesh, we use \ac{MISE} \cite{Mescheder2019CVPR}.
\ac{MISE} evaluates points in an equally spaced grid from coarse to fine.
By only evaluating the points of interest close to the isosurface
the number of
calculations is reduced considerably. Subsequently, the marching cubes
algorithm is applied and the
resulting mesh is refined by minimizing a loss term
for each vertex using the proximity
to the desired threshold value and the gradient information for faces of the mesh.
This approach removes artefacts from the marching cubes algorithm and
requires that gradients \wrt{} the position of input points are available.
We query the $\bff{}^\bc_{\text{\nn}}$ function for all face-center positions of the resulting mesh
and color the mesh based on these semantic predictions.
\figref{fig:voxelization} compares the mesh visualization and voxelized
output that is obtained from the completion function.

We create a ground segmentation image to inspect the
completion function at positions which are hidden in the scene.
First, semantic segmentation is applied to the input point cloud.
The LiDAR points that are identified to belong to one of the ground classes
are selected.
Then, the positions of the selected ground points are used
for a bi-variate spline interpolation of all ground positions.
A dense regular top-view grid of predicted ground positions is extracted.
We query the completion function and display the predictions for the
previously selected ground classes as image.

\section{Experiments}

In this section, we first describe the details of our training dataset
and how it differs from the published Semantic KITTI scene completion
dataset.
Next, we introduce other published methods for real-world outdoor semantic scene completion
and compare the quantitative results on the closed test set through the
public benchmark.
Finally, we perform an ablation study about the upsampling architecture, hyperparameter
choices and semantic supervision signal.

\subsection{Dataset and LiDAR Accumulation}

The Semantic~KITTI authors construct the semantic scene completion task from LiDAR scans of the KITTI~Odometry
dataset \cite{Geiger2012CVPR} and their corresponding semantic annotations~\cite{Behley2019ICCV}.
The LiDAR sensor is a Velodyne HDL-64 that rotates with a frequency
of \SI{10}{\hertz}.
The continuously measured LiDAR points from a full revolution are
bundled into a LiDAR \emph{scan}.
The cut between scans is the negative $x$-axis in sensor coordinates so
that every scan begins and ends looking backwards.
LiDAR points are annotated with their respective semantic class.

\begin{figure}[t]
    \centering{}
    \includegraphics[width=\linewidth]%
    {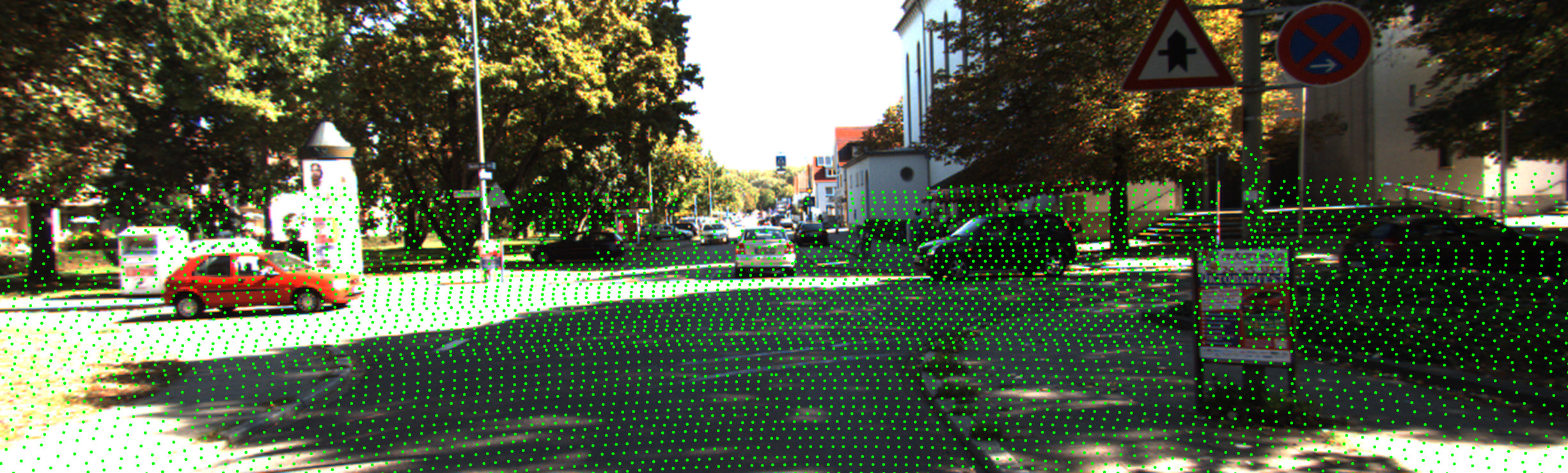}%
    \caption{%
    LiDAR scan (green) projected into reference RGB image.
    The vertical field of view of the KITTI LiDAR sensor only covers a
    range up to a few degrees over the horizon.
    Nevertheless, the resulting scene completion training targets cover
    objects at more than \SI{2}{\meter} over the ground
    since they are accumulated from more distant positions.
    \label{fig:lidar_rgb_proj}%
    }

\end{figure}

The recordings are made up of 21 sequences in total.
The data is split on a per-sequence basis: Ten sequences for training (\num{19130} point clouds),
one sequence for validation (\num{4071} point clouds)
and eleven sequences for testing (\num{20351} point clouds).
In the KITTI~Odometry dataset the LiDAR scans are already ego-motion corrected.
All points within a single \SI{360}{\degree} scan are transformed into the coordinate system
located at the sensor's position in the moment the sensor was looking in the direction
of the vehicle's front.
In addition, the Semantic~KITTI authors provide a frame-by-frame point cloud registration.
Sequences and registration are crucial as they allow to accumulate LiDAR measurements
of a longer time span into a single fixed \emph{reference coordinate system}.
This process creates the annotations of the semantic scene completion task
without requiring any additional manual annotations.
The Semantic~KITTI completion task combines the sequence of future LiDAR scans
to generate the completion target of the scene at the time of the input LiDAR scan.
This includes movements of dynamic objects and therefore requires to predict
object motion to solve the task in full.
\secref{sec:dyn_obj} details how we deviate from this handling
of dynamic objects and explains the \emph{static scene}
accumulation targets that we propose instead.

The Semantic~KITTI scene completion task uses a voxelized scene as output
representation.
A voxelized input LiDAR scan is also provided next to the raw LiDAR scan
from the KITTI~Odometry dataset.
However, we do not use the provided voxelized scene to train our method
as it is designed to classify individual positions.
Instead of creating a labeled voxel grid from accumulated LiDAR measurements
we use all of the individual points as training targets.
The accumulated point clouds are sub-sampled to include only a maximum of 10 points
within each original Semantic~KITTI voxel.
This reduces the overall dataset size and eliminates
a large part of the redundancy in regions that are scanned by the sensor in multiple frames.
The second column of \figref{fig:val_qualitative} shows examples of the
accumulation result.
The input LiDAR point cloud is shown as on overlay over the prediction in the first column (left).
We use the same
extent for accumulation as the Semantic~KITTI scene completion dataset:
A square with \SI{51.2}{\meter} edge length where the ego-vehicle is located
in the middle of an edge facing the center of the square.

The difficulty of the scene completion task gets apparent when looking at the pronounced sparsity of the input point cloud in a distance of around \SI{50}{\meter}
from the sensor.
In sparse regions most geometric details have to be inferred from scene context.
It is apparent that there are geometric and semantic ambiguities within the 3D scenes
which cannot be decided with high confidence from the single input LiDAR scan.
\figref{fig:lidar_rgb_proj} shows a projection of the LiDAR point cloud into the
camera view of the ego-vehicle.
The Velodyne HDL-64 sensor features a vertical field of view
that at the top only covers a few degrees over the level horizon.
Thus, in the vicinity of the ego-vehicle the LiDAR only covers a
height of about \SI{2}{\meter} over ground.
The scene completion target does however include geometry further up because
it includes LiDAR points that were recorded from a greater distance of
the ego-vehicle.
This is another prominent ambiguity of the training data that requires
a method to guess \eg\ if there is a traffic sign attached to a pole without
actual evidence from the sensor.

\subsection{Handling of Dynamic Objects\label{sec:dyn_obj}}

\begin{figure}[t]
    \centering{}
    \definecolor{colhighlight}{gray}{0.9}
    \newcolumntype{H}{>{\columncolor{colhighlight}}r}
    \subfloat[Static scene]{%
    \includegraphics[width=.49\linewidth]%
    {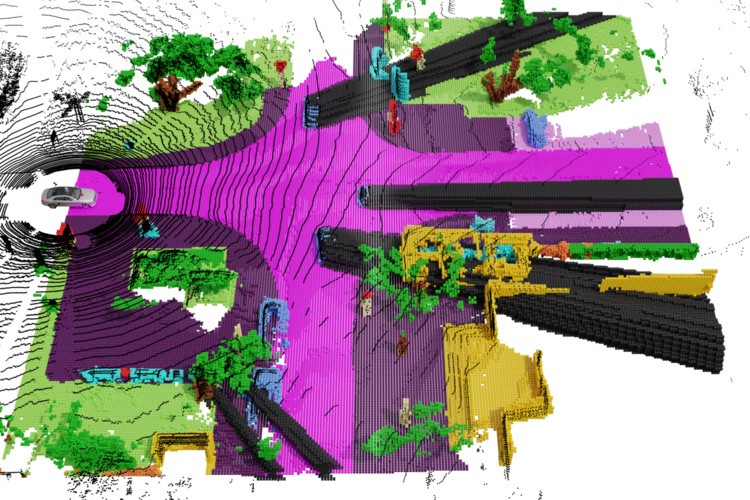}%
    \label{fig:dataset_static}}
    \hspace{1mm}%
    \subfloat[Spatio-temporal tubes]{%
    \includegraphics[width=.49\linewidth]%
    {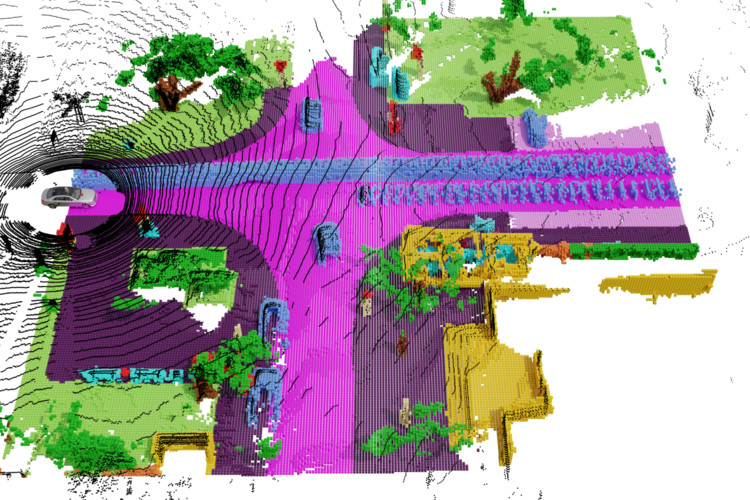}%
    \label{fig:dataset_temporal}}
    \vspace{2mm}
    \renewcommand{\arraystretch}{1.0} 
    \setlength{\tabcolsep}{3pt} 
    \footnotesize%
    \begin{tabularx}{\linewidth}{@{}lr|rrrr@{}}%
        \toprule%
        Dataset variant & \makecell[cr]{Occupied \\ IoU} &
        \makecell[cr]{Semantic \\ mIoU} & \makecell[cr]{Car \\ IoU} & \makecell[cr]{Person\\IoU} & \makecell[cr]{Bicyclist\\IoU} \\ \midrule
        (\protect\subref{fig:dataset_static}) Static scene &%
        57.8 &
        \textbf{26.1} & \textbf{51.3} & \textbf{15.7} & \textbf{24.7}\\[4pt]
        \makecell[cl]{(\protect\subref{fig:dataset_temporal}) Spatio-temporal\\object tubes} &%
        57.6 &
        24.0 & 45.6 & 3.3 & 0.9 \\ \bottomrule
    \end{tabularx}%
    %
    \caption{%
    Our dataset ((\protect\subref{fig:dataset_static}) static scene)
    and the official benchmark ((\protect\subref{fig:dataset_temporal}) spatio-temporal tubes) handle dynamic objects differently.
    We remove all free space targets within the shadows of dynamic objects (marked as black regions) to obtain a consistent static scene.
    We evaluate the same model on both variants to measure this difference quantitatively.
    The impact on overall reconstruction performance in terms of 
    IoU for \emph{occupied} and \emph{free space} class is marginal
    because of the prevalence of voxels belonging to static objects.
    However, the impact on IoU of small object classes that are primarily dynamic (e.g. \emph{Person}, \emph{Bicyclist})
    is significant and leads to an increase in mIoU over all classes of about \SI{2.1}{\percent}.
    The comparison highlights that our method is in fact able to
    recognize smaller traffic participants.
    But an additional requirement to predict their motion will hide this ability.
    \label{fig:benchmark_voxelization}%
    }
\end{figure}

We use \emph{static} training and evaluation data
for the semantic scene completion task.
We regard this variant as more suitable for a meaningful evaluation of performance compared
to the handling of dynamic objects in the original scene completion annotations.
The KITTI~Odometry scenes contain dynamic objects such as moving cars and pedestrians.
These objects are additionally annotated with a \emph{dynamic} flag.
The original Semantic~KITTI scene completion data 
accumulates the occupied voxels from dynamic objects
in the reference frame just as the voxels of any other static object.
Effectively this creates \emph{spatio-temporal tubes}
of moving traffic participants along their respective path.
Therefore, fully solving the Semantic~KITTI scene completion task
requires predicting the future trajectories of traffic participants.
As we focus on geometric reconstruction of the scene in the instant of the input
LiDAR scan we take a different approach for dynamic objects:
When accumulating LiDAR measurements, we only keep the single current scan on dynamic
objects. By omitting the following scans over dynamic objects no trajectory \emph{tube}
is created. Next, it is necessary to ensure that no free space points get sampled
within the extent of a dynamic object. As the object potentially moves from its initial position,
the following LiDAR scans will record the initial position as free space.
So to prevent free space targets within the actual object we record the shadow
cast by the object in the first frame and treat the occluded regions as unseen
regions where no free space points are sampled (see black regions in \figref{fig:dataset_static}).
These two measures make the replicated geometry consistent in the presence
of dynamic objects.
The resulting set of training targets reflects the true scene at the moment of the input LiDAR
scan. Areas where we cannot obtain consistent targets from future frames are ignored in the training.

In \figref{fig:benchmark_voxelization} we compare the two approaches for dynamic objects and show an example.
We quantitatively measure the difference in performance when using the different
dataset targets for evaluation.
Note, that in this comparison, our method is trained on our \emph{static} version of
the data in both cases.
This allows us to better judge the performance reported by the benchmark on the
private test set.
We see that there is almost no quantitative difference for the geometric completion
evaluation (\emph{Occupied IoU}) because static objects are prevalent over
dynamic objects.
However, for semantic scene completion we expect a significant difference.
Object classes with a large proportion of dynamic voxels perform much worse
if a method does not predict the object's movement.
By not requiring our method to predict complicated object trajectories
of even completely invisible objects we generate a consistent supervision signal.
Qualitative results of other methods~\cite{Behley2019ICCV,yan2021sparse,Cheng2020} on Semantic KITTI show that they do not predict tubes as well, but instead also complete dynamic objects as if they were static.
Having said this, the benchmark metric of course penalizes all methods equally
for not predicting spatio-temporal objects tubes for dynamic objects.

\subsection{Scene Completion Evaluation}

In accordance with the scene completion benchmark
\cite{Behley2019ICCV} we use the
\ac{mIoU} metric to assess both geometric completion performance
and semantic segmentation accuracy.
This metric is calculated on a per-voxel basis for the
semantic scene completion task and
on a per-point basis for single-scan LiDAR semantic segmentation.
The semantic scene completion task is ranked by the
\ac{mIoU} value over all semantic classes including the free space class.
The mere geometric completion performance is rated by the IoU value over all
occupied classes combined, that is all classes except for free space.

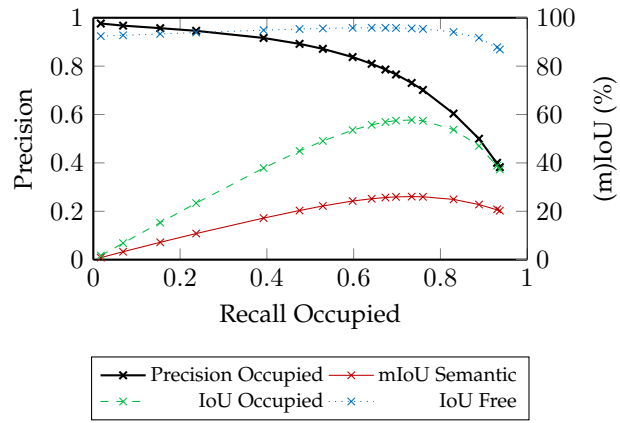
\begin{figure}[tb]
    \centering{}
   \begin{tikzpicture}[trim axis left, trim axis right]
        \definecolor{clr1}{RGB}{190,0,0}
        \definecolor{clr2}{RGB}{0,190,70}
        \definecolor{clr3}{RGB}{0,112,191}
        \begin{axis}[
            axis y line*=left,
            name=ax1,
            xmin=0.0,
            xmax=1.0,
            ymin=0.0,
            ymax=1.0,
            ylabel=Precision,
            xlabel=Recall Occupied,
            height=4.8cm,
            width=7.35cm,
            thick,
        ]
                \addplot [thick, mark=x] table [x=recall, y=precision, col sep=comma]%
                {data/validation_pr/data_pr.csv}; \label{plot_precision}
        \end{axis}
        \begin{axis}[
            axis y line*=right,
            axis x line=none,
            xmin=0.0, xmax=1.0,
            ymin=0, ymax=100,
            ylabel=(m)IoU (\%),
            height=4.8cm,
            width=7.35cm,
            legend cell align={right},
            legend style={
                at={(0.5,-0.4)},
                anchor=north,
                legend columns=2,
                nodes={scale=0.8, transform shape},
            },
        ]
            \addlegendimage{/pgfplots/refstyle=plot_precision}\addlegendentry{Precision Occupied}
            \addplot [mark=x, clr1] table [
                x=recall,
                y expr=\thisrow{miou}*100,
                col sep=comma,
                ]%
            {data/validation_pr/data_pr.csv}; \addlegendentry{mIoU Semantic}
            
            \addplot [mark=x, clr2, dashed, mark options={solid}] table [
                x=recall,
                y expr=\thisrow{occupied}*100,
                col sep=comma,
                ]%
            {data/validation_pr/data_pr.csv}; \addlegendentry{IoU Occupied}
            
            \addplot [mark=x, clr3, dotted, mark options={solid}] table [
                x=recall,
                y expr=\thisrow{freespace}*100,
                col sep=comma,
                ]%
            {data/validation_pr/data_pr.csv}; \addlegendentry{IoU Free}

        \end{axis}
    \end{tikzpicture}
    \caption{%
    \textbf{Precision-recall curve for the occupied class.}
    We plot the (m)IoU values for occupied, free and semantic classes of the baseline network variant. Markers are at the free space thresholds that are evaluated, interpolation in between.
    \label{fig:validation_pr}%
    }

\end{figure}

The threshold $\theta_{\text{empty voxel}}\in (0, 1)$ is selected individually for each network
variant based on the training set.
This ensures that precision and recall values are balanced out, resulting in the
respective maximum value for completion IoU and semantic mIoU.
\figref{fig:validation_pr} plots the precision-recall-curve for the occupied class
on the validation set together with IoU values for our best performing network.

We apply \ac{TTA} to our best performing approach for better comparison to the concurrent work JS3CNet. The regular predictions and predictions with \ac{TTA} are submitted separately to the benchmark. \ac{TTA} is implemented by augmenting the input point cloud at test time and averaging over the lattice grid predictions before generating the final voxel grid.

\subsection{Semantic Scene Completion Benchmark Results}


\begin{table*}[t]

    \definecolor{colhighlight}{gray}{0.9}
    \newcolumntype{H}{>{\columncolor{colhighlight}}r}

    \centering

    \caption{Quantitative scene completion results for our method and recently published approaches on the Semantic KITTI Scene Completion Benchmark (in Intersection-over-Union, higher is better). $\dagger$: Method uses \acl{TTA}.}
    \label{tab:completion_eval}
    \centering
    \setlength{\tabcolsep}{2pt}
    \begin{tabularx}{\linewidth}{@{}Xr|rrrrrrrrrrrrrrrrrrrrl@{}}
        \toprule %
        \multicolumn{2}{r|}{\textbf{Geometric Completion}} &
        \multicolumn{19}{l}{\textbf{Semantic Completion}} \\
         \textbf{Method / IoU [\%]} &
         \textbf{Occ.} &
%
         \textbf{mIoU} &
         \rotsem{\semcolor[road] Road} &
         \rotsem{\semcolor[sidewalk] Sidewalk} &
         \rotsem{\semcolor[parking] Parking} &
         \rotsem{\semcolor[otherground] other gr.} &
         \rotsem{\semcolor[building] Building} &
         \rotsem{\semcolor[fence] Fence} &
         \rotsem{\semcolor[car] Car} &
         \rotsem{\semcolor[truck] Truck} &
         \rotsem{\semcolor[othervehicle] other veh.} &
         \rotsem{\semcolor[bicycle] Bicycle} &
         \rotsem{\semcolor[motorcycle] Motorcycle} &
         \rotsem{\semcolor[person] Person} &
         \rotsem{\semcolor[bicyclist] Bicyclist} &
         \rotsem{\semcolor[motorcyclist] Motorcycl.} &
         \rotsem{\semcolor[vegetation] Vegetation} &
         \rotsem{\semcolor[trunk] Trunk} &
         \rotsem{\semcolor[terrain] Terrain} &
         
         \rotsem{\semcolor[pole] Pole} &
         \rotsem{\semcolor[trafficsign] Tr. Sign} &
         \hphantom{|} \\ \midrule
%
\makecell[cl]{TS3D \cite{Garbade2019CVPRWorkshops,Behley2019ICCV}} &
50.6 &
%
17.7 &%
62.2 & 31.6 & 23.3 &  6.5 & 34.1 &%
24.1 &%
30.7 & 4.9 &%
0.1 &
0.0 & 0.0 &
0.0 &  0.0 &  0.0 &%
40.1 & 21.9 & 33.1 &%
16.9 &  6.9 \\
\makecell[cl]{LMSCNet-singlescale \cite{Roldao2020}} &%
56.7 &
%
17.6 &%
64.8 & 34.7 & 29.0 & 4.6 & 38.1 &%
21.3 &%
30.9 & 1.5 &%
0.8 &%
0.0 & 0.0 &%
0.0 & 0.0 & 0.0 &%
41.3 & 19.9 & 32.1 &%
15.0 & 0.8 \\
\makecell[cl]{JS3CNet $\dagger$ \cite{yan2021sparse}} &%
56.6 &
23.8 &%
64.7 & 39.9 & 34.9 & \textbf{14.1} & 39.4 &%
30.4 &%
33.3 & \textbf{7.2} &%
12.7 &%
14.4 & 8.8 &%
8.0 & 5.1 & 0.4 &%
43.1 & 19.6 & 40.5 &%
18.9 & 15.9 \\
\makecell[cl]{S3CNet \cite{Cheng2020}} &%
45.6 &
\textbf{29.5} &%
42.0 & 22.5 & 17.0 & 7.9 & \textbf{52.2} &%
\textbf{31.3} &%
31.2 & 6.7 &%
\textbf{16.1} &%
\textbf{41.5} & \textbf{45.0} &%
\textbf{45.9} & \textbf{35.8} & \textbf{16.0} &%
39.5 & \textbf{34.0} & 21.2 &%
\textbf{31.0} & \textbf{24.3} \\
\midrule%
%
\textbf{\methodname{} (ours)} &%
57.7 &
%
22.7 &
67.9 & 42.9 & 40.1 & 11.4 & 40.4 &%
29.0 &%
34.8 & \phantom{0}4.4 &%
\phantom{0}4.8 &%
\phantom{0}3.6 & \phantom{0}2.4 &%
\phantom{0}2.5 & \phantom{0}1.1 & \phantom{0}0.0 &%
42.2 & 26.5 & 39.1 &%
21.3 & 17.5 \\
%
%
%
\methodname{} + \acs{TTA} $\dagger$ &%
\textbf{58.9} &
%
23.6 &
\textbf{69.6} & \textbf{44.5} & \textbf{41.8} & 12.7 & 41.3 &%
30.5 &%
\textbf{35.4} & \phantom{0}4.7 &%
\phantom{0}4.7 &%
\phantom{0}3.6 & \phantom{0}2.7 &%
\phantom{0}2.4 & \phantom{0}1.0 & \phantom{0}0.0 &%
\textbf{43.8} & 27.4 & \textbf{40.9} &%
22.1 & 18.5 \\
\bottomrule%
        \end{tabularx}
\end{table*}

We compare our approach against four recently published deep-learning-based methods
on the challenging outdoor LiDAR semantic scene completion task.
Quantitative results are reported by their respective authors on the benchmark and
are compared in \tabref{tab:completion_eval}.

The performance of our method surpasses all other methods in pure geometric completion
performance (\SI{57.7}{\percent}).
Here we exceed the second-best performing method LMSCNet-singlescale \cite{Roldao2020}
by a margin of \SI{1.0}{\percent}.
The authors of JS3CNet~\cite{yan2021sparse} only report benchmark results with \ac{TTA}, so we use \ac{TTA} as well for comparison. 
JS3CNet achieves a marginally higher mIoU (+\SI{0.2}{\percent}) than our method with \ac{TTA}, while being considerable inferior in geometric completion (-\SI{2.3}{\percent} IoU).
JS3CNet is more accurate on small object classes and less accurate on the larger ground classes.
S3CNet~\cite{Cheng2020} outperforms all other methods by a large margin on the semantics of small object classes,
resulting in the best mIoU value.
For the other object classes, it does however perform comparably or even worse to our method.
Overall, when it comes to geometric accuracy, S3CNet underperforms significantly.
This might be a result of the semantic post-processing steps.


\setlength{\tabcolsep}{3.5pt} 
\begin{table}[tb]
    \definecolor{colhighlight}{gray}{0.9}
    \newcolumntype{H}{>{\columncolor{colhighlight}}r}
    \caption{Network parameter count for architecture variants}
    \label{tab:network_params}
    \centering
    \begin{tabularx}{\linewidth}{@{}l|Hrrrr@{}}
        \toprule
         & & \multirow{2}{*}{\makecell[cc]{Point\\feat.}} & \multicolumn{2}{c}{Encoder} & Decoder \\
         \textbf{Variant} & \textbf{$\Sigma$} & & Convs & Upsample & \\%
         \midrule{}%
\methodname{} & \num{9892788} & \num{1280} & \num{7123648} & \num{1656192} & \num{1111668} \\
         \scssnetUpsample{} & \num{7712308} & \num{1280} & \num{7123648} & \num{0} & \num{587380} \\%
\scssnetSingle{} & \num{9556340} & \num{1280} & \num{7123648} & \num{1656192} & \num{775220} \\
Feature interp. & \num{9897364} & \num{1280} & \num{7123648} & \num{1656192} & \num{1116244} \\
         \bottomrule
    \end{tabularx}
\end{table}

\subsection{Ablation Study}

We use the Semantic KITTI validation split and the static scene data variant for evaluation of the ablation study.
All ablation results are listed in \tabref{tab:ablations_reduced}.


\setlength{\tabcolsep}{1.7pt} 

\begin{table}[tb]
    \definecolor{colhighlight}{gray}{0.9}
    \newcolumntype{H}{>{\columncolor{colhighlight}}r}
    \caption{Quantitative results of baseline and ablations on the validation set (higher is better)}
    \label{tab:ablations_reduced}
    \centering
    \begin{tabularx}{\linewidth}{@{}clHrr|H@{}}
        \toprule
        & &
        \multicolumn{3}{@{}c@{}}{\textbf{Geometric Completion}} &
        \makecell[cr]{Semantic\\Completion} \\
        & \textbf{Variation} & \textbf{IoU Occ.} & %
        Precision & Recall & 
        \textbf{mIoU} 
        \\ \midrule
        & \makecell[cl]{\methodname{} (Baseline)} & %
        57.8 & 73.1 & 73.4 & %
        \textbf{26.1}
        \\%
        & \makecell[cl]{(\methodname{} + \acs{TTA}}) & %
        (58.5) & 74.2 & 73.5 & %
        (26.9)%
        \\ \midrule{}
        \multirow{3}{*}{\rotatebox[origin=c]{90}{\textbf{Arch.}}} %
        & \makecell[cl]{\scssnetUpsample{}} & %
        55.4 & 71.9 & 70.8 & %
        23.8 %
        \\ %
        & \makecell[cl]{\scssnetSingle{}} &
        57.1 & 72.7 & 72.6 & %
        24.2 %
        \\ %
        & \makecell[cl]{Feature interpolation} &
        57.4 & 73.0 & 73.0 & %
        25.5 %
        \\ %
        \midrule{} %
        \multirow{4}{*}{\rotatebox[origin=c]{90}{\textbf{Cell size}}} %
        & \makecell[cl]{Cell size \SI{75.0}{\percent}} &
        54.1 & 70.6 & 69.8 & %
        23.8 %
        \\ %
        & \makecell[cl]{Cell size \SI{87.5}{\percent}} &
        57.1 & 73.8 & 71.7 & %
        25.6
        \\ %
        & \makecell[cl]{Cell size \SI{150}{\percent}} &
        56.7 & 71.6 & 73.1 & %
        24.1 %
        \\ %
        & \makecell[cl]{Cell size \SI{200}{\percent}} &
        56.7 & 72.5 & 72.3 & %
        23.3
        \\ %
        \midrule{} %
        \multirow{3}{*}{\rotatebox[origin=c]{90}{\textbf{Loss}}} %
        & \makecell[cl]{$\lambda_S=15,\lambda_G=1$} &
        55.6 & 71.1 & 71.7 & %
        24.0 %
        \\ %
        & \makecell[cl]{$\lambda_S=3.75,\lambda_G=4$} &
        \textbf{58.2} & 74.5 & 72.7 & %
        24.7 %
        \\ %
        & \makecell[cl]{$\lambda_C=0$} &
        56.9 & 72.0 & 73.0 & %
        25.0 %
        \\ %
        \midrule{} %
        \multirow{2}{*}{\rotatebox[origin=c]{90}{\textbf{Data}}} &
        \makecell[cl]{Simplified sem.} &
        57.8 & 74.1 & 72.3 & %
        (38.8)%
        \\%
        & \makecell[cl]{Without sem.} &
        57.9 & 73.6 & 73.1 & %
        (57.9) %
        \\ \bottomrule
    \end{tabularx}
\end{table}

\psection{Multi-resolution upsampling and decoder variants (\tabref{tab:ablations_reduced}, architecture)}%
The individual local functions are arranged in a grid
where each cell has an edge length of \SI{0.32}{\meter}.
The encoder uses a number of pooling layers and
generally produces feature maps at lower
resolutions of up to 16 times the output grid size. 
Our baseline \methodname{} variant achieves a high resolution
output grid by two independent upsample approaches.
The first is upsampling and concatenating the lower resolution
feature maps progressively in the encoder.
The second is to supply pairs of relative coordinates and conditioning vectors
for different resolutions.
The decoder then handles the fusion of multiple feature maps.
The conditioned-batch normalization (CBN) works as an attention mechanism
between latent vector and query position.
This variant is unique to decoder architecture based on \acp{DIF}.

We drop one of the two upsample approaches at a time resulting in
two model architecture variants:
\emph{\scssnetUpsample{}} does not have
transposed convolutions for upsampling in the encoder.
\emph{\scssnetSingle{}} uses only the highest resolution feature map in the decoder.
Both completion and semantic scene completion performance is highest when
using the baseline model that can rely on both upsample pathways.
Building the decoder only on the high resolution feature map in \emph{\scssnetSingle{}}
reduces performance to a lesser extent than removing the transposed convolutions
in the encoder in \emph{\scssnetUpsample{}}.
In both cases the drop in semantic scene completion is more noticeable than the drop
in pure geometric completion.
\emph{\scssnetUpsample{}} reduces the number of trainable parameters
compared to the baseline to about \SI{78}{\percent} (\tabref{tab:network_params}).
The transposed convolutions account for a considerable share of the total parameters of the encoder.
This experiment indicates that a decoder based on coarse grid cells
together with coordinates as an attention mechanism
can reduce the number of network weights required for upsampling.

Inspired by \cite{Peng2020ECCV,Chibane_2020_CVPR}, we construct a continuous representation decoder
without the use of \ac{cbn}. This third architecture variant \emph{feature interpolation} performs bilinear interpolation on each
resolution of the 2D feature grid to obtain
a latent feature vector corresponding directly to the query position.
As this feature only contains information about the $xy$-position
we also concatenate the $z$-position of the query position onto this \emph{positioned} vector.
The resulting decoder structure contains almost the same number of parameters.
While the overall performance is comparable to the baseline, the accuracy in semantic mIoU declines.

\psection{Grid cell size (\tabref{tab:ablations_reduced}, cell size)}%
We review the impact of the architecture's grid cell size by scaling the base cell size of \SI{0.32}{\meter}
to \{\SI{75.0}{\percent}, \SI{81.25}{\percent}, \SI{150}{\percent}, \SI{200}{\percent}\} of
its original value.
The lower resolution feature maps as well as the input voxelization resolution are scaled accordingly.
Larger grid cells tend to have only a negligible impact on the large ground object classes.
However, semantic mIoU drops due to overall lower accuracy over all classes.

\psection{Loss weighting (\tabref{tab:ablations_reduced}, loss)}%
The individual loss weights $\lambda_{\{S,G,L\}}$ of the baseline network are $\lambda_S=7.5,\lambda_G=2.0,\lambda_C=1.0$.
We vary this weighting towards a larger contribution of the semantic loss, a larger contribution of the geometric loss, and a disabled consistency loss.
Reducing the semantic loss weight does help with geometric reconstruction accuracy.
However, the semantic segmentation accuracy does not improve over the baseline level
by a higher relative weighting.

\psection{Impact of semantic supervision signal on geometric completion quality (\tabref{tab:ablations_reduced}, Data})%
Previous work uses deep neural networks to perform geometric
scene completion both with and without semantic understanding of objects or scenes.
This choice primarily depends on the existence on semantic ground truth annotations.
Previous experiments suggest a correlation between semantic classification
of objects on the SUNGC dataset and the accuracy on geometric completion of
the scene \cite{Song2017CVPR}.
We investigate if the semantic supervision signal helps with understanding
objects in the scene and
therefore also with geometric reconstruction.

We compare our baseline model on the validation set with two models that are
trained on variants of the training dataset.
First, we map the 19 semantic classes of the semantic KITTI dataset to a
simpler set of only 9 classes. For instance, similar object classes are pooled
into categories for small and large traffic participants.
Secondly, we omit semantic classes altogether and only differentiate between \emph{occupied}
and \emph{free} while training.
Quantitative results are listed in \tabref{tab:ablations_reduced} grouped under \emph{Data}.
The performance on geometric completion is almost unaffected
by semantic supervision: \SI{57.6}{\percent} for the baseline vs.
\SI{57.9}{\percent} without semantic predictions.
It is still noteworthy that the seemingly more difficult task of
semantic scene completion is solved by a network of the same
size almost without a loss in geometric completion performance.
This suggests that the semantic and geometric completion task
are indeed related.

\psection{Impact of scene completion training data}%
\label{sec:abl_semseg}%
We analyze the single frame segmentation performance measured
by the mIoU over the segmentation of all input LiDAR points.
For this purpose, we train our method on single LiDAR scans with free space sampling and
compare it to the baseline trained for scene completion on accumulated data.
The networks are identical and the accumulated scene completion targets
are a super set of the semantic segmentation of a single LiDAR scan.
The segmentation performance of the scene completion model with accumulated
supervision is almost \SI{6}{\percent} lower than
that of the model only trained on single frame segmentation.
Smaller object classes see the strongest declines.
The quantitative results of this comparison and the differences in IoU scores are listed in the appendix in \tabref{tab:semantic_vs_completion}.

%


\section{Discussion}


It is noteworthy that we can compete on a benchmark
based on a voxelized representation
even though we do not use voxels as input or training targets.
The voxelized scene that we generate from a post-processing step is more
accurate than the end-to-end learned voxelization of other methods.
We believe that our continuous representation
benefits the learning of a spatially accurate scene representation.
Voxelization causes quantization noise in the input signal and the
supervision signal which we can avoid entirely.

\begin{figure}[tb]
    \definecolor{colorbarA}{RGB}{188,74,63}
    \definecolor{colorbarB}{RGB}{255,255,0}
    \definecolor{colorbarC}{RGB}{69,188,104}
    \centering{}
    \renewcommand{\arraystretch}{.5} 
    \setlength{\tabcolsep}{1pt} 
    
    \begin{tabularx}{\linewidth}{@{}XX@{}}
        \includegraphics[trim=0 100 0 100,clip,width=\linewidth]%
        {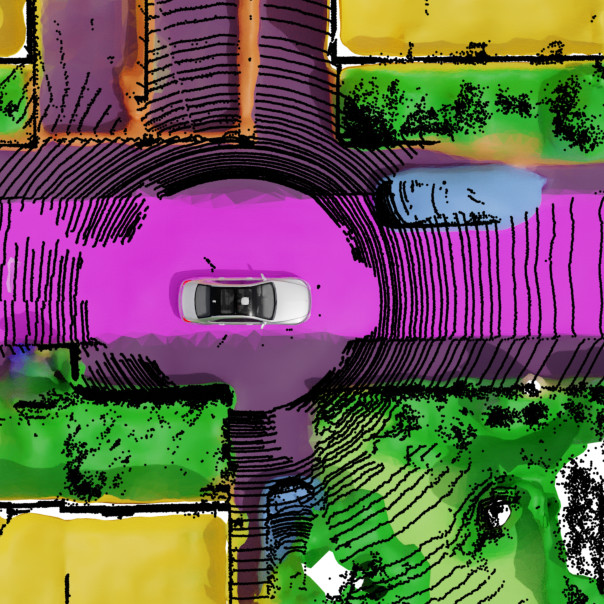} &
        \includegraphics[trim=0 213 0 213,clip,width=\linewidth]%
        {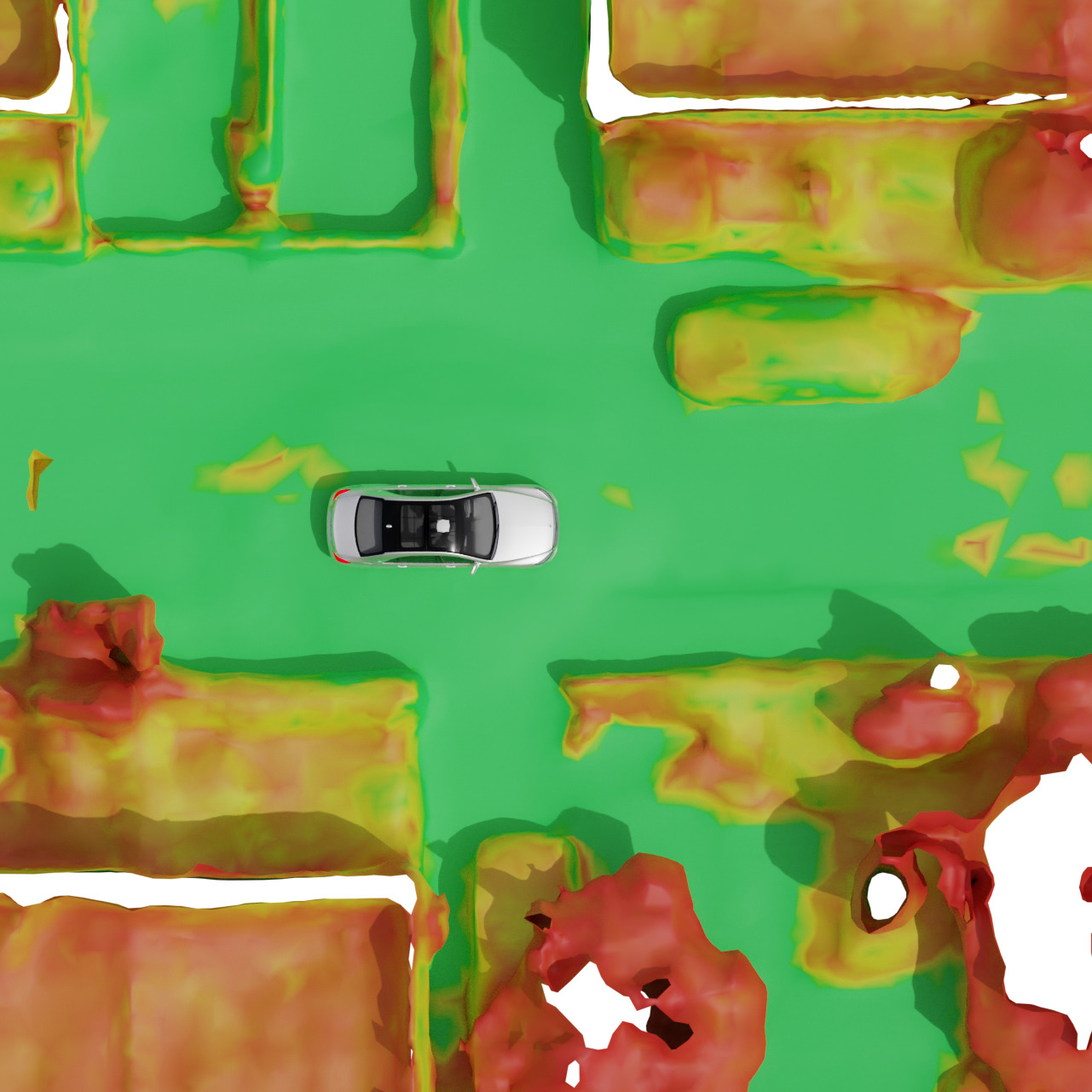}\\
        \includegraphics[trim=0 110 0 110,clip,width=\linewidth]%
        {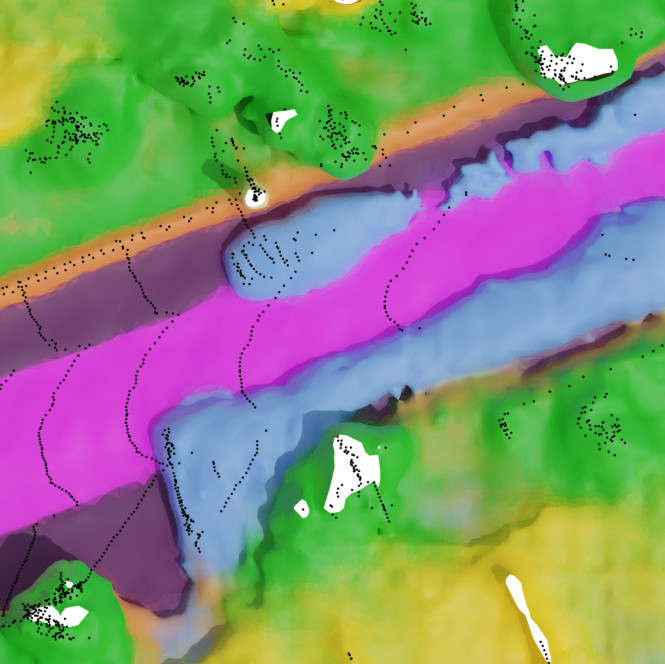} &
        \begin{tikzpicture}
            \node[anchor=south west,inner sep=0] (image) at (0,0) {\includegraphics[trim=0 114 0 114,clip,width=\linewidth]{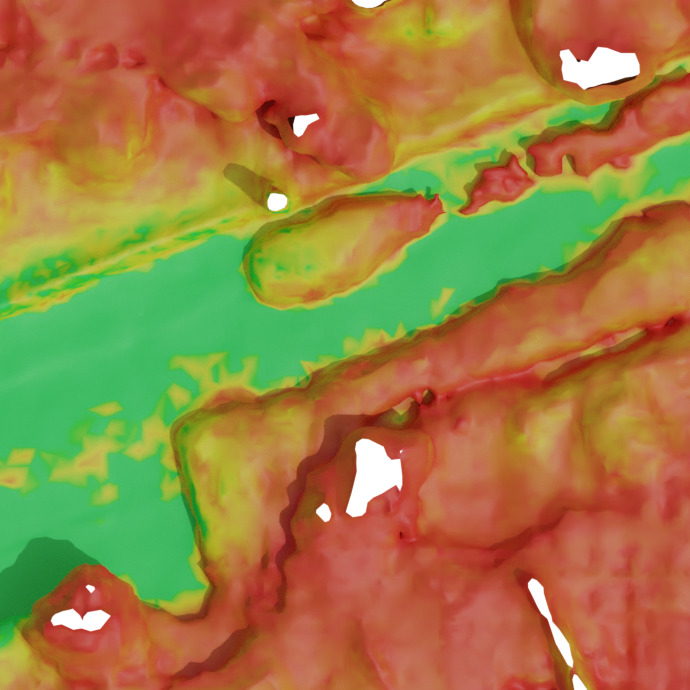}};
            \begin{scope}[x={(image.south east)},y={(image.north west)},shift={(.05,.4)}]
                \simplecolorbar{0.5}{.05}{colorbarA,colorbarB,colorbarC}{0.0}{3.5}{1.0}%
            \end{scope}
        \end{tikzpicture}\\%
    \end{tabularx}
    \caption{%
    Left: Top-view scene completion. Right: Magnitude of the gradient of the free space probability \wrt{} the surface normal. %
    The top row demonstrates a highly predictable completion of road surface,
    the boundaries of the sidewalk, and a car in proximity to the ego-vehicle.
    Far away from the ego-vehicle, the bottom row shows how our method guesses the most
    likely classification of each individual scene coordinate
    in the absence of almost all evidence from actual measurements.
    The scene completion function is \emph{softer} at these object boundaries (red surfaces).
    \label{fig:completion_uncertainty}%
    }
\end{figure}

We have analyzed how the generated scene completion function
behaves when confronted with sparse measurements.
The ground segmentation images illustrate that the representation
generalizes to areas that are never directly observed with the LiDAR sensor.
Our method learns to interpolate the course of the road, sidewalks
or parking areas between measurements.
\figref{fig:completion_uncertainty} highlights completion modes
for areas that are highly predictable (top row)
and areas where completion is based on a best guess from the
prior data distribution obtained from the training dataset (bottom row).
For the latter part we say that the \ac{DNN} completes the scene from experience
when presented with practically no evidence from measurements.
We examine another aspect of the scene completion function
and plot the results in the right column of \figref{fig:completion_uncertainty}.
As before, we create a mesh that approximates the decision surface of the
completion function at a certain threshold for the free space
probability.
In addition we determine the gradient of the free space value
\wrt{} to the surface normal.
The magnitude of this gradient is now transformed into a pseudo-color of the mesh.
With a larger magnitude the transition from free space
to an occupied class gets sharper.
It is clearly visible that the ground level has a sharp
transition even in high distances as it is easier to predict.
Smaller objects show generally smaller gradients at their surfaces.
But it is also noteworthy that the invisible rear side of objects
as well as the predicted clouds of parking-car-probabilities
have a small magnitude.
Meaning that there is a softer transition in the completion function.
It appears that the free space gradient correlates with the
certainty of the spatial position of a surface.
However, it can not be considered a well-calibrated measure
of uncertainty in the output, but probably more as an indication of such.

We identify a failure mode of our method when it comes
to the representation of fine geometric details and
the drop of single frame segmentation performance as analyzed in 
the ablation about completion vs. segmentation training data (\secref{sec:abl_semseg}, final paragraph).
This loss in segmentation performance is significant
given that the segmentation is
derived from the exact same input point cloud.
We do not have a definitive explanation for the magnitude of this circumstance.
A possibility is to attribute the drop
to the domination of the learning process by the completion task that
leads to poorer performance on the segmentation task.
An effect that can similarly be observed in multi-task learning setups.
Another effect that contributes is that
the completion task exhibits many ambiguities in the areas where
the input point cloud is sparse.
There predictions are dominated by the dataset prior where
small object classes are underrepresented.
The convolutional architecture shares all weights over the spatial scene extent
so that this kind of label noise contributes
to the blurring of smaller object classes.

\section{Conclusion and Future Work}

We presented a novel approach to predict a semantically
completed 3D scene from a single LiDAR scan.
Our method is able to infer 3D geometry and semantics in sparsely measured areas
from context and prior experience.
In doing so we address two essential challenges:
The first is to use LiDAR data and the included free space information as supervision
signal.
The second is being able to process large spatial extents for outdoor use while maintaining
a high spatial resolution of the predicted completion at the same time.
The key aspect is to encode LiDAR point clouds in a structured latent representation
that is then decoded using local deep implicit functions at multiple resolutions.
The output representation can be post-processed to obtain a voxel representation or 3D meshes
for visualization purposes.
We believe that we have set an important LiDAR-only baseline
in the emerging field of large-extent outdoor scene completion.

Our approach surpasses all other methods on the challenging
voxel-based Semantic~KITTI scene completion benchmark
in terms of geometric completion \acs{IoU} (+\SI{1.0}{\percent}).
The ablation experiments demonstrated the advantage of the
multi-resolution latent grid over a single
resolution and verify the selected hyper-parameters.
We showed that learning semantic classes along with geometry does
not induce a performance penalty on the geometric completion performance.
Uncertainty is inherent in the real-world scene completion task.
As future work it will be rewarding to address this uncertainty
by means of calibrating the network output or learning of a mapping to
uncertainty from the input data.
A well-calibrated uncertainty estimate will
help to take full advantage of learning-based scene completion.

\ifCLASSOPTIONcaptionsoff
  \newpage
\fi



%

\bibliographystyle{IEEEtran} 
\bibliography{conf_abrv_long,IEEEabrv,references}

%

\begin{IEEEbiography}[{\includegraphics[width=1in,height=1.25in,keepaspectratio,trim=0 .7in 0 .1in, clip]{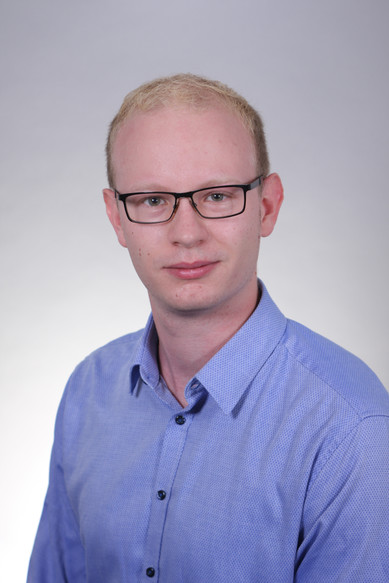}}]{Christoph B. Rist}
received his Bachelor’s degree in 2014 and his Master's degree in 2017,
both in Electrical Engineering and Information Technology
from Karlsruhe Institute of Technology, Germany.
He is currently pursuing his Ph.D. degree in the Intelligent Vehicles group at TU Delft (The Netherlands) while working in the Corporate Research of Mercedes-Benz AG in Stuttgart (Germany).
His current research focuses on LiDAR perception for autonomous driving.
\end{IEEEbiography}

\begin{IEEEbiography}[{\includegraphics[width=1in,height=1.25in,clip,keepaspectratio]{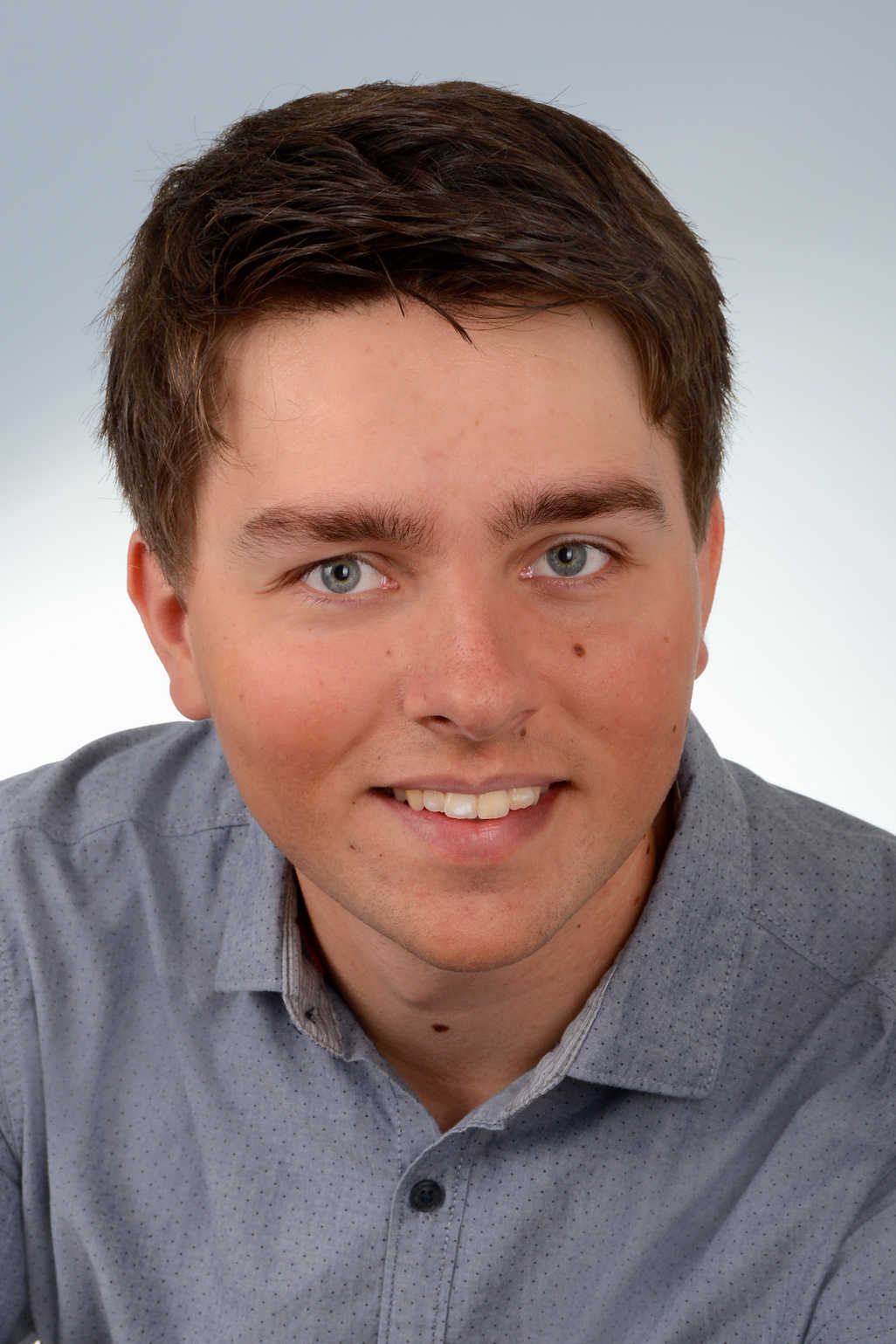}}]{David Emmerichs}
received his Bachelor’s degree in 2016 and his Master's degree in 2018 in Physics from the RWTH Aachen University, Germany.
He is currently pursuing his Ph.D. degree at IWR, Heidelberg University while working in the Corporate Research of Mercedes-Benz AG, Stuttgart (both Germany).
His current research focuses on LiDAR perception for autonomous driving.
\end{IEEEbiography}


\begin{IEEEbiography}[{\includegraphics[width=1in,height=1.25in,clip,keepaspectratio]{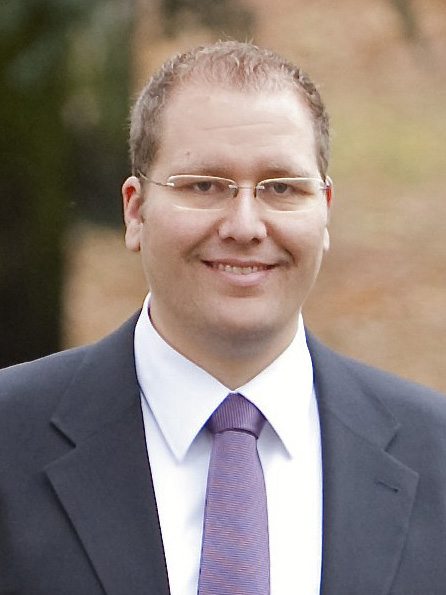}}]{Markus Enzweiler}
received the Ph.D. degree in computer science
from the Univ. of Heidelberg, Germany (2011). From 2010, he was with Mercedes-Benz AG R\&D in Stuttgart, Germany, most recently as a Technical Manager for LiDAR and camera. He co-developed
the Daimler vision-based pedestrian detection system which is available in
Mercedes-Benz cars. In 2021, he moved to Esslingen University of Applied Sciences as a Full Professor for Autonomous Mobile Systems. His current research focuses on scene understanding for mobile robotics. 
In 2012, he received
the IEEE Intelligent Transportation Systems Society Best PhD
Dissertation Award and the Uni-DAS Research Award for his
work on vision-based pedestrian recognition. He is part of the
team that won the 2014 IEEE Intelligent Transportation Systems
Outstanding Application Award. In 2014, he was honored with a
Junior-Fellowship of the Gesellschaft für Informatik.
\end{IEEEbiography}

\begin{IEEEbiography}
[{\includegraphics[width=1in,height=1.25in,clip,keepaspectratio]{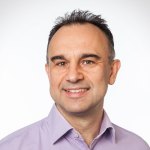}}]{Dariu M. Gavrila}
received the Ph.D. degree in computer science from the Univ. of Maryland at College Park, USA, in 1996. From 1997, he was with Daimler R\&D, Ulm, Germany, where he became a Distinguished Scientist. He led the vision-based pedestrian detection research, which was commercialized 2013-2014 in various  Mercedes-Benz models. In 2016, he moved to TU Delft (The Netherlands), where he since heads the Intelligent Vehicles group as Full Professor. His research deals with sensor-based detection of humans and analysis of behavior, recently in the context of the self-driving cars in urban traffic. He received the Outstanding Application Award 2014 and the Outstanding Researcher Award 2019, both from the IEEE Intelligent Transportation Systems Society.  
\end{IEEEbiography}





%


\appendices



\section{Neural network architecture, implementation and hyperparameters\label{sec:arch}}

\tabref{tab:network_architecture} lists all layers, inputs and operations of
our \ac{DNN} architecture.
\tabref{tab:hyperparameters} lists hyperparameters for the latent grid size, 
network training, and inference.
We use TensorFlow to implement online data processing, neural network weight optimization, and network inference.
The Adam optimizer is used for optimization.
We use linear learning rate warm-up
over the first $ 2*(1-\beta_2)^{-1} $ training iterations as proposed by \cite{DBLP:journals/corr/abs-1910-04209}
for untuned warmup of Adam's learning rate.
%
%

\newcommand{\widthFirstColumn}{2.8cm}
\definecolor{LightCyan}{rgb}{0.88,1,1}
\definecolor{LightA}{rgb}{1,0.88,1}
\definecolor{LightB}{rgb}{1,1,0.88}
\begin{table*}[p]
\centering
\setlength{\extrarowheight}{0pt} 
\caption{\textbf{Network architecture:} Detailed network architecture and input format definition.
The ID of each row is used to reference the output of the row.
$\uparrow$ indicates that the layer immediately above is an input.
$N$ denotes the number of LiDAR points falling within the $xy$-grid extent.
$M$ denotes the number of input cells in the $x$ and $y$ direction. We are assuming a square $N=M$ $xy$-extent.
$O$ denotes the number of positions to classify.%
\label{tab:network_architecture}
}
\resizebox{.90\linewidth}{!}{%
    \setlength{\tabcolsep}{5pt}
    \setlength{\fboxsep}{0pt}%
\begin{tabular}{@{}rlllp{7.0cm}@{}}
\toprule
\textbf{ID} & \textbf{Inputs} & \textbf{Layer/operation} & \textbf{Output shape} & \textbf{Parameters/output description} \\
\midrule
\multicolumn{5}{c}{\textbf{Input features from voxelized LiDAR scan}} \\ \midrule{}
1 & LiDAR & $x_v, y_v, z_v$ & $[L\times3]$ & Position of each point in voxel coordinates\\
2 & LiDAR & $x_m, y_m, z_m$ & $[L\times3]$ & Position of each point relative to voxel mean\\
3 & LiDAR & $r$             & $[L]$ & Reflectivity value of each point\\
4 & LiDAR & $d$             & $[L]$ & $xy$-distance from sensor of each point\\
5 & 1,2,3,4 & Concat features & $[L\times8]$ & \\\midrule
\multicolumn{5}{c}{\textbf{Point feature extractor}} \\ \midrule{}
6 & $\uparrow$ & Dense layer & $[L\times128]$ & Pointwise-dense features. Units: 128 \\
7 & $\uparrow$ & Batch Norm & \dittotikz & \\
8 & $\uparrow$ & ReLU & \dittotikz & \\
9 & $\uparrow$ & ReduceMax & $[M\!\times\! M\!\times\! 128]$ & Maximum over point features within each voxel\\ \midrule
\multicolumn{5}{c}{\textbf{CNN feature encoder}} \\ \midrule
10 & $\uparrow$ & (Conv+BN+ReLU) x2 & $[M\!\times\! M\!\times\! 128]$ & Kernel size $3\!\times\! 3$, stride 1 \\
11 & $\uparrow$ & Conv+BN+ReLU & $[M/2\!\times\! M/2\!\times\! 128]$ & Kernel size $3\!\times\! 3$, stride 2 \\
12 & $\uparrow$ & (Conv+BN+ReLU) x3 & \dittotikz & Kernel size $3\!\times\! 3$, stride 1 \\
13 & $\uparrow$ & Conv+BN+ReLU & $[M/4\!\times\! M/4\!\times\! 256]$ & Kernel size $3\!\times\! 3$, stride 2 \\
14 & $\uparrow$ & (Conv+BN+ReLU) x5 & \dittotikz & Kernel size $3\!\times\! 3$, stride 1 \\
15 & $\uparrow$ & Conv+BN+ReLU & $[M/8\!\times\! M/8\!\times\! 256]$ & Kernel size $3\!\times\! 3$, stride 2 \\
16 & $\uparrow$ & (Conv+BN+ReLU) x5 & \dittotikz & Kernel size $3\!\times\! 3$, stride 1 \\
17 & $\uparrow$ & Conv+BN+ReLU & $[M/32\!\times\! M/32\!\times\! 64]$ & Kernel size $3\!\times\! 3$, stride 4 \\
18 & $\uparrow$ & (Conv+BN+ReLU) x2 & \dittotikz & Kernel size $3\!\times\! 3$, stride 1 \\
19 & $\uparrow$ & Deconv+BN+ReLU & $[M/8\!\times\! M/8\!\times\! 64]$ & Kernel size $4\!\times\! 4$, stride 4 \\
20 & $\uparrow$, 16 & Concat features & $[M/8\!\times\! M/8\!\times\! 320]$ &  \\
21 & $\uparrow$ & Deconv+BN+ReLU & $[M/2\!\times\! M/2\!\times\! 256]$ & Kernel size $4\!\times\! 4$, stride 4 \\
22 & 14         & Deconv+BN+ReLU & $[M/2\!\times\! M/2\!\times\! 256]$ & Kernel size $2\!\times\! 2$, stride 2 \\
23 & 9          & Conv+ReLU & $[M/2\!\times\! M/2\!\times\! 64]$ & Kernel size $3\!\times\! 3$, stride 2 \\
24 & 12,21,22,23  & Concat features & $[M/2\!\times\! M/2\!\times\! 704]$ & Also includes skip connection from 12\\
25 & 18 & Conv+ReLU & $[M/32\!\times\! M/32\!\times\! 256]$ & Kernel size $2\!\times\! 2$, stride 1 (feature map $\bc_1$) \\
26 & 20 & Conv+ReLU & $[M/8\!\times\! M/8\!\times\! 256]$ & Kernel size $2\!\times\! 2$, stride 1 (feature map $\bc_2$) \\
27 & 24 & Conv+ReLU & $[M/2\!\times\! M/2\!\times\! 128]$ & Kernel size $2\!\times\! 2$, stride 1 (feature map $\bc_3$) \\ \midrule
\multicolumn{5}{c}{\textbf{Latent feature selection}} \\ \midrule
28 & -              & Query positions    & $[O\times 3]$            & Positions to be classified by the decoder \\
29 & 28             & Cell assignment   & $[O\times 2]$            & Assign cells to each query position \\
30 & 28             & Cell assignment   & \dittotikz            & Assign cells to each query position \\
31 & 28             & Cell assignment   & \dittotikz            & Assign cells to each query position \\
32 & 18, 29     & Gather features $\bc_1$  & $[O\times 256]$     & Features of hierarchy 1 \\
33 & 20, 30     & Gather features $\bc_2$  & $[O\times 256]$    & Features of hierarchy 2 \\
34 & 24, 31     & Gather features $\bc_3$  & $[O\times 128]$    & Features of hierarchy 3 \\
35 & 32, 33     & Concat $[\bc_1, \bc_2]$         & $[O\times 512]$    & Concat features of hierarchy 1, 2  \\
36 & 32, 33, 34 & Concat $[\bc_1, \bc_2, \bc_3]$  & $[O\times 640]$    & Concat features of hierarchy 1, 2, 3  \\
37 & 28, 29     & Calc. position $\bp_1$  & $[O\times 3]$  & Query position in local cell-relative coordinate system\\
38 & 28, 30     & Calc. position $\bp_2$  & \dittotikz     & Query position in local cell-relative coordinate system\\
39 & 28, 31     & Calc. position $\bp_3$  & \dittotikz     & Query position in local cell-relative coordinate system\\ \midrule
\multicolumn{5}{c}{\textbf{Decoder with conditioned batch-normalization}} \\ \midrule
40 & 32              & Dense            & $[O\times 64]$        & Conditioning vector. Units: $2 * 32$ as mean $\mu$, var $\sigma$\\
41 & 35              & Dense            & \dittotikz         & Conditioning vector. Units: $2 * 32$ as mean $\mu$, var $\sigma$\\
42 & 36              & Dense            & \dittotikz         & Conditioning vector. Units: $2 * 32$ as mean $\mu$, var $\sigma$\\
43 & 37              & Dense+BN         & $[O\times 32]$        &    \\
44 & $\uparrow$, 40  & Conditioning     & $[O\times32]$         & Apply feature-wise mean and var $\by = \sigma\odot\bx+\mu $  \\
45 & $\uparrow$      & ReLU             & $[O\times32]$         &    \\
46 & $\uparrow$, 38  & Concat           & $[O\times 35]$        &    \\
47 & $\uparrow$      & Dense+BN         & $[O\times 32]$        &    \\
48 & $\uparrow$, 41  & Conditioning     & $[O\times32]$         & Apply feature-wise mean and var $\by = \sigma\odot\bx+\mu $  \\
49 & $\uparrow$      & ReLU             & $[O\times32]$         &    \\
50 & $\uparrow$, 39  & Concat           & $[O\times 35]$        &    \\
51 & $\uparrow$      & Dense+BN         & $[O\times 32]$        &    \\
52 & $\uparrow$, 42  & Conditioning     & $[O\times32]$         & Apply feature-wise mean and var $\by = \sigma\odot\bx+\mu $  \\
53 & $\uparrow$      & ReLU             & $[O\times32]$         &    \\
54 & $\uparrow$      & (Dense+ReLU) x2  & $[O\times 32]$        &    \\
55 &  $\uparrow$      & Dense            & $[O\times 20]$        & Output logits vector $[z_1,\dots{},z_n]$ of single local function $f_{L}$ \\
\bottomrule
\end{tabular}
}
\end{table*}

\begin{table*}[htbp]
\centering
\caption{\textbf{Hyperparameters}}
\label{tab:hyperparameters}
\begin{tabular}{llp{11cm}}
\toprule
\textbf{Parameter}    &   \textbf{Value}    &    \textbf{Description}           \\
\midrule
\multicolumn{3}{c}{\textbf{Grid setup and spatial augmentations}} \\ \midrule{}
$\delta$                    & \SI{0.32}{\meter}                             & Edge length of highest resolution hierarchy cells. 
Lower resolutions have edge lengths of $4\delta=\SI{1.28}{\meter}$ and $16\delta=\SI{5.12}{\meter}$.\\
$\delta / 2$                & \SI{0.16}{\meter}                             & Edge length of LiDAR input cells.
The ratio $1/2$ between LiDAR input cells and output cells is determined by the network architecture (\tabref{tab:network_architecture}).\\
$M_{\text{training}}\times M_{\text{training}}$ & $\SI{256}{} \times \SI{256}{} $  & Number of input cells when training.
Equals a spatial extent of \SI{40.96}{\meter}. This extent is randomly translated to off-center positions. \\
$\sigma$ & \SI{8}{\meter} & Standard deviation of normally-distributed grid offsets \\
$a_{\text{transl}}, b_{\text{transl}}$ & $\pm\SI{0.05}{\meter}$ & Boundaries of random uniform translation of input point cloud  \\
$a_{\text{scale}}, b_{\text{scale}}$ & $\pm\SI{5}{\percent}$ & Boundaries of random uniform scaling of input point cloud  \\
\midrule
\multicolumn{3}{c}{\textbf{Loss and optimizer}} \\ \midrule{}%
$\lambda_S$                 & 7.5           & Constant weighting of semantic loss $\cL{}_S$. \\
$\lambda_G$                 & 2.0           & Constant weighting of geometric loss $\cL{}_G$. \\
$\lambda_C$                 & 1.0           & Constant weighting of consistency loss $\cL{}_C$. \\
$B$                         & $ 8 = (2 \times 4\;\text{GPUs})$                        & Training batch size                           \\
l                         & \num{1.0e-3}                                  & Initial learning rate                \\
$d_\text{step}$, $d_\text{rate}$  & \SI{40}{\kilo\nothing}; \num{.5}      & Staircase learning rate decay    \\
$w_\text{warmup}$, $w_\text{rate}$  & $ 2*(1-\beta_2)^{-1} = 2000 $           & Learning rate warm-up steps    \\
Adam $\beta_1$, $\beta_2$   & \num{0.9}; \num{0.999}                         & Adam optimizer momenta                \\
\midrule
\multicolumn{3}{c}{\textbf{Inference}} \\ \midrule{}%
$\theta_{\text{free space}}$     & \num{0.3}  & Visualization meshes use the isosurface at free space probability $\theta_{\text{free space}}$.\\
$\theta_{\text{empty voxel}}$ \methodname{}     & \num{0.04}  & Empty voxel threshold\\
$\theta_{\text{empty voxel}}$ \scssnetUpsample{}     & \num{0.05}  & Empty voxel threshold for upsampling variant\\
$\theta_{\text{empty voxel}}$ \scssnetSingle{}     & \num{0.04}  & Empty voxel threshold for upsampling variant\\
$\theta_{\text{empty voxel}}$ Feature Interpolation     & \num{0.04}  & Empty voxel threshold for feature interpolation variant\\
\bottomrule
\end{tabular}
\end{table*}

\section{Stable Formulation of the Geometric and Consistency Loss Terms\label{sec:loss}}

Our network head outputs a probability vector
$\bff^\bc_{\text{\nn{}}}(\bp) = [f_1,\dots,f_N,f_{N+1}]^{\intercal}$ for a
query position $\bp$. The first $N$
entries are defined to represent semantic object classes while the last entry $f_{N+1}$
is defined to represent the additional free space class.
These probabilities are obtained by softmax-normalization of the network output logits.
The following sections explain how numerical stability issues prevent us from
calculating the
cross-entropy loss terms from the softmax output.
However, the paper has so far derived the definitions of geometric and consistency loss
only from the softmax probabilities.
Therefore, the following sections deduce stable formulations
for the geometric and consistency loss from the unscaled output logits
as required to calculate the loss and gradients for training.

\subsection{Stable Cross-Entropy}

Probability-like vectors are computed from the unscaled network logits $z_i$
by applying exponential normalization ("softmax").
In order to calculate the cross-entropy loss from logits $z_i$ and true probabilities
$y_i$
\begin{align}
    \label{eq:ce_with_logsoftmax}\cL &= \sum_i y_i\,\log f_i = \sum_i y_i\, \text{logsoftmax}(z_i)
\end{align}
the logsoftmax function is required.
It is common and necessary for implementations of \eqnref{eq:ce_with_logsoftmax} to take advantage of the identity
$\text{softmax}(\bz) = \text{softmax}(\bz + c)$ to allow for the equivalent stable formulation
\begin{align}
\label{eq:logsoftmax_normalization}\text{logsoftmax}(z_i) &= \log\frac{\exp(z_i)}{\sum_j \exp(z_j)} \\
\label{eq:logsoftmax}&= z_i - b - \log \sum_j \exp(z_j - b)\\
&\hphantom{=}\quad \text{with}\;\; b = \max{(\bz)}\nonumber
\end{align}
to avoid both numerical under- or overflow issues that would otherwise arise in the formulation $\log(\exp(.))$.

\subsection{Geometric Loss\label{sec:geometric_loss}}%
The geometric loss loss
differentiates between the binary \textit{occupied} label of a given point.
The geometric loss is formulated separately from the semantic loss
to also include LiDAR measurements without a semantic annotation.
The geometric loss $\cL{}_G$ makes it necessary to conceptually sum up
the output
probabilities $f_1, \dots{},f_{N}$
of all the semantic object classes (all classes except for the
free space class).
It is straightforward to formulate the cross-entropy loss $\cL{}_G$ 
from the complete probability vector $\bff \in \nR_{N+1}$:
\begin{align}
\label{eq:reconstructionAPPENDIX}
\cL{}_G &= y_{\text{occupied}} \log (f_{\text{occupied}}) + (1-y_{\text{occupied}}) \log (f_{\text{free}})\\
\label{eq:gl}&= y_{\text{occupied}} \log\left(\sum_{i=1}^{N}f_{i}\right) + (1-y_{\text{occupied}}) \log (f_{N+1})
\end{align}
However, the fused formulation of logsoftmax (\eqnref{eq:logsoftmax}) prevents us from just adding up the softmax outputs $\sum_{i}^{N}f_{i}$
and inserting them into \eqnref{eq:gl}.
Instead, we use the same normalization trick as in \eqnref{eq:logsoftmax} to define free space and occupied logits as %
\begin{align}
    z_{\text{occupied}} &= b_{\text{occ}} + \log\sum_{i=1}^{N} \exp(z_i - b_{\text{occ}}) \\
    & \quad \text{with}\;\; b_{\text{occ}} = \max(z_1,\dots{},z_{N}) \\
    z_{\text{free}} &= z_{N+1}
\end{align}
from the outputs $z_1, \dots{},z_{N+1}$ so that the equality
\begin{align}
    \text{softmax}\begin{pmatrix}z_{\text{occupied}} \\ z_{\text{free}} \end{pmatrix}
    &= \begin{pmatrix}\sum_{i=1}^{N}f_{i} \\ f_{N+1}\end{pmatrix}
\end{align}
holds.
Thus, the converted logits $z_{\text{free}}$ and $z_{\text{occupied}}$
can be used for the computation of the geometric loss as in \eqnref{eq:ce_with_logsoftmax}.

\subsection{Consistency Loss\label{sec:consistency_loss}}%
The same numerical stability issue needs to be considered when implementing the consistency loss.
We can write the consistency loss $\cL{}_{C}$ as Jensen-Shannon-Divergence (JSD)
between $M$ probability vectors
$\bP{}^{(1)}, \dots, \bP{}^{(M)}$:
\begin{align}
    \bP^{(\alpha)} &= [p_1^{(\alpha)},\dots{},p_n^{(\alpha)}] \in \nR^n\\
    \cL{}_{C} & = \mathrm{JSD}\left(\bP{}^{(1)}, \dots, \bP{}^{(M)}\right)\\
     \label{eq:jsd2}& = H\left(\overline{\bP}\right)
     - \frac1{M} \sum_{\alpha=1}^M H\left(\bP{}^{(\alpha)}\right)\\
     &\text{with}\quad\overline{\bP} = (\overline{p}_1,\dots{},\overline{p}_n) = \frac{1}{M} \sum_{\alpha=1}^{M}\bP{}^{(\alpha)}\nonumber\\
     \label{eq:jsd_written_out}& = \left(-\sum_{i=1}^{n}\left(\overline{p}_i\right) \log \left(\overline{p}_i\right)\right)
         + \frac1{M} \sum_{\alpha=1}^M  \sum_{i=1}^n p_{i}^{(\alpha)} \log p_{i}^{(\alpha)}
\end{align}
The loss $\cL{}_{C}$ equals the difference between the entropy of the averaged probability distribution
$H\left(\overline{\bP}\right)$ and the average of
the individual entropies $\frac1{M} \sum_{\alpha=1}^M H\left(\bP{}^{(\alpha)}\right)$.
As in \eqnref{eq:ce_with_logsoftmax},
this term needs to be calculated from the unscaled output logits $z_{i}^{(\alpha)}$.
The probabilities
\begin{align}
p_{i}^{(\alpha)} &= \frac{\exp(z_{i}^{(i)})}{\sum_{j=1}^{n} \exp (z_{j}^{(\alpha)})}
\end{align}
are again deduced from the logits $z_{i}^{(\alpha)}$ by softmax normalization.
The second term %
of \eqnref{eq:jsd_written_out}
is calculated using the definition of the stable logsoftmax function
in \eqnref{eq:logsoftmax}.
Special consideration is necessary to compute 
the logarithm of averaged probabilities $\log\overline{p}_i$
in the first term in \eqnref{eq:jsd_written_out}.
For a stable computation of $\log\overline{p}_i$ from logits $z_{i}^{(\alpha)}$ we are going to use the definitions
\begin{align}
    \hat{z}^{(\alpha)} &= \max\left(z_{1}^{(\alpha)}, \dots{},z_{n}^{(\alpha)}\right) \\
    \hat{z}_{i} &= \max\left(z_{i}^{(1)} - \hat{z}^{(1)}, \dots{},z_{i}^{(M)} - \hat{z}^{(M)}\right) \\
    s^{(\alpha)} &= \sum_{i=1}^{n} \exp \left(z_{i}^{(\alpha)} - \hat{z}^{(\alpha)}\right)
\end{align}
of maximum vectors $\hat{z}_{1},\dots,\hat{z}_{n}$ and $\hat{z}^{(1)},\dots,\hat{z}^{(M)}$
and $s^{(\alpha)}$ as shorthand notation for the softmax denominator of the shifted logits.
$\hat{z}^{(\alpha)}$ shifts the softmax calculation of each individual output logit vector.
Thereafter, $\hat{z}_{i}$ shifts the computation of the exponential over the given vector entry $i$.
Again, this ensures that the argument of $\log(\exp(.))$ is within an acceptable range.
Thus, $\log\overline{p}_i$ is written as
\begin{align}
    \log \overline{p}_i &= \log \left(\frac1{M}\sum_{\alpha=1}^M p_i^{(\alpha)}\right)\\
    &= \log \left(\frac1{M}\sum_{\alpha=1}^M \frac{\exp(z_{i}^{(\alpha)})}{\sum_{j=1}^{n} \exp (z_{j}^{(\alpha)})}\right) \\
    &= \log \left(\sum_{\alpha=1}^M \frac{\exp(z_{i}^{(\alpha)} - \hat{z}^{(\alpha)})}{s^{(\alpha)}}\right) - \log (M) \\
    &= \log \left(\sum_{\alpha=1}^M \left(\prod_{\beta=1,\beta\neq \alpha}^{M} s^{(\beta)}\right) \exp(z_{i}^{(\alpha)} -\hat{z}^{(\alpha)})\right) \nonumber\\
    &\hphantom{=}- \log \prod_{\alpha=1}^{M} s^{(\alpha)} - \log (M) \\
    \label{eq:consistency_result} &= \hat{z}_{i} + \log\left( \sum_{\alpha=1}^{M}  \left(\prod_{\beta=1,\beta\neq \alpha}^{M} s^{(\beta)}\right)
      \exp \left(z_{i}^{(\alpha)} - \hat{z}^{(\alpha)} - \hat{z}_{i} \right)
      \right) \nonumber\\
      &\hphantom{=}- \log \prod_{\alpha=1}^{M} s^{(\alpha)} - \log (M)
\end{align}
to gain a formulation that is stable against over or underflow. We insert \eqnref{eq:consistency_result} into
\eqnref{eq:jsd_written_out} and implement the resulting statement as our consistency loss function.



\setlength{\tabcolsep}{1.7pt} 

\begin{table*}[tbp]
    \definecolor{colhighlight}{gray}{0.9}
    \newcolumntype{H}{>{\columncolor{colhighlight}}r}
%
    \caption{Quantitative results of baseline and ablations on the validation set (higher is better)}
    \label{tab:ablations_full}
    \centering
    \begin{tabularx}{\linewidth}{@{}clHrr|Hrrrrrrrrrrrrrrrrrrr@{}}
        \toprule
        & &
        \multicolumn{3}{@{}c@{}}{\textbf{Geometric Completion}} &
        \multicolumn{20}{@{}c@{}}{\textbf{Semantic Completion}} \\
        & \textbf{Variation} & \rotsem{\textbf{IoU Occ.}} & \rotsem{Precision} & \rotsem{Recall} & 
        \rotsem{\textbf{mIoU}} &
        \rotsem{\semcolor[road] Road} &
        \rotsem{\semcolor[sidewalk] Sidewalk} &
        \rotsem{\semcolor[parking] Parking} &
        \rotsem{\semcolor[otherground] other gr.} &
        \rotsem{\semcolor[building] Building} &
        \rotsem{\semcolor[fence] Fence} &
        \rotsem{\semcolor[car] Car} &
        \rotsem{\semcolor[truck] Truck} &
        \rotsem{\semcolor[othervehicle] other veh.} &
        \rotsem{\semcolor[bicycle] Bicycle} &
        \rotsem{\semcolor[motorcycle] Motorcycle} &
        \rotsem{\semcolor[person] Person} &
        \rotsem{\semcolor[bicyclist] Bicyclist} &
        \rotsem{\semcolor[motorcyclist] Motorcycl.} &
        \rotsem{\semcolor[vegetation] Vegetation} &
        \rotsem{\semcolor[trunk] Trunk} &
        \rotsem{\semcolor[terrain] Terrain} &
        \rotsem{\semcolor[pole] Pole} &
        \rotsem{\semcolor[trafficsign] Tr. Sign} %
        \\ \midrule
        & \makecell[cl]{\methodname{} (Baseline)} & %
        57.8 & 73.1 & 73.4 & %
        \textbf{26.1}%
        & \textbf{71.2} & \textbf{43.8} & 31.8 & \phantom{0}\textbf{3.3} & 38.6 & \textbf{13.6} & 51.3 & \textbf{32.3} %
        & \textbf{10.6} & \phantom{0}4.3 & \phantom{0}3.3 & \textbf{15.7} & 24.7 & \phantom{0}0.0 %
        & 40.1 & 19.6 & 50.6 & 25.7 & 14.0 %
        \\
        & \makecell[cl]{(\methodname{} + \acs{TTA})} & %
        (58.5) & 74.2 & 73.5 & %
        (26.9)%
        & 72.2 & 45.4 & 36.0 & \phantom{0}2.8 & 39.2 & 14.7 & 52.8 & 33.2 %
        & 9.3 & 4.1 & 4.4 & 16.4 & 24.9 & \phantom{0}0.0 %
        & 41.8 & 19.7 & 52.1 & 26.3 & 15.0 %
        \\ \midrule{}
        \multirow{3}{*}{\rotatebox[origin=c]{90}{\textbf{Arch.}}} %
        & \makecell[cl]{\scssnetUpsample{}} &
        55.4 & 71.9 & 70.8 & %
        23.8 %
        & 63.5 & 37.1 & 30.6 & \phantom{0}0.1 & 37.1 & 13.5 & 48.5 & 22.6 %
        & \phantom{0}7.5 & \phantom{0}\textbf{7.6} & \phantom{0}2.6 & 15.1 & 24.9 & \phantom{0}0.0 %
        & 39.4 & 18.5 & 45.0 & 24.9 & 13.3 %
        \\ %
        & \makecell[cl]{\scssnetSingle{}} &
        57.1 & 72.7 & 72.6 & %
        24.2 %
        & 69.6 & 42.9 & \textbf{33.7} & \phantom{0}2.5 & 37.3 & 12.7 & 50.0 & 27.3 %
        & \phantom{0}5.9 & \phantom{0}2.5 & \phantom{0}3.5 & 13.1 & 11.9 & \phantom{0}0.0 %
        & 40.6 & 18.6 & 49.5 & 23.8 & 14.7 %
        \\ %
        & \makecell[cl]{Feature interpolation} &
        57.4 & 73.0 & 73.0 & %
        25.5 %
        & 68.5 & 43.4 & 30.0 & \phantom{0}2.4 & \textbf{39.4} & 12.9 & 51.4 & 23.3 %
        & 10.5 & \phantom{0}2.7 & \phantom{0}2.6 & 14.7 & \textbf{34.9} & \phantom{0}0.0 %
        & 40.6 & 19.1 & 47.9 & 26.3 & 14.3 %
        \\ %
        \midrule{} %
        \multirow{4}{*}{\rotatebox[origin=c]{90}{\textbf{Cell size}}} %
        & \makecell[cl]{Cell size \SI{75.0}{\percent}} &
        54.1 & 70.6 & 69.8 & %
        23.8 %
        & 62.0 & 39.3 & 27.3 & \phantom{0}1.7 & 37.3 & 14.4 & 50.2 & 30.4 %
        & 13.0 & \phantom{0}5.1 & \phantom{0}3.4 & 13.3 & 16.1 & \phantom{0}0.0 %
        & 39.3 & 16.6 & 43.0 & 24.8 & 14.7 %
        \\ %
        & \makecell[cl]{Cell size \SI{87.5}{\percent}} &
        57.1 & 73.8 & 71.7 & %
        25.6
        & 68.3 & 42.0 & 33.2 & \phantom{0}1.6 & 38.2 & 13.9 & 51.4 & 29.6 %
        & 9.8 & \phantom{0}3.3 & \phantom{0}\textbf{5.3} & 14.0 & 25.2 & \phantom{0}0.0 %
        & \textbf{41.1} & 18.6 & 49.0 & \textbf{26.7} & 14.9 %
        \\ %
        & \makecell[cl]{Cell size \SI{150}{\percent}} &
        56.7 & 71.6 & 73.1 & %
        24.1 %
        & 69.0 & 41.1 & 30.1 & \phantom{0}1.3 & 37.8 & 11.6 & 49.6 & 31.7 %
        & 5.6 & \phantom{0}1.9 & \phantom{0}2.2 & 10.4 & 22.2 & \phantom{0}0.0 %
        & 40.2 & 17.3 & 48.8 & 24.0 & 13.2 %
        \\ %
        & \makecell[cl]{Cell size \SI{200}{\percent}} &
        56.7 & 72.5 & 72.3 & %
        23.3
        & 71.0 & 42.4 & 25.8 & \phantom{0}1.0 & 37.5 & \phantom{0}8.5 & 48.4 & 22.0 %
        & 12.8 & \phantom{0}2.3 & \phantom{0}2.8 & 2.7 & 20.9 & \phantom{0}0.0 %
        & 39.6 & 16.6 & \textbf{51.3} & 23.2 & 13.4 %
        \\ %
        \midrule{} %
        \multirow{3}{*}{\rotatebox[origin=c]{90}{\textbf{Loss}}} %
        & \makecell[cl]{$\lambda_S=15,\lambda_G=1$} &
        55.6 & 71.1 & 71.7 & %
        24.0 %
        & 68.4 & 42.3 & 29.8 & \phantom{0}2.3 & 37.9 & 12.9 & 49.6 & 29.1 %
        & \phantom{0}7.2 & \phantom{0}1.4 & \phantom{0}2.0 & 14.3 & 15.6 & \phantom{0}0.0 %
        & 39.0 & 18.7 & 46.7 & 25.3 & 13.7 %
        \\ %
        & \makecell[cl]{$\lambda_S=3.75,\lambda_G=4$} &
        \textbf{58.2} & 74.5 & 72.7 & %
        24.7 %
        & 70.3 & 42.7 & 28.8 & \phantom{0}0.6 & 38.4 & 13.0 & \textbf{51.9} & 23.2 %
        & 11.9 & \phantom{0}2.9 & \phantom{0}6.1 & 12.3 & 16.9 & \phantom{0}0.0 %
        & 40.9 & \textbf{20.0} & 48.5 & 26.1 & \textbf{15.6} %
        \\ %
        & \makecell[cl]{$\lambda_C=0$} &
        56.9 & 72.0 & 73.0 & %
        25.0 %
        & 67.7 & 40.5 & 30.2 & \phantom{0}0.5 & 38.0 & 13.3 & 50.8 & 25.9 %
        & \phantom{0}7.8 & \phantom{0}5.4 & \phantom{0}2.3 & \textbf{15.7} & 28.1 & \phantom{0}0.0 %
        & 40.7 & 18.8 & 47.6 & 26.4 & 14.9 %
        \\ %
        \midrule{} %
        \multirow{2}{*}{\rotatebox[origin=c]{90}{\textbf{Data}}} &
        \makecell[cl]{Simplified sem.} &
        57.8 & 74.1 & 72.3 & %
        (38.8)%
        & 70.0 & \clsrule{3}{.54cm}{45.0} & \clsrule{2}{.22cm}{38.0} & 51.9 %
        & \clsrule{2}{.27cm}{30.2} & \clsrule{5}{1.08cm}{11.8} & 40.5 & \clsrule{3}{.51cm}{46.7} & 15.3 %
        \\
        & \makecell[cl]{Without sem.} &
        57.9 & 73.6 & 73.1 & %
        (57.9) %
        & \clsrule{19}{5.4cm}{57.9} %
        \\ \bottomrule
    \end{tabularx}
\end{table*}


\setlength{\tabcolsep}{3pt} 
\setlength{\extrarowheight}{1pt} 

\begin{table*}[t]
\caption{Cross-evaluation of LiDAR segmentation performance when trained for scene completion. Categories sorted by gain in IoU.}%
\label{tab:semantic_vs_completion}%
\centering
\begin{tabularx}{\linewidth}{@{}lrr|rcrrrrrrrrrrrrr@{}}
\toprule
Trained on & 
\makecell[cr]{Occupied \\ IoU} & \makecell[cr]{Semantic \\ mIoU} &%
\makecell[cr]{Single frame\\mIoU} & %
& %
Diff. & %
\rotsemXXX{M.cycle} & %
\rotsemXXX{Truck} & %
\rotsemXXX{Bicyclist} & %
\rotsemXXX{Person} & %
\rotsemXXX{Pole} & %
\rotsemXXX{Other veh.} & %
\rotsemXXX{Trunk} & %
\rotsemXXX{Parking} & %
\rotsemXXX{Bicycle} & %
\rotsemXXX{Tr. Sign} & %
\rotsemXXX{(Others)} & %
\rotsemXXX{Fence} %
\\ \midrule
Time accumulated & \textbf{57.8} & \textbf{26.1} & 50.0 & %
\multirow{2}{*}{\rotatebox[origin=c]{90}{\scalebox{2}{$\lcurvearrowleft$}}} & %
\multirow{2}{*}{\textbf{+5.9}} & %
\multirow{2}{*}{+30.7} & %
\multirow{2}{*}{+16.8} & %
\multirow{2}{*}{+16.0} & %
\multirow{2}{*}{+12.8} & %
\multirow{2}{*}{+9.5} & %
\multirow{2}{*}{+9.2} & %
\multirow{2}{*}{+7.2} & %
\multirow{2}{*}{+4.5} & %
\multirow{2}{*}{+3.3} & %
\multirow{2}{*}{+2.4} & %
\multirow{2}{*}{($<2.0$)} & %
\multirow{2}{*}{-1.2} %
\\
Single frame & %
12.5 & 9.9 & \textbf{55.9} & \\ \bottomrule
\end{tabularx}
\end{table*}

\section{Quantitative results of ablation study\label{sec:quan}}

The quantitative results of the ablation study of section 4.5 in the manuscript
are listed in \tabref{tab:ablations_full}.
The listing includes the class-individual IoU values.

\section{Impact of scene completion training data}%

\tabref{tab:semantic_vs_completion} includes the comparison
of the single frame segmentation performance as discussed in section 4.5 \emph{impact of scene completion training data} in the manuscript.

\section{Qualitative Results\label{sec:qual}}

\begin{figure*}[p]
\setlength{\tabcolsep}{7.0pt} 
\centering
\begin{tabular}{@{}lllllllllllllllllll@{}}
         \rotsemXX{\semcolor[road] Road} &
         \rotsemXX{\semcolor[sidewalk] Sidewalk} &
         \rotsemXX{\semcolor[parking] Parking} &
         \rotsemXX{\semcolor[otherground] other gr.} &
         \rotsemXX{\semcolor[building] Building} &
         \rotsemXX{\semcolor[fence] Fence} &
         \rotsemXX{\semcolor[car] Car} &
         \rotsemXX{\semcolor[truck] Truck} &
         \rotsemXX{\semcolor[othervehicle] other veh.} &
         \rotsemXX{\semcolor[bicycle] Bicycle} &
         \rotsemXX{\semcolor[motorcycle] Motorcycle} &
         \rotsemXX{\semcolor[person] Person} &
         \rotsemXX{\semcolor[bicyclist] Bicyclist} &
         \rotsemXX{\semcolor[motorcyclist] Motorcycl.} &
         \rotsemXX{\semcolor[vegetation] Vegetation} &
         \rotsemXX{\semcolor[trunk] Trunk} &
         \rotsemXX{\semcolor[terrain] Terrain} &
         \rotsemXX{\semcolor[pole] Pole} &
         \rotsemXX{\semcolor[trafficsign] Tr. Sign} \\
\end{tabular}
\setlength{\tabcolsep}{1pt} 
\def\scalemargin{0.04}
\def\scalemarginh{0.04}
\def\scalesection{12.5 / 102.4}
\def\scaleheight{0.02}
\def\customrowspace{-.5mm}
\begin{tabularx}{\linewidth}{@{}XXXX@{}}
\includegraphics[width=\linewidth]{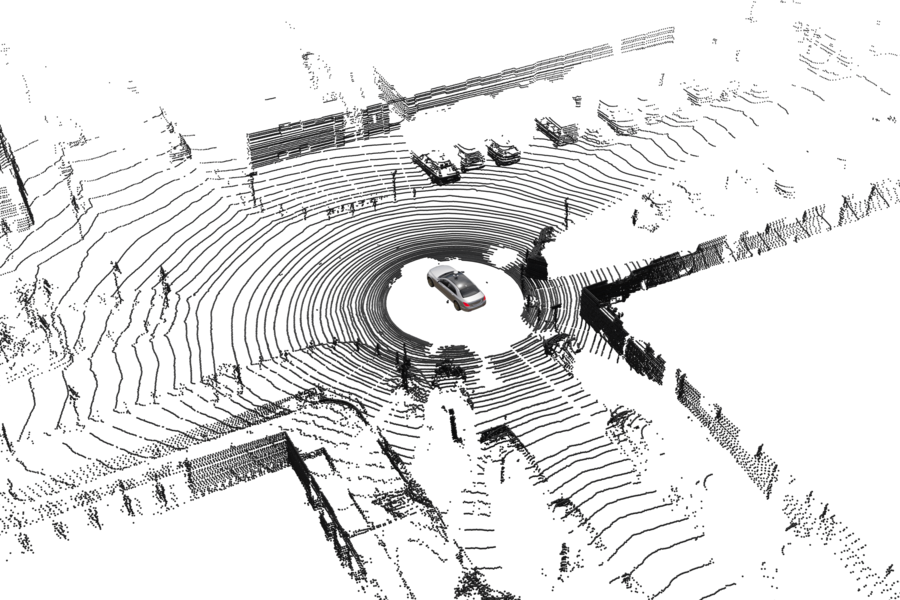} &
\includegraphics[width=\linewidth]{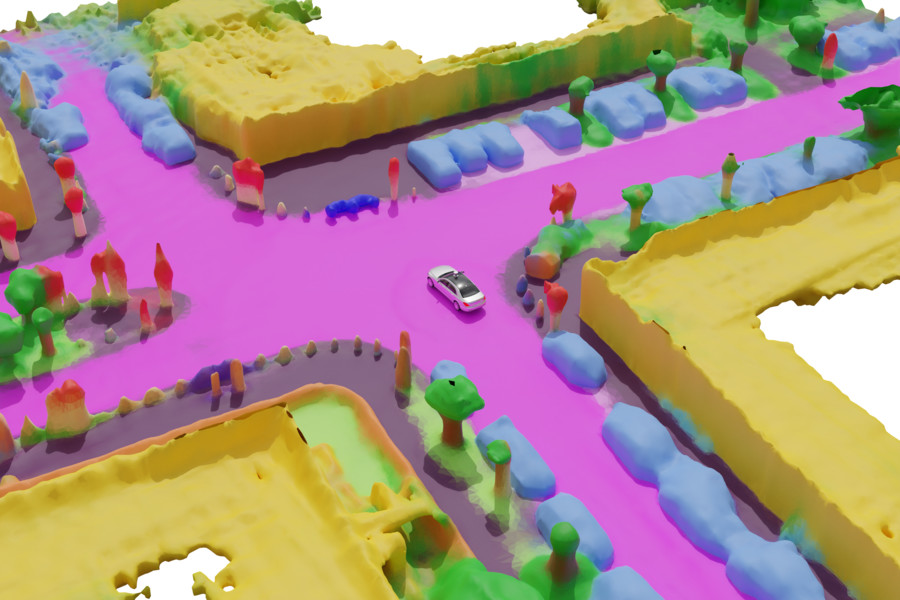} &
\makescaleimage{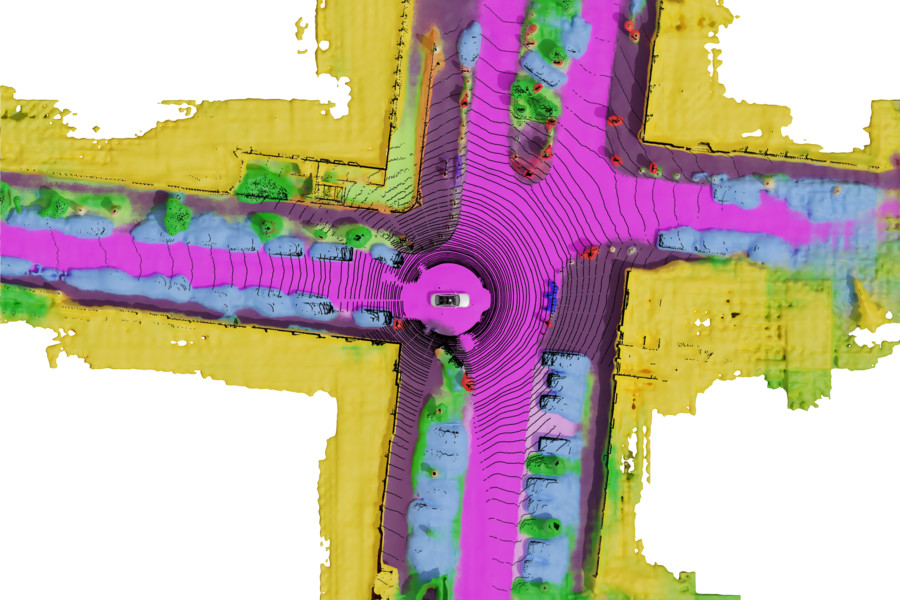}{25}{50} &
\makescaleimage{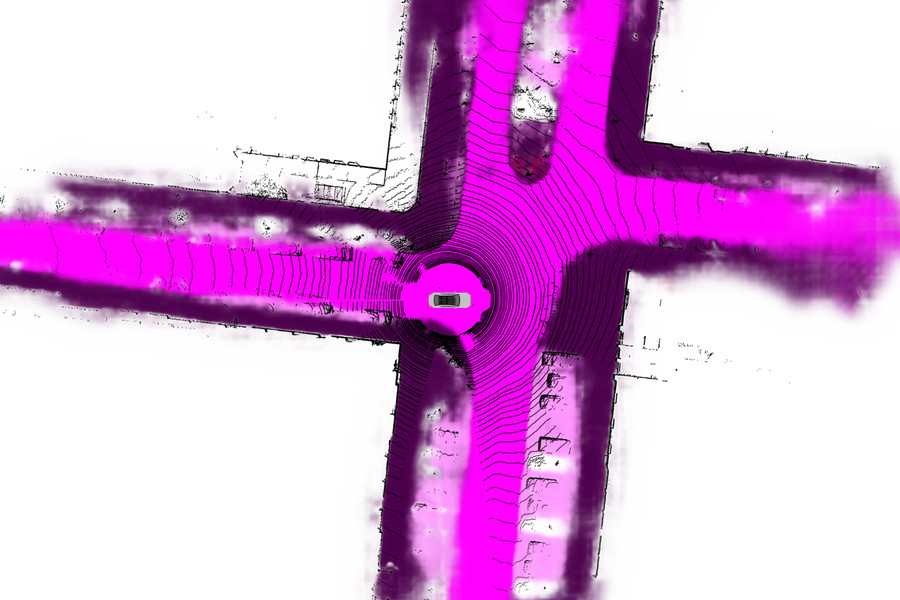}{25}{50} \\[\customrowspace]
\includegraphics[width=\linewidth]{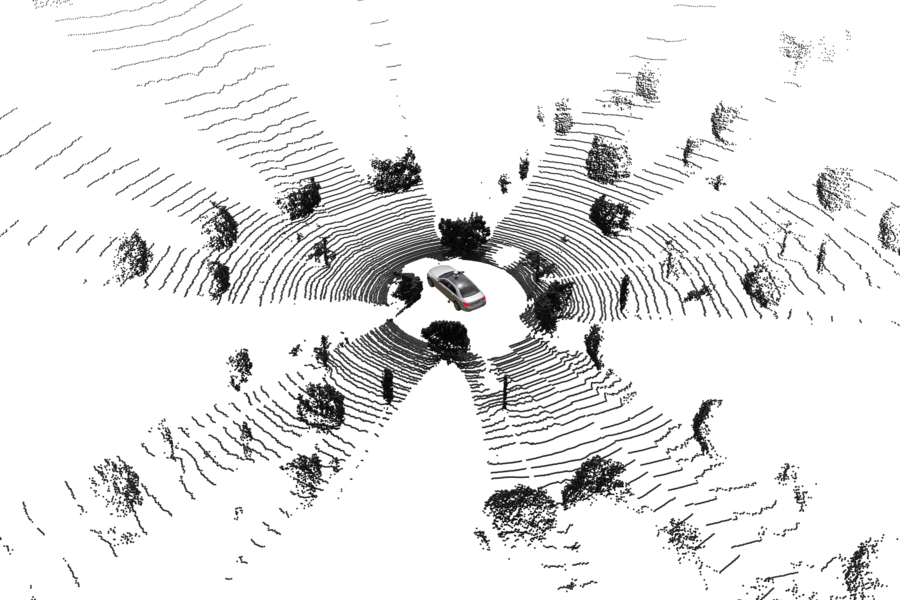} &
\includegraphics[width=\linewidth]{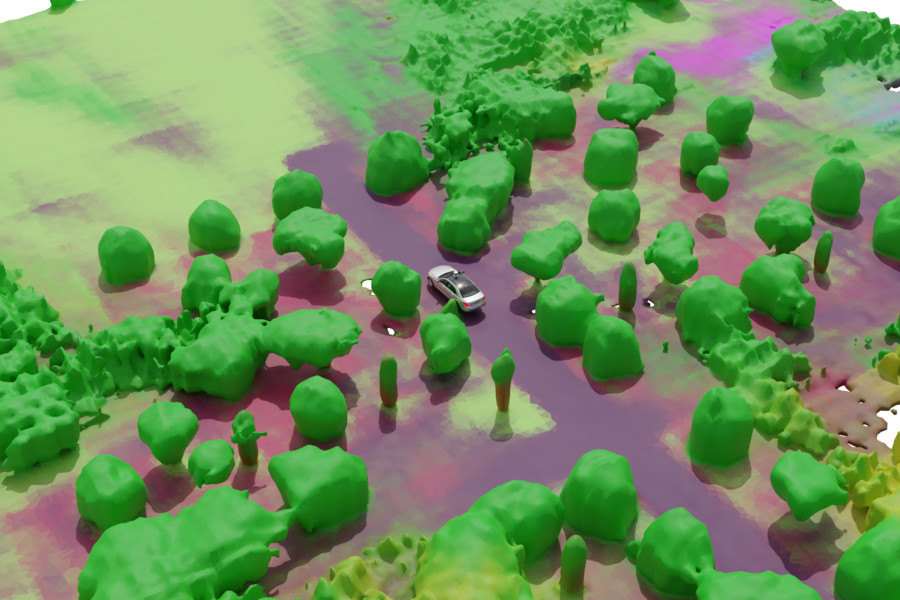} &
\makescaleimage{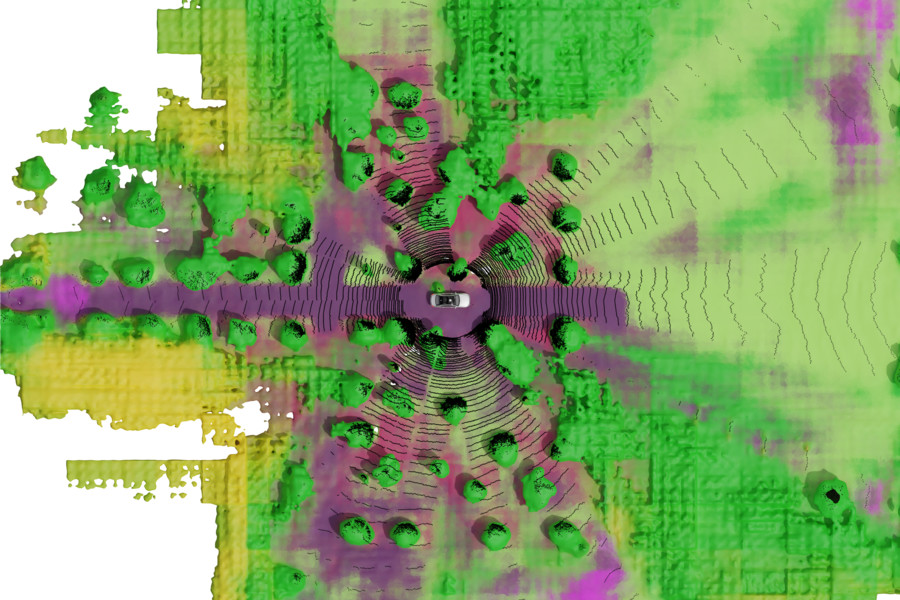}{25}{50} &
\makescaleimage{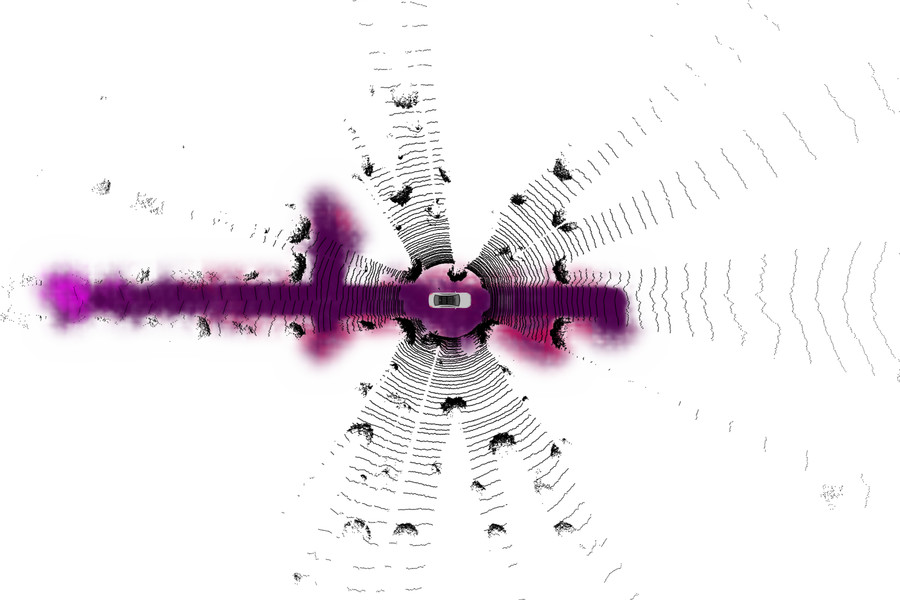}{25}{50} \\[\customrowspace]
\includegraphics[width=\linewidth]{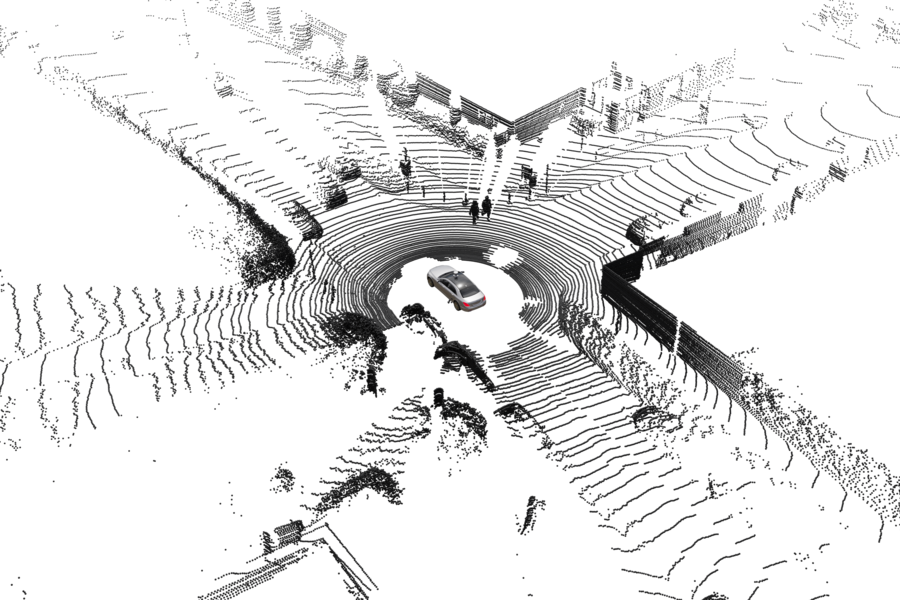} &
\includegraphics[width=\linewidth]{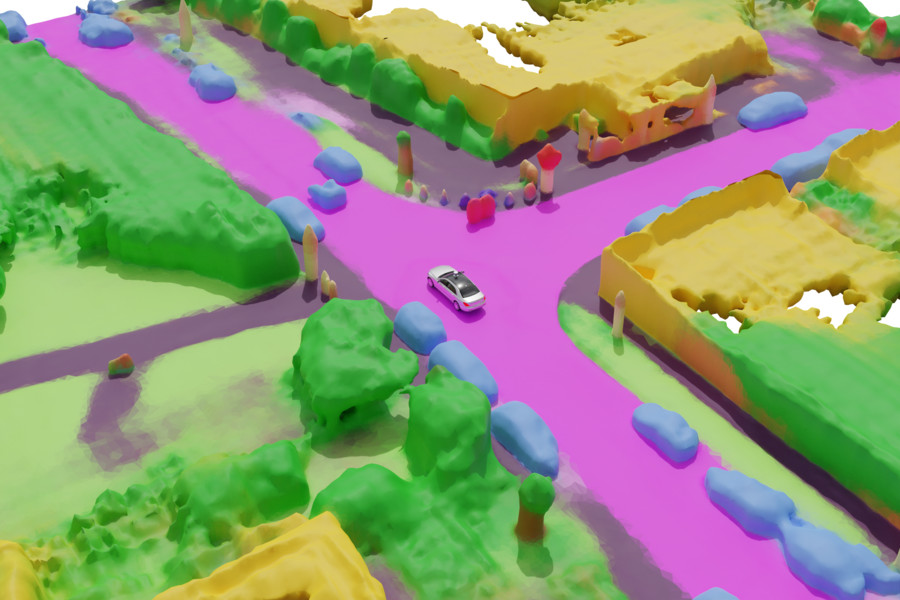} &
\makescaleimage{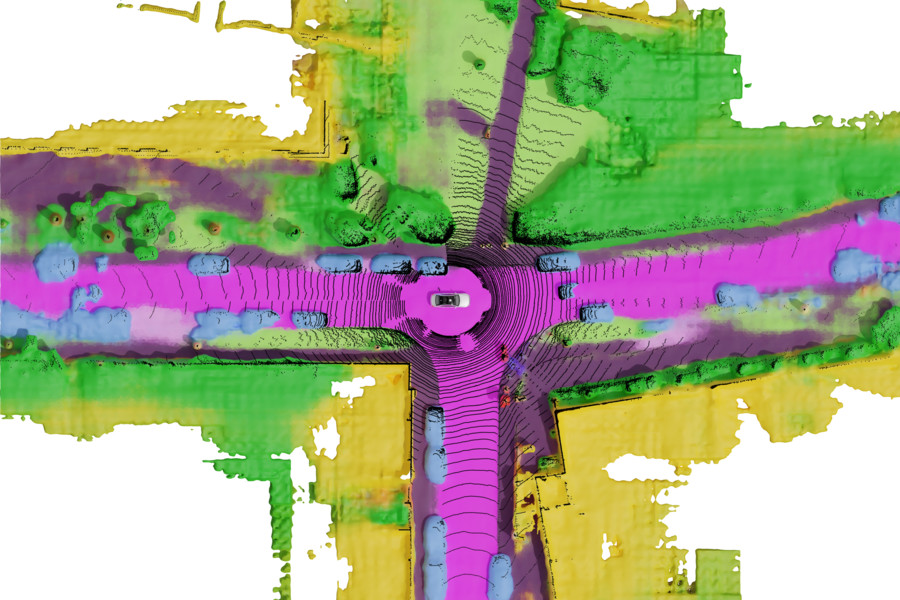}{25}{50} &
\makescaleimage{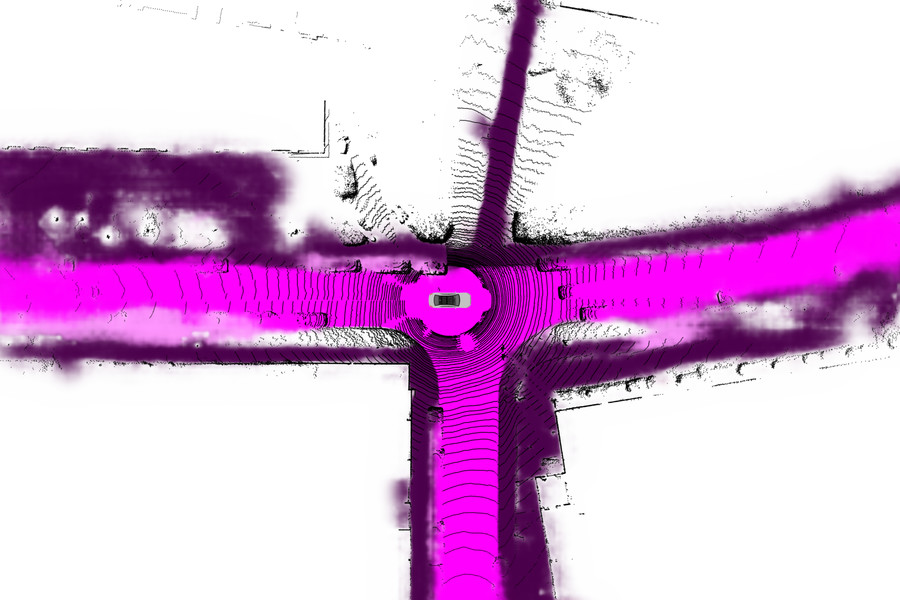}{25}{50} \\[\customrowspace]
\includegraphics[width=\linewidth]{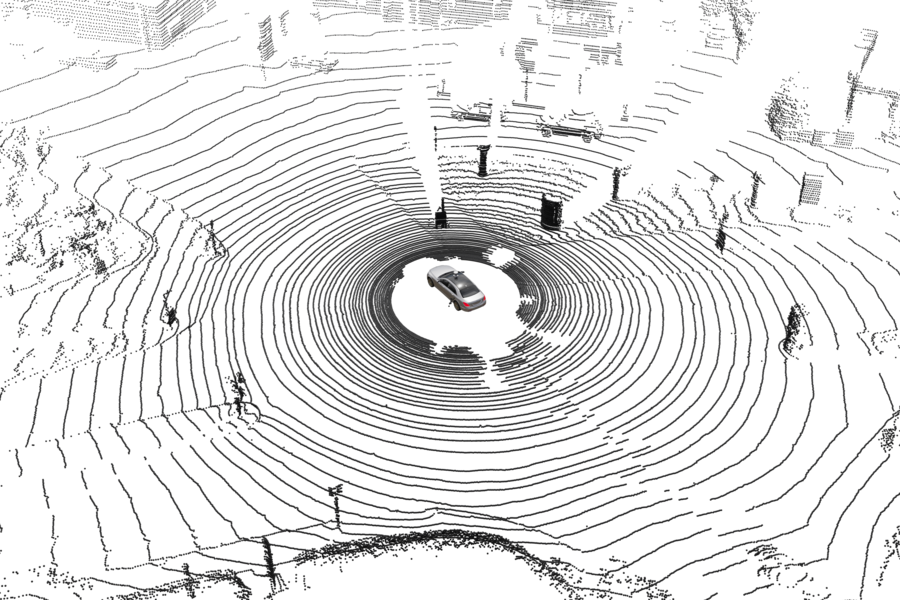} &
\includegraphics[width=\linewidth]{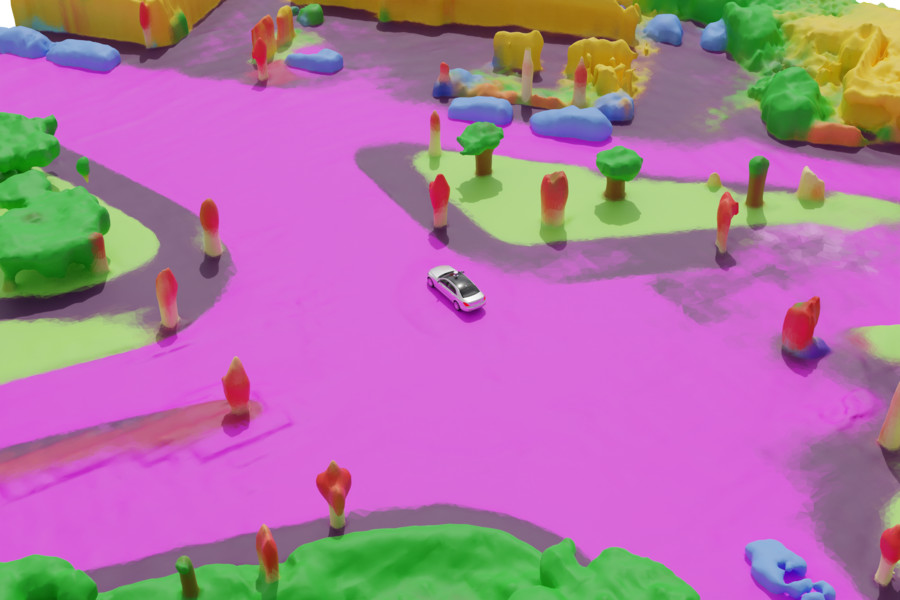} &
\makescaleimage{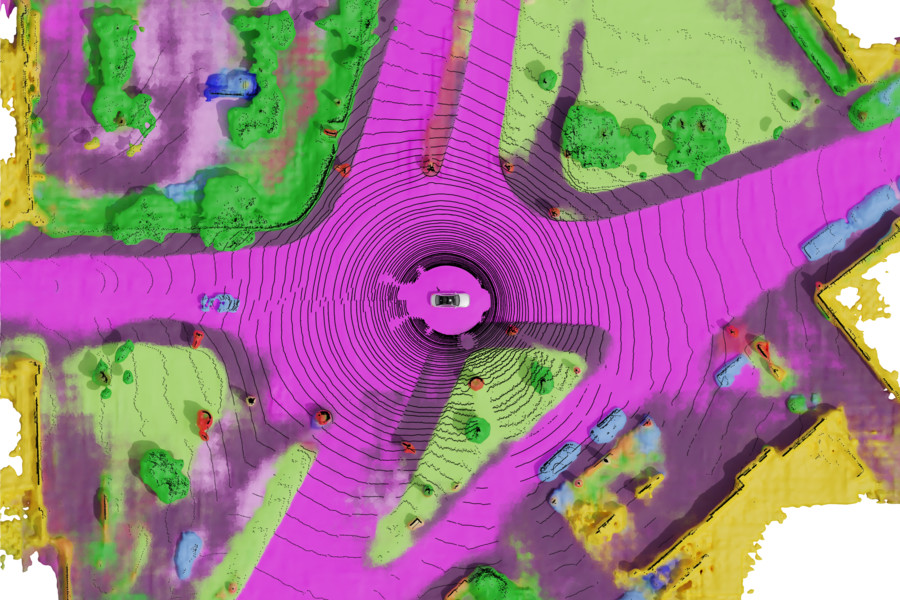}{25}{50} &
\makescaleimage{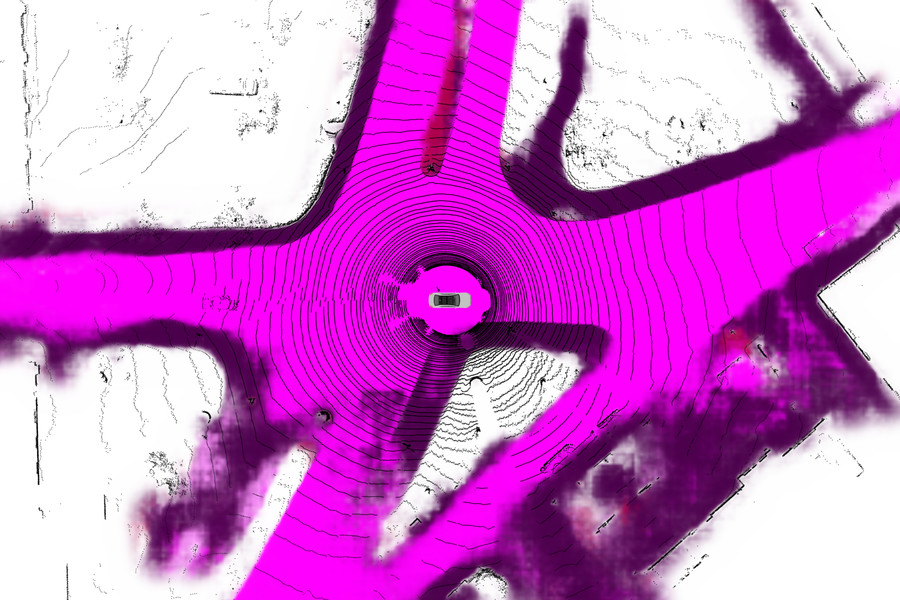}{25}{50} \\[\customrowspace]
\includegraphics[width=\linewidth]{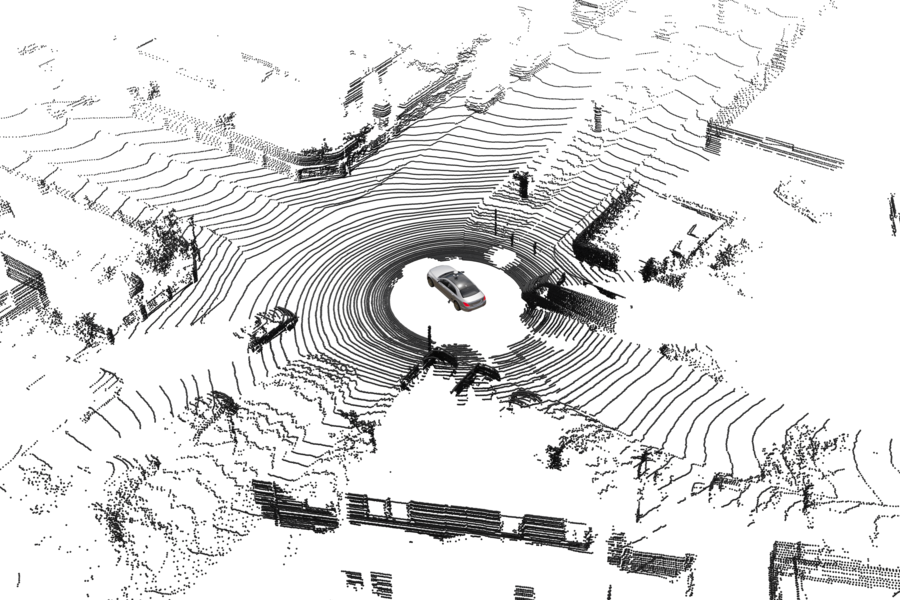} &
\includegraphics[width=\linewidth]{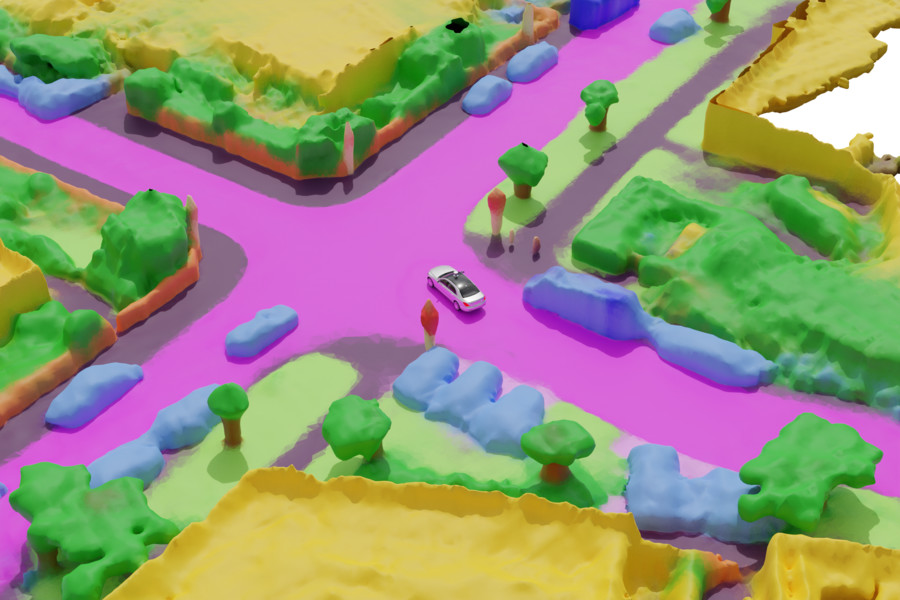} &
\makescaleimage{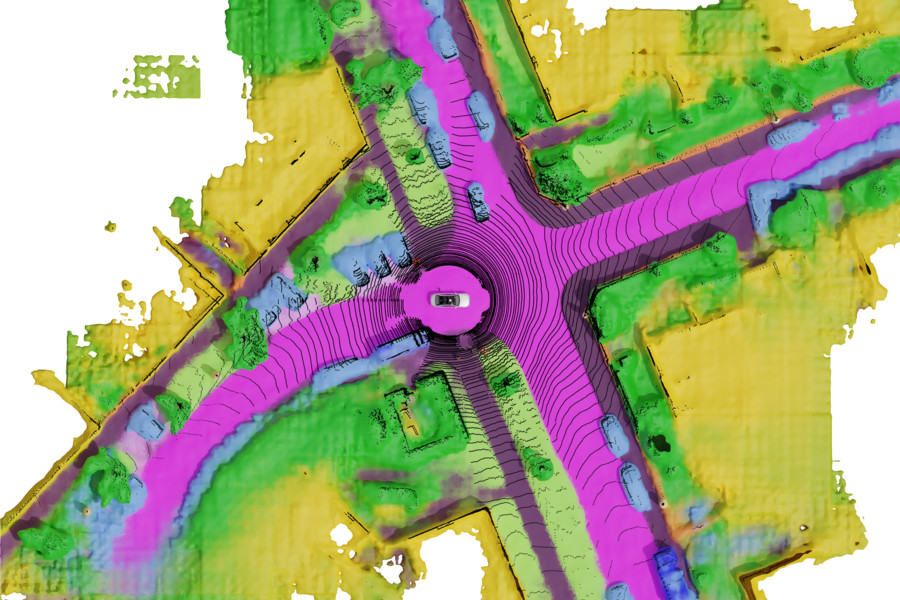}{25}{50} &
\makescaleimage{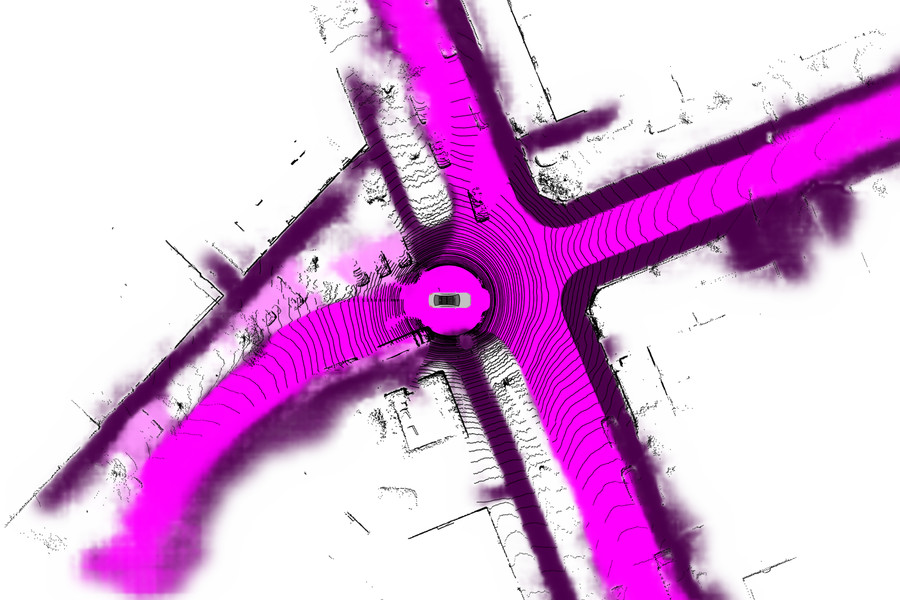}{25}{50} \\[\customrowspace]
\includegraphics[width=\linewidth]{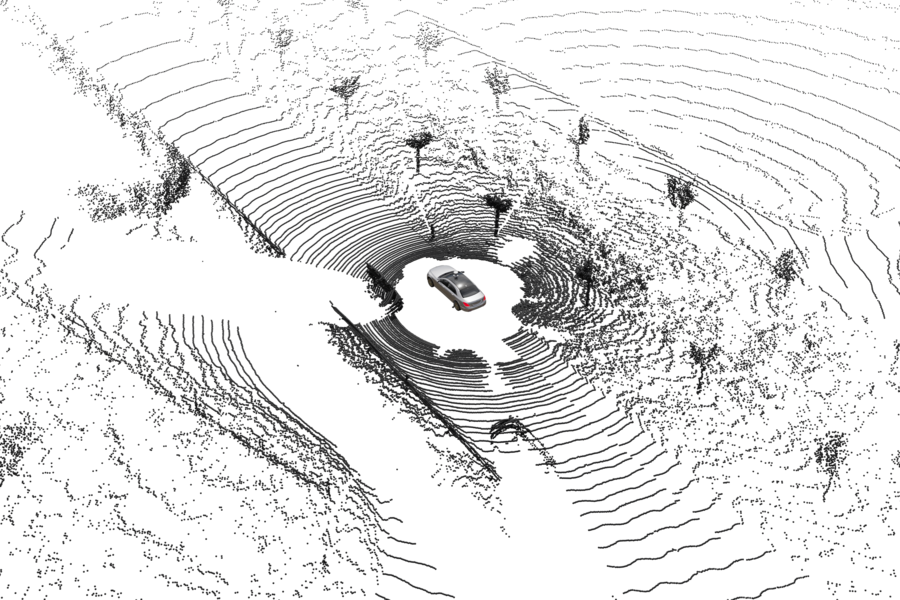} &
\includegraphics[width=\linewidth]{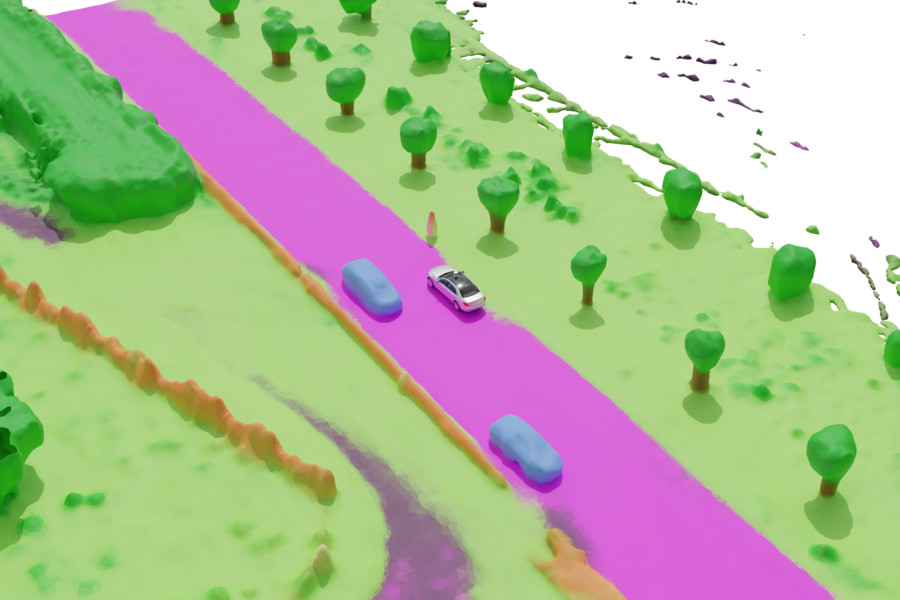} &
\makescaleimage{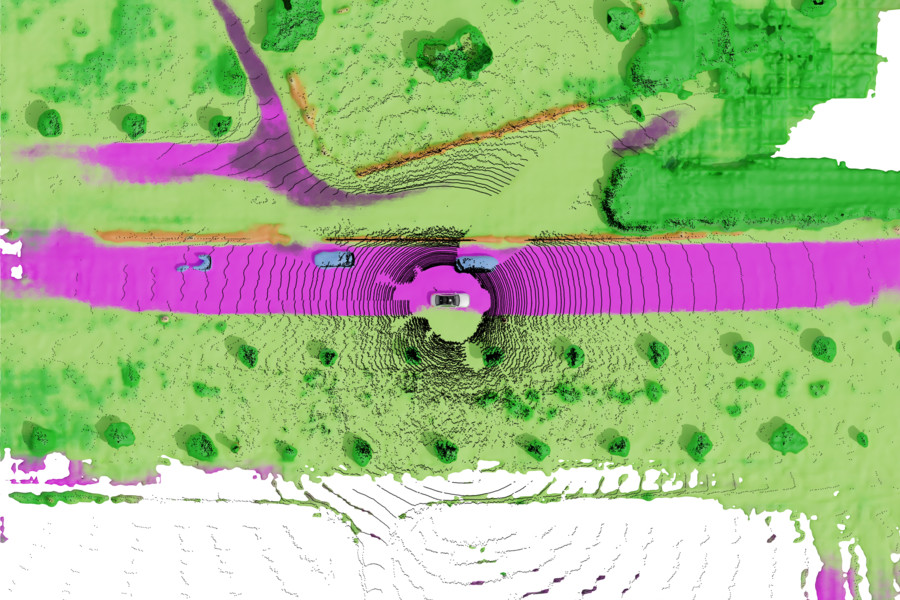}{25}{50} &
\makescaleimage{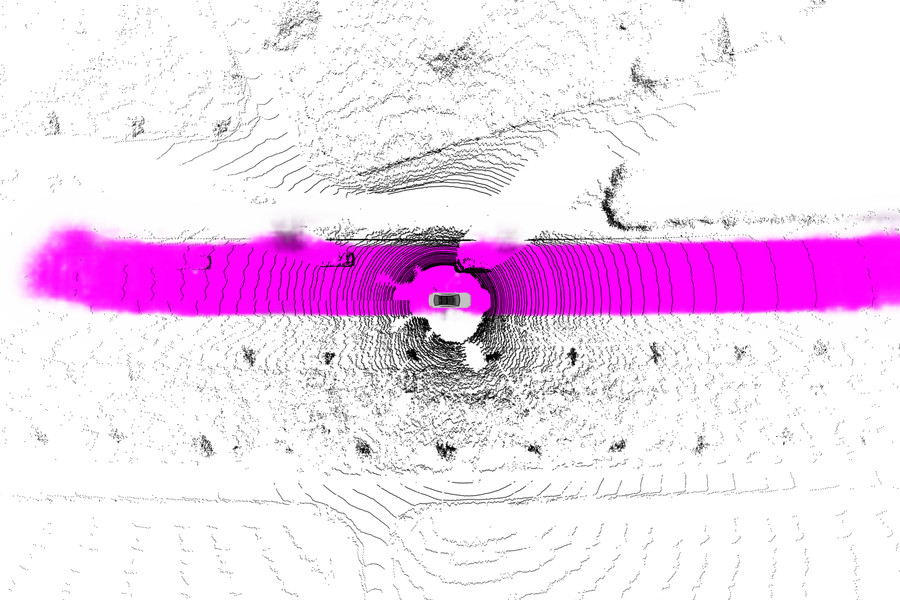}{25}{50} \\[\customrowspace]
\includegraphics[width=\linewidth]{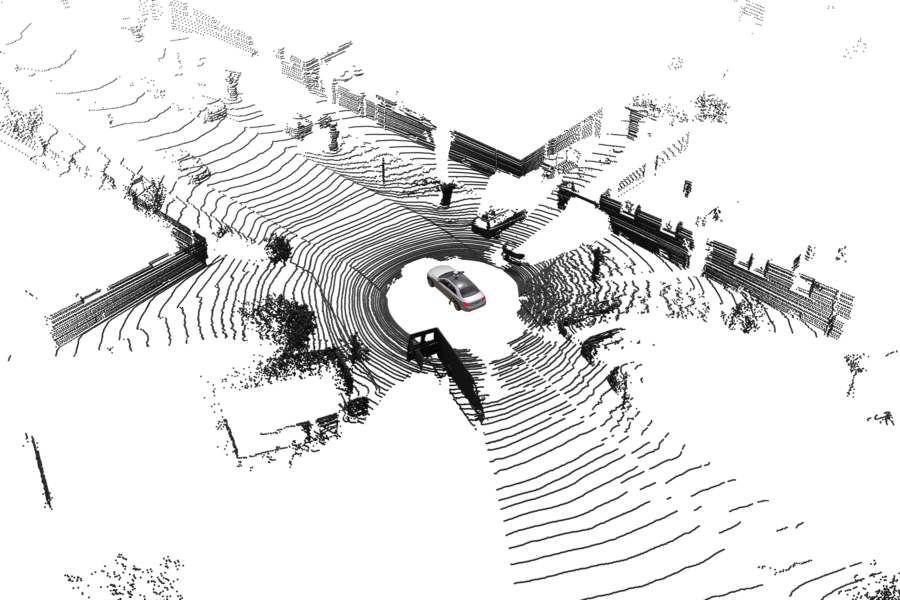}%
&%
\includegraphics[width=\linewidth]{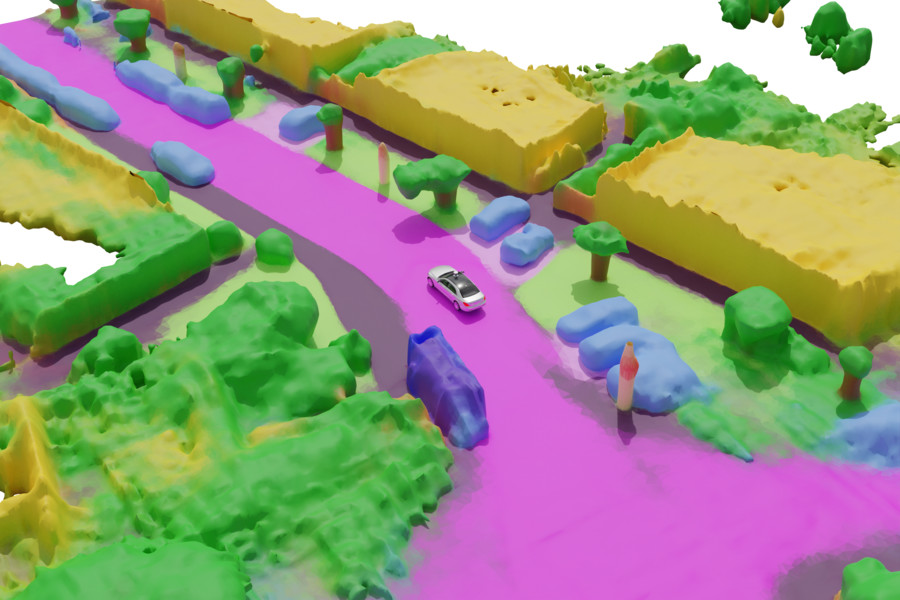}%
& %
\makescaleimage{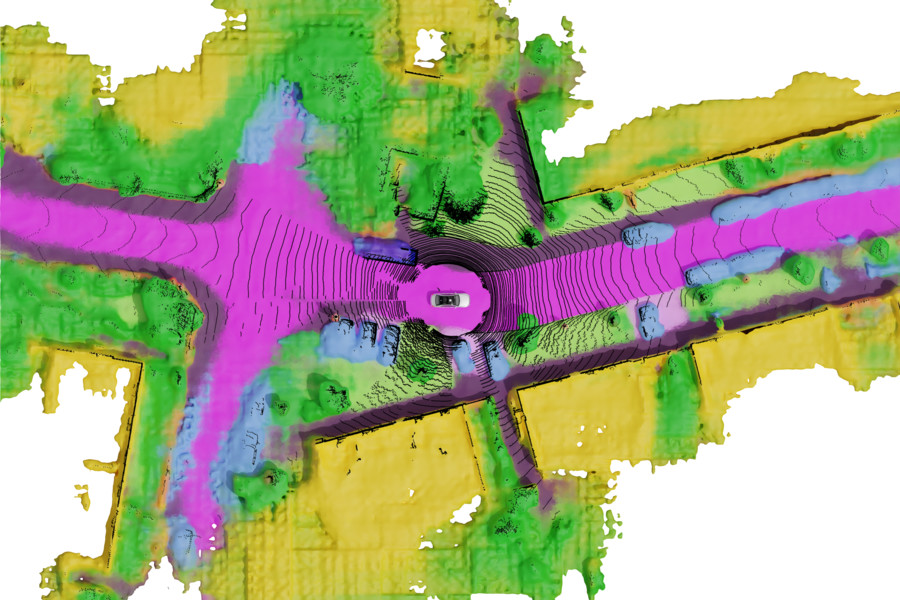}{25}{50}%
&%
\makescaleimage{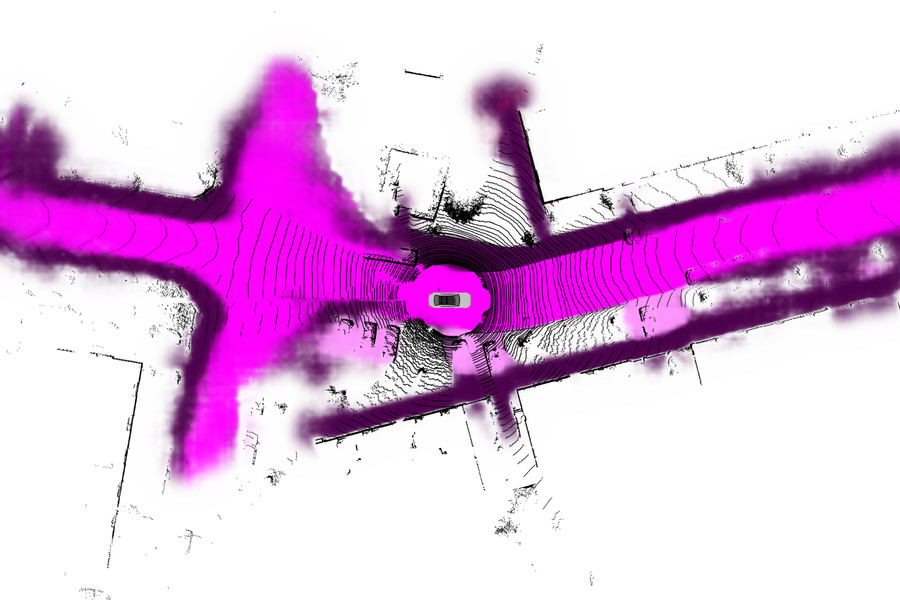}{25}{50}%
\\
\end{tabularx}
\caption{Qualitative results on the Semantic~KITTI test set. Columns from left to right: Input LiDAR scan, completed scene, top view, and ground segmentation.
The selected scenes for each row highlight the diversity of the test sequences.
The scene completion function is derived from a single LiDAR input scan
(leftmost column).
The renderings of the free space isosurface give an intuition how the function behaves for the different scenes.
The ground plane is estimated from semantic predictions and segmented into classes road, sidewalk,
parking space, and other ground.
The top-view images cover an extent of \SI{51.2}{\meter} behind
and in front of the ego-vehicle.}
\label{fig:showcase}
\end{figure*}

We visualize the scene completion function on example scenes of the Semantic~KITTI test set in \figref{fig:showcase}.
The scene extent is not limited to the scene size of the Semantic~KITTI benchmark,
but instead quadrupled in terms of area to cover the full \SI{360}{\degree} LiDAR
scan. The corresponding ground level segmentation is displayed in the rightmost column.


\end{document}